\documentclass[10pt,twocolumn,letterpaper]{article}

\usepackage[pagenumbers]{cvpr} 

\usepackage{graphicx}
\usepackage{amsmath}
\usepackage{amssymb}
\usepackage{booktabs}
\usepackage{array}
\usepackage{multirow}
\usepackage[table]{xcolor}

\usepackage[pagebackref,breaklinks,colorlinks]{hyperref}

\usepackage[capitalize]{cleveref}
\crefname{section}{Sec.}{Secs.}
\Crefname{section}{Section}{Sections}
\Crefname{table}{Table}{Tables}
\crefname{table}{Tab.}{Tabs.}

\DeclareMathOperator*{\argmin}{arg\,min}

\newcommand{\EE}{\mathbb{E}}
\newcommand{\RR}{\mathbb{R}}

\newcommand{\cC}{\mathcal{C}}
\newcommand{\cL}{\mathcal{L}}

\newcommand{\vece}{\mathbf{e}}
\newcommand{\vech}{\mathbf{h}}

\newcommand{\vecr}{\mathbf{r}}

\newcommand{\vecu}{\mathbf{u}}
\newcommand{\vecv}{\mathbf{v}}

\newcommand{\vecz}{\mathbf{z}}

\newcommand{\bM}{\mathbf{M}}
\newcommand{\bS}{\mathbf{S}}
\newcommand{\bX}{\mathbf{X}}
\newcommand{\bZ}{\mathbf{Z}}

\newcommand{\sg}{\mathrm{sg}}
\newcommand{\given}{{\mkern1.5mu | \mkern1.5mu}}

\newcommand{\Q}{\mathcal{Q}}
\newcommand{\RQ}{\mathcal{R\mkern-1.5mu Q}}
\newcommand{\ARmodel}{RQ-Transformer}

\newcommand{\Nsp}{N_{\text{spatial}}}
\newcommand{\Ndep}{N_{\text{depth}}}

\newcommand{\Lrec}{\cL_{\text{recon}}}
\newcommand{\Lcommit}{\cL_{\text{commit}}}

\def\cvprPaperID{10217} 
\def\confName{CVPR}
\def\confYear{2022}

\begin{document}

\title{Autoregressive Image Generation using Residual Quantization}

\author{Doyup Lee\thanks{Equal contribution} \\
POSTECH, Kakao Brain\\
{\tt\small doyup.lee@postech.ac.kr}
\and
Chiheon Kim\footnotemark[1] \\
Kakao Brain\\
{\tt\small chiheon.kim@kakaobrain.com}
\and
Saehoon Kim\\
Kakao Brain\\
{\tt\small shkim@kakaobrain.com}
\and
Minsu Cho\\
POSTECH\\
{\tt\small mscho@postech.ac.kr}
\and
Wook-Shin Han\thanks{Corresponding author}\\
POSTECH\\
{\tt\small wshan@postech.ac.kr}
}

\maketitle

\begin{abstract}
For autoregressive (AR) modeling of high-resolution images, vector quantization (VQ) represents an image as a sequence of discrete codes. A short sequence length is important for an AR model to reduce its computational costs to consider long-range interactions of codes. However, we postulate that previous VQ cannot shorten the code sequence and generate high-fidelity images together in terms of the rate-distortion trade-off. In this study, we propose the two-stage framework, which consists of Residual-Quantized VAE (RQ-VAE) and RQ-Transformer, to effectively generate high-resolution images. Given a fixed codebook size, RQ-VAE can precisely approximate a feature map of an image and represent the image as a stacked map of discrete codes. Then, RQ-Transformer learns to predict the quantized feature vector at the next position by predicting the next stack of codes. Thanks to the precise approximation of RQ-VAE, we can represent a 256$\times$256 image as 8$\times$8 resolution of the feature map, and RQ-Transformer can efficiently reduce the computational costs. Consequently, our framework outperforms the existing AR models on various benchmarks of unconditional and conditional image generation. Our approach also has a significantly faster sampling speed than previous AR models to generate high-quality images.

\end{abstract}


\begin{figure}
    \centering
    \includegraphics[height=0.475\textwidth, width=0.475\textwidth]{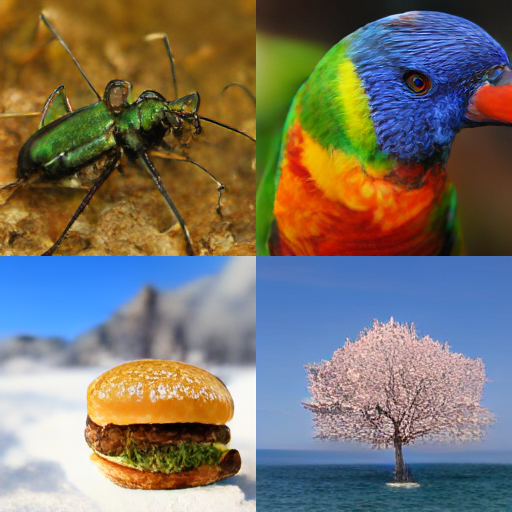}
    \caption{Examples of our conditional generation for 256$\times$256 images. The images in the first row are generated from the classes of ImageNet. The images in the second row are generated from text conditions (``A cheeseburger in front of a mountain range covered with snow.'' and ``a cherry blossom tree on the blue ocean''). The text conditions are unseen during the training.}
    \label{fig:teaser}
\end{figure}

\section{Introduction}
\label{sec:intro}
Vector quantization (VQ) becomes a fundamental technique for autoregerssive (AR) models to generate high-resolution images \cite{VQVAE,iGPT,VQGAN,DALL-E,ImageBART}.
Specifically, an image is represented as a sequence of discrete codes, after the feature map of the image is quantized by VQ and rearranged by an ordering such as raster scan~\cite{PixelRNN}.
After the quantization, AR model is trained to sequentially predict the codes in the sequence.
That is, AR models can generate high-resolution images without predicting whole pixels in an image.

We postulate that reducing the sequence length of codes is important for AR modeling of images.
A short sequence of codes can significantly reduce the computational costs of an AR model, since an AR uses the codes in previous positions to predict the next code.
However, previous studies have a limitation to reducing the sequence length of images in terms of the rate-distortion trade-off~\cite{shannon1959coding}.
Namely, VQ-VAE~\cite{VQVAE} requires an exponentially increasing size of codebook to reduce the resolution of the quantized feature map, while conserving the quality of reconstructed images.
However, a huge codebook leads to the increase of model parameters and the codebook collapse problem~\cite{Jukebox}, which makes the training of VQ-VAE unstable.

In this study, we propose a \emph{Residual-Quantized VAE (RQ-VAE)}, which uses a residual quantization (RQ) to precisely approximate the feature map and reduce its spatial resolution.
Instead of increasing the codebook size, RQ uses a fixed size of codebook to recursively quantize the feature map in a coarse-to-fine manner.
After $D$ iterations of RQ, the feature map is represented as a stacked map of $D$ discrete codes.
Since RQ can compose as many vectors as the codebook size to the power of $D$, RQ-VAE can precisely approximate a feature map, while conserving the information of the encoded image without a huge codebook.
Thanks to the precise approximation, RQ-VAE can further reduce the spatial resolution of the quantized feature map than previous studies~\cite{VQVAE,VQGAN,DALL-E}.
For example, our RQ-VAE can use 8$\times$8 resolution of feature maps for AR modeling of 256$\times$256 images.

In addition, We propose \emph{RQ-Transformer} to predict the codes extracted by RQ-VAE.
For the input of RQ-Transformer, the quantized feature map in RQ-VAE is converted into a sequence of feature vectors.
Then, RQ-Transformer predicts the next $D$ codes to estimate the feature vector at the next position.
Thanks to the reduced resolution of feature maps by RQ-VAE, RQ-Transformer can significantly reduce the computational costs and easily learn the long-range interactions of inputs.
We also propose two training techniques for RQ-Transformer, \emph{soft labeling} and \emph{stochastic sampling} for the codes of RQ-VAE.
They further improve the performance of RQ-Transformer by resolving the exposure bias~\cite{ExposureBias} in the training of AR models.
Consequently, as shown in Figure~\ref{fig:teaser}, our model can generate high-quality images.

Our main contributions are summarized as follows. 1) We propose RQ-VAE, which represents an image as a stacked map of discrete codes, while producing high-fidelity reconstructed images. 2) We propose RQ-Transformer to effectively predict the codes of RQ-VAE and its training techniques to resolve the exposure bias. 3) We show that our approach outperforms previous AR models and significantly improves the quality of generated images, computational costs, and sampling speed.


\section{Related Work}

\paragraph{AR Modeling for Image Synthesis}
AR models have shown promising results of image generation~\cite{VQVAE,iGPT,DCTransformer,VQGAN,VQVAE2,ImageBART} as well as text~\cite{GPT3} and audio~\cite{Jukebox} generation.
AR modeling of raw pixels is possible~\cite{PixelCNN,PixelCNN++,PixelRNN,iGPT}, but it is infeasible for high-resolution images due to the slow speed and low quality of generated images.
Thus, previous studies incorporate VQ-VAE~\cite{VQGAN}, which uses VQ to represent an image as discrete codes, and uses an AR model to predict the codes of VQ-VAE.
VQ-GAN~\cite{VQGAN}, improves the perceptual quality of reconstructed images using adversarial~\cite{GAN,isola2017image} and perceptual loss~\cite{LedigC2017vggloss}.
However, when the resolution of the feature map is further reduced, VQ-GAN cannot precisely approximate the feature map of an image due to the limited size of codebook.

\paragraph{VQs in Other Applications}
Composite quantizations have been used in other applications to represent a vector as a composition of codes for the precise approximation under a given codebook size~\cite{PQ,AQ,SQ,li2021trq,li2017performance,ferdowsi2017regularized}.
For the nearest neighbor search, product quantization (PQ)~\cite{PQ} approximates a vector as the sum of linearly independent vectors in the codebook.
As a generalized version of PQ, additive quantization (AQ)~\cite{AQ} uses the dependent vectors in the codebook, but finding the codes is an NP-hard task~\cite{cooper1990computational}.
Residual quantization (RQ, also known as stacked quantization)~\cite{juang1982multiple,SQ} iteratively quantizes a vector and its residuals and represents the vector as a stack of codes, which has been used for neural network compression~\cite{li2021trq,li2017performance,ferdowsi2017regularized}.
For AR modeling of images, our RQ-VAE adopts RQ to discretize the feature map of an image.
However, different from previous studies, RQ-VAE uses a single shared codebook for all quantization steps.

\section{Methods} \label{sec:methods}
\begin{figure*}
    \centering
    \includegraphics[width=6.7in]{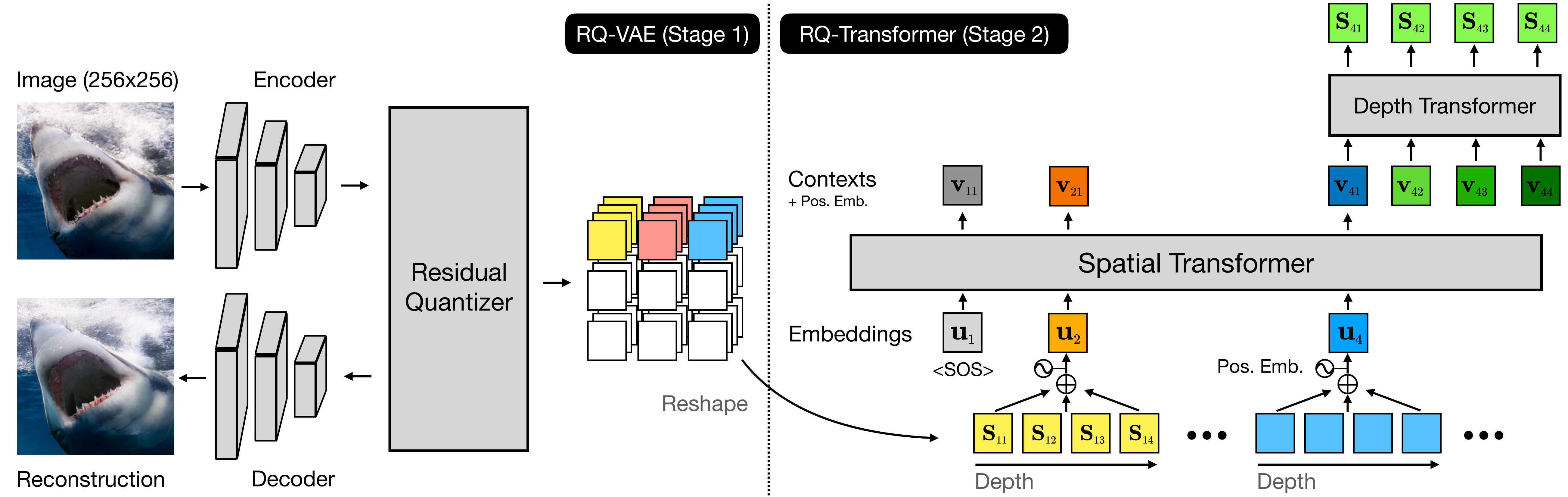}
    \caption{An overview of our two-stage image generation framework composed of RQ-VAE and RQ-Transformer. In stage 1, RQ-VAE uses the residual quantizer to represent an image as a stack of $D=4$ codes. After the stacked map of codes is reshaped, RQ-Transformer predicts the $D$ codes at the next position. More details are available in Section~\ref{sec:methods}.}
    \label{fig:SQ_overview}
\end{figure*}

We propose the two-stage framework with \emph{RQ-VAE} and \emph{\ARmodel} for AR modeling of images (see Figure~\ref{fig:SQ_overview}).
RQ-VAE uses a codebook to represent an image as a stacked map of $D$ discrete codes.
Then, our \ARmodel\ autoregressively predicts the next $D$ codes at the next spatial position.
We also introduce how our RQ-Transformer resolves the exposure bias~\cite{ExposureBias} in the training of AR models.

\subsection{Stage 1: Residual-Quantized VAE}
In this section, we first introduce the formulation of VQ and VQVAE.
Then, we propose RQ-VAE, which can precisely approximate a feature map without increasing the codebook size, and explain how RQ-VAE represents an image as a stacked map of discrete codes.

\subsubsection{Formulation of VQ and VQ-VAE}
Let a \emph{codebook} $\cC$ be a finite set $\{(k, \vece(k))\}_{k \in [K]}$, which consists of the pairs of a \emph{code} $k$ and its \emph{code embedding} $\vece(k) \in \RR^{n_z}$, where $K$ is the codebook size and $n_z$ is the dimensionality of code embeddings.
Given a vector $\vecz \in \RR^{n_z}$, $\Q (\vecz; \cC)$ denotes VQ of $\vecz$, which is the code whose embedding is nearest to $z$, that is, 
\begin{equation}
\label{eq:Q}
    \Q(\vecz; \cC) = \argmin_{k \in [K]} \|\vecz - \vece(k)\|_2^2.
\end{equation}

After VQ-VAE encodes an image into a discrete code map, VQ-VAE reconstructs the original image from the encoded code map.
Let $E$ and $G$ be an encoder and a decoder of VQ-VAE.
Given an image $\bX \in \RR^{H_{o} \times W_{o} \times 3}$, VQ-VAE extracts the feature map $\bZ = E(\bX) \in \RR^{H \times W \times n_z}$, where $(H, W)=(H_{o}/f, W_{o}/f)$ is the spatial resolution of $\bZ$, and $f$ is a downsampling factor.
By applying the VQ to each feature vector at each position, VQ-VAE quantizes $\bZ$ and returns its code map $\bM \in [K]^{H\times W}$ and its quantized feature map $\hat{\bZ} \in \RR^{H\times W \times n_z}$ as
\begin{equation}
    \bM_{hw} = \Q (\bZ_{hw}; \cC), \quad
    \hat{\bZ}_{hw} = \vece(\bM_{hw}),
\end{equation}
where $\bZ_{hw} \in \RR^{n_z}$ is a feature vector at $(h,w)$, and $\bM_{hw}$ is its code.
Finally, the input is reconstructed as $\hat{\bX} = G(\hat\bZ)$.

We remark that reducing the spatial resolution of $\hat{\bZ}$, $(H,W)$, is important for AR modeling, since the computational cost of an AR model increases with $HW$.
However, since VQ-VAE conducts a lossy compression of images, there is a trade-off between reducing $(H,W)$ and conserving the information of $\bX$.
Specifically, VQ-VAE with the codebook size $K$ uses $HW\log_2 K$ bits to represent an image as the codes.
Note that the best achievable reconstruction error depends on the number of bits in terms of the rate-distortion theory~\cite{shannon1959coding}.
Thus, to further reduce $(H,W)$ to $(H/2, W/2)$ but preserve the reconstruction quality, VQ-VAE requires the codebook of size $K^4$.
However, VQ-VAE with a large codebook is inefficient due to the codebook collapse problem~\cite{Jukebox} with unstable training.

\subsubsection{Residual Quantization} \label{section:RQ}
Instead of increasing the codebook size, we adopt a residual quantization (RQ) to discretize a vector $\vecz$.
Given a quantization depth $D$, RQ represents $\vecz$ as an \emph{ordered} $D$ codes
\begin{equation}
\label{eq:SQ}
    \RQ (\vecz; \cC, D) = (k_1, \cdots, k_D) \in [K]^D,
\end{equation}
where $\cC$ is the codebook of size $|\cC|=K$, and $k_d$ is the code of $\vecz$ at depth $d$.
Starting with $0$-th residual $\vecr_0 = \vecz$, RQ recursively computes $k_d$, which is the code of the residual $\vecr_{d-1}$, and the next residual $\vecr_d$ as 
\begin{equation} \label{eq:SQ_details}
\begin{aligned}
    k_d &= \Q(\vecr_{d-1}; \cC), \\
    \vecr_d &= \vecr_{d-1} - \vece(k_d),
\end{aligned}
\end{equation}
for $d=1,\cdots,D$.
In addition, we define $\hat{\vecz}^{( d)}= \sum_{i=1}^d \vece(k_i)$ as the partial sum of up to $d$ code embeddings, and $\hat{\vecz}:=\hat{\vecz}^{( D)}$ is the quantized vector of $\vecz$.

The recursive quantization of RQ approximates the vector $\vecz$ in a coarse-to-fine manner.
Note that $\hat{\vecz}^{( 1)}$ is the closest code embedding $\vece(k_1)$ in the codebook to $\vecz$.
Then, the remaining codes are subsequently chosen to reduce the quantization error at each depth.
Hence, the partial sum up to $d$, $\hat{\vecz}^{( d)}$, provides a finer approximation as $d$ increases.

Although we can separately construct a codebook for each depth $d$, a single shared codebook $\cC$ is used for every quantization depth.
The shared codebook has two advantages for RQ to approximate a vector $\vecz$.
First, using separate codebooks requires an extensive hyperparameter search to determine the codebook size at each depth, but the shared codebook only requires to determine the total codebook size $K$.
Second, the shared codebook makes all code embeddings available for every quantization depth.
Thus, a code can be used at every depth to maximize its utility.

We remark that RQ can more precisely approximate a vector than VQ when their codebook sizes are the same.
While VQ partitions the entire vector space $\RR^{n_z}$ into $K$ clusters, RQ with depth $D$ partitions the vector space into $K^D$ clusters at most.
That is, RQ with $D$ has the same partition capacity as VQ with $K^D$ codes.
Thus, we can increase $D$ for RQ to replace VQ with an exponentially growing codebook.

\subsubsection{RQ-VAE}
In Figure~\ref{fig:SQ_overview}, we propose \emph{RQ-VAE} to precisely quantize a feature map of an image. 
RQ-VAE is also comprised of the encoder-decoder architecture of VQ-VAE, but the VQ module is replaced with the RQ module above.
Specifically, RQ-VAE with depth $D$ represents a feature map $\bZ$ as a stacked map of codes $\bM \in [K]^{H\times W\times D}$ and extracts $\hat{\bZ}^{( d)} \in \mathbb{R}^{H\times W \times n_z}$, which is the quantized feature map at depth $d$ for $d\in[D]$ such that
\begin{equation} \label{eq:depth_sum}
\begin{aligned}
    \bM_{hw} &= \RQ (E(\bX)_{hw}; \cC, D), 
    \\
    \hat{\bZ}_{hw}^{( d)} &= \sum_{d'=1}^d \vece(\bM_{hwd'}).
\end{aligned}
\end{equation}
For brevity, the quantized feature map $\hat{\bZ}^{( D)}$ at depth $D$ is also denoted by $\hat{\bZ}$.
Finally, the decoder $G$ reconstructs the input image from $\hat{\bZ}$ as $\hat{\bX} = G(\hat\bZ)$.

Our RQ-VAE can make AR models to effectively generate high-resolution images with low computational costs.
For a fixed downsampling factor $f$, RQ-VAE can produce more realistic reconstructions than VQ-VAE, since RQ-VAE can precisely approximate a feature map using a given codebook size. 
Note that the fidelity of reconstructed images is critical for the maximum quality of generated images.
In addition, the precise approximation by RQ-VAE allows more increase of $f$ and decrease of $(H,W)$ than VQ-VAE, while preserving the reconstruction quality.
Consequently, RQ-VAE enables an AR model to reduce its computational costs, increase the speed of image generation, and learn the long-range interactions of codes better.

\paragraph{Training of RQ-VAE}
To train the encoder $E$ and the decoder $G$ of RQ-VAE, we use the gradient descent with respect to the loss $\cL = \Lrec + \beta\Lcommit$ with a multiplicative factor $\beta > 0$,
The \emph{reconstruction loss} $\Lrec$ and the \emph{commitment loss} $\Lcommit$ are defined as
\begin{gather}
    \Lrec = \|\bX - \hat{\bX}\|_2^2, \\
    \Lcommit = \sum_{d=1}^{D} \left\|
        \bZ - \sg \left[\hat{\bZ}^{( d)}\right]
    \right\|_2^2,
\end{gather}
where $\sg[\cdot]$ is the stop-gradient operator, and the straight-through estimator~\cite{VQVAE} is used for the backpropagation through the RQ module.
Note that $\Lcommit$ is the sum of quantization errors from every $d$, not a single term $\|\bZ - \sg [\hat{\bZ}]\|_2^2$.
It aims to make $\hat{\bZ}^{(d)}$ sequentially decrease the quantization error of $\bZ$ as $d$ increases.
Thus, RQ-VAE approximates the feature map in a coarse-to-fine manner and keeps the training stable.
The codebook $\cC$ is updated by the exponential moving average of the clustered features~\cite{VQVAE}.

\paragraph{Adversarial Training of RQ-VAE}
RQ-VAE is also trained with adversarial learning to improve the perceptual quality of reconstructed images.
The patch-based adversarial loss~\cite{isola2017image} and the perceptual loss~\cite{johnson2016perceptual} are used together as described in the previous study~\cite{VQGAN}.
We include the details in the supplementary material.

\subsection{Stage 2: \ARmodel}
In this section, we propose \ARmodel\ in Figure~\ref{fig:SQ_overview} to autoregressively predict a code stack of RQ-VAE.
After we formulate the AR modeling of codes extracted by RQ-VAE, we introduce how our \ARmodel\ efficiently learns the stacked map of discrete codes.
We also propose the training techniques for \ARmodel\ to prevent the exposure bias~\cite{ExposureBias} in the training of AR models.

\subsubsection{AR Modeling for Codes with Depth $D$}
After RQ-VAE extracts a code map $\bM \in [K]^{H\times W \times D}$, the raster scan order~\cite{PixelRNN} rearranges the spatial indices of $\bM$ to a 2D array of codes $\mathbf{S} \in [K]^{T \times D}$ where $T = HW$.
That is, $\mathbf{S}_t$, which is a $t$-th row of $\mathbf{S}$, contains $D$ codes as
\begin{equation} \label{eq:D_stack_codes}
    \bS_t = (\bS_{t1}, \cdots, \bS_{tD}) \in [K]^D \quad \text{for $t \in [T]$.}
\end{equation}

Regarding $\bS$ as discrete latent variables of an image, AR models learn $p(\bS)$ which is autoregressively factorized as
\begin{equation}
    p(\bS) = \prod_{t=1}^T \prod_{d=1}^D p(\bS_{td} \given \bS_{<t,d}, \bS_{t,<d}).
\end{equation}

\subsubsection{\ARmodel\ Architecture} \label{sec:rqt_arch}
A na\"ive approach can unfold $\bS$ into a sequence of length $T D$ using the raster-scan order and feed it to the conventional transformer~\cite{transformer}.
However, it neither leverages the reduced length of $T$ by RQ-VAE and nor reduces the computational costs.
Thus, we propose \ARmodel\ to efficiently learn the codes extracted by RQ-VAE with depth $D$.
As shown in Figure~\ref{fig:SQ_overview}, \ARmodel\ consists of \emph{spatial transformer} and \emph{depth transformer}.

\paragraph{Spatial Transformer}
The spatial transformer is a stack of \emph{masked} self-attention blocks to extract a context vector that summarizes the information in previous positions.
For the input of the spatial transformer, we reuse the learned codebook of RQ-VAE with depth $D$.
Specifically, we define the input $\vecu_t$ of the spatial transformer as
\begin{equation}
    \vecu_t = \mathrm{PE}_T(t) + \sum_{d=1}^D \vece(\bS_{t-1,d}) \quad \text{for $t > 1$},
\end{equation}
where $\mathrm{PE}_T(t)$ is a positional embedding for spatial position $t$ in the raster-scan order. Note that the second term is equal to the quantized feature vector of an image in Eq.~\ref{eq:depth_sum}.
For the input at the first position, we define $\vecu_1$ as a learnable embedding, which is regarded as the start of a sequence.
After the sequence $(\vecu_t)_{t=1}^T$ is processed by the spatial transformer, a context vector $\vech_{t}$ encodes all information of $\bS_{<t}$ as
\begin{equation}
    \vech_{t} = \mathrm{SpatialTransformer}(\vecu_{1}, \cdots, \vecu_{t}).
\end{equation}

\paragraph{Depth Transformer}
Given the context vector $\vech_{t}$, the depth transformer autoregressively predicts $D$ codes $(\bS_{t1}, \cdots, \bS_{tD})$ at position $t$.
At position $t$ and depth $d$, the input $\vecv_{td}$ of the depth transformer is defined as the sum of the code embeddings of up to depth $d-1$ such that
\begin{equation} \label{eq:v_td}
    \vecv_{td} = \mathrm{PE}_D(d)+\sum_{d'=1}^{d-1} \vece(\bS_{td'}) \quad \text{for $d > 1$,}
\end{equation}
where $\mathrm{PE}_D(d)$ is a positional embedding for depth $d$ and shared at every position $t$.
We do not use $PE_T(t)$ in $\vecv_{td}$, since the positional information is already encoded in $\vecu_t$.
For $d=1$, we use $\vecv_{t1} = \mathrm{PE}_D(1)+\vech_t$.
Note that the second term in Eq.~\ref{eq:v_td} corresponds to a quantized feature vector $\hat{\bZ}_{hw}^{(d-1)}$ at depth $d-1$ in Eq.~\ref{eq:depth_sum}.
Thus, the depth transformer predicts the next code for a finer estimation of $\hat{\bZ}_t$ based on the previous estimations up to $d-1$.
Finally, the depth transformer predicts the conditional distribution $\mathbf{p}_{td}(k)=p(\bS_{td}=k|\bS_{<t,d}, \bS_{t,<d})$ as
\begin{equation}
    \mathbf{p}_{td} = \mathrm{DepthTransformer}
    (\vecv_{t1}, \cdots, \vecv_{td}).
\end{equation}
\ARmodel\ is trained to minimize $\cL_{AR}$, which is the negative log-likelihood (NLL) loss:
\begin{equation} \label{eq:nll_loss}
    \cL_{AR} = \EE_{\bS} \EE_{t,d} \left[
        -\log p(\bS_{td}|\bS_{<t,d}, \bS_{t,<d})
    \right].
\end{equation}

\paragraph{Computational Complexity}

Our \ARmodel\ can efficiently learn and predict the $T\!\times\!D$ code maps of RQ-VAE, since \ARmodel\ has lower computational complexity than the na\"ive approach, which uses the unfolded 1D sequence of $TD$ codes.
When computing $TD$ length of sequences, a transformer with $N$ layers has $O(NT^2D^2)$ of computational complexity~\cite{transformer}.
On the other hand, let us consider a \ARmodel\ with total $N$ layers, where the number of layers in the spatial transformer and depth transformer is $\Nsp$ and $\Ndep$, respectively.
Then, the spatial transformer requires $O(\Nsp T^2)$ and the depth transformer requires $O(\Ndep TD^2)$, since the maximum sequence lengths for the spatial transformer and depth transformer are $T$ and $D$, respectively.
Hence, the computational complexity of \ARmodel\ is $O(\Nsp T^2 + \Ndep TD^2)$, which is much less than $O(NT^2D^2)$.
In Section~\ref{sec:exp_efficiency}, we show that our \ARmodel\ has a faster speed of image generation than previous AR models.


\begin{figure*}
    \centering
    \includegraphics[height=0.5\textwidth, width=\textwidth]{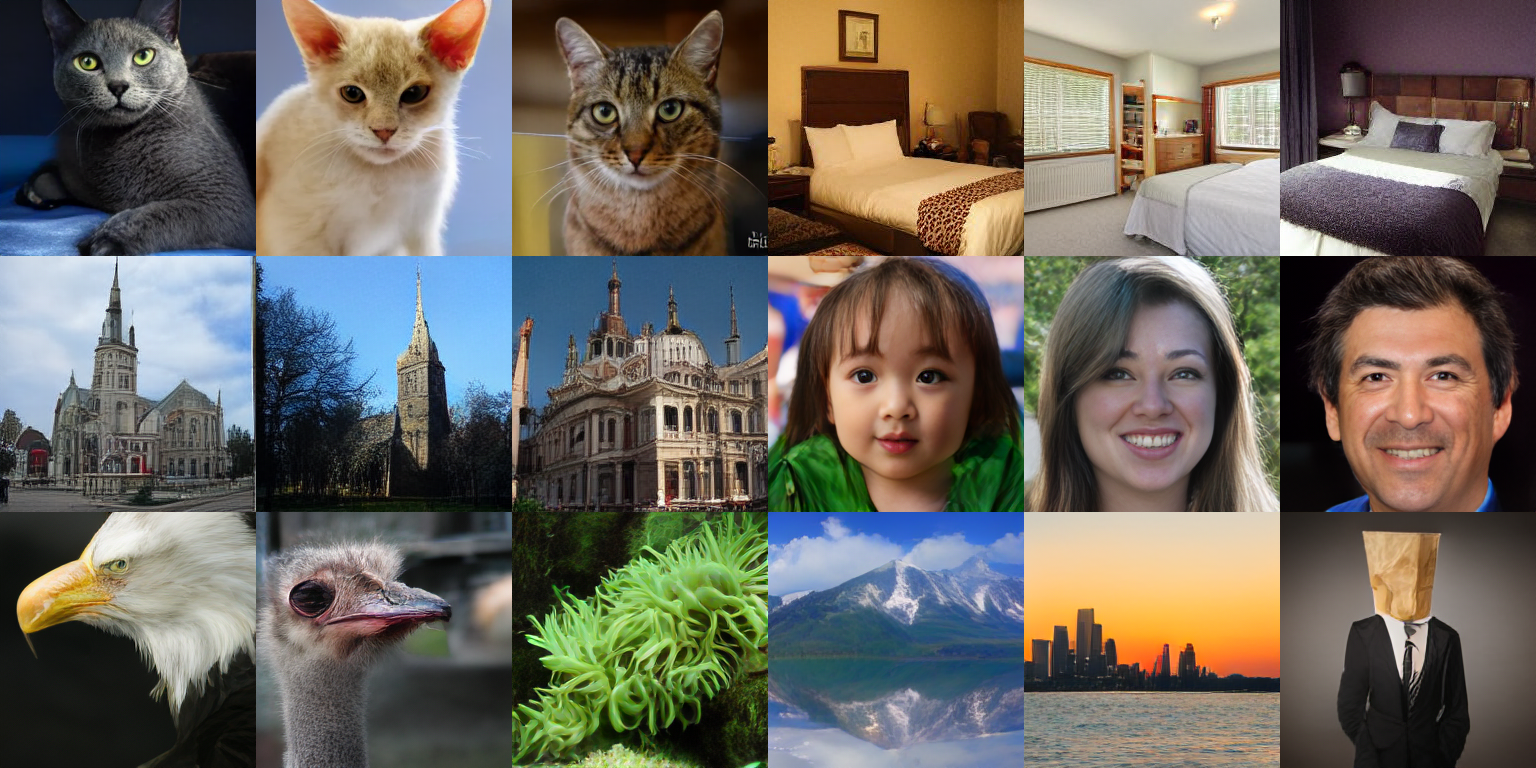}
    \caption{Generated 256$\times$256 images by our models. First row: LSUN-$\{\text{cat, bedroom}\}$. Second row: LSUN-church and FFHQ. Third row: ImageNet and CC-3M. The text conditions of CC-3M are ``Mountains and hills reflecting over a surface," ``A sunset over the skyline of a city," and ``Businessman with a paper bag on head," respectively.}
    \label{fig:Sampling}
\end{figure*}

\subsubsection{Soft Labeling and Stochastic Sampling}
\newcommand{\soft}{\text{soft}}
The exposure bias~\cite{ExposureBias} is known to deteriorate the performance of an AR model due to the error accumulation from the discrepancy of predictions in training and inference.
During an inference of \ARmodel, the prediction errors can also accumulate along with the depth $D$, since finer estimation of the feature vector becomes harder as $d$ increases.

Thus, we propose \emph{soft labeling} and \emph{stochastic sampling} of codes from RQ-VAE to resolve the exposure bias.
Scheduled sampling~\cite{bengio2015scheduled} is a way to reduce the discrepancy.
However, it is unsuitable for a large-scale AR model, since multiple inferences are required at each training step and increase the training cost.
Instead, we leverage the geometric relationship of code embeddings in RQ-VAE.
We define a categorical distribution on $[K]$ conditioned by a vector $\vecz \in \RR^{n_z}$ as $\Q_\tau(k|\vecz)$, where $\tau > 0$ is a temperature 
\begin{equation} \label{eq:code_distribution}
    \Q_\tau(k \given \vecz) \propto e^{-\|\vecz - \vece(k)\|_2^2 / \tau} \quad \text{for $k \in [K]$}.
\end{equation}
As $\tau$ approaches zero, $\Q_\tau$ is sharpened and converges to the one-hot distribution $\Q_0(k \given \vecz) = \mathbf{1}[k = \Q(\vecz; \cC)]$.

\paragraph{Soft Labeling of Target Codes}
Based on the distance between code embeddings, soft labeling is used to improve the training of \ARmodel\ by explicit supervision on the geometric relationship between the codes in RQ-VAE.
For a position $t$ and a depth $d$, let $\bZ_t$ be a feature vector of an image and $\vecr_{t,d-1}$ be a residual vector at depth $d-1$ in Eq.~\ref{eq:SQ_details}.
Then, the NLL loss in Eq.~\ref{eq:nll_loss} uses the one-hot label $\Q_{0}(\cdot \given \vecr_{t,d-1})$ as the supervision of $\bS_{td}$.
Instead of the one-hot labels, we use the softened distribution $\Q_\tau(\cdot \given \vecr_{t,d-1})$ as the supervision.

\paragraph{Stochastic Sampling for Codes of RQ-VAE}
Along with the soft labeling above, we propose stochastic sampling of the code map from RQ-VAE to reduce the discrepancy in training and inference.
Instead of the deterministic code selection of RQ in Eq.~\ref{eq:SQ_details}, we select the code $\bS_{td}$ by sampling from $\Q_{\tau}(\cdot | \vecr_{t,d-1})$. 
Note that our stochastic sampling is equivalent to the original code selection of SQ in the limit of $\tau \to 0$. 
The stochastic sampling provides different compositions of codes $\bS$ for a given feature map of an image.

\section{Experiments} \label{sec:Experiments}
In this section, we empirically validate our model for high-quality image generation.
We evaluate our model on unconditional image generation benchmarks in Section~\ref{sec:exp_uncond} and conditional image generation in Section~\ref{sec:exp_cond}. 
The computational efficiency of RQ-Transformer is shown in Section~\ref{sec:exp_efficiency}.
We also conduct an ablation study to understand the effectiveness of RQ-VAE in Section~\ref{sec:exp_abl}.

For a fair comparison, we adopt the model architecture of VQ-GAN~\cite{VQGAN}.
However, since RQ-VAEs convert 256$\times$256$\times$3 RGB images into 8$\times$8$\times$4 codes, we add an encoder and decode block to RQ-VAE and further decreases the resolution of the feature map by half.
All RQ-Transformers have $\Nsp=24$ and $\Ndep=4$ except for the model of 1.4B parameters that has $\Nsp=42$ and $\Ndep=6$.
We include all details of implementation in the supplementary material.

\begin{table} 
\centering
\small
\caption{Comparison of FIDs for unconditional image generation on LSUN-$\{\text{Cat, Bedroom, Church}\}$~\cite{yu2015lsun} and FFHQ~\cite{karras2019style}.}
\label{tab:result_uncond}
\begin{tabular}{l|cccc}
\toprule
    & Cat & Bedroom & Church & FFHQ \\ \hline
VDVAE~\cite{VDVAE} & - & - & - & 28.5 \\ \hline
DDPM~\cite{ho2020denoising} & 19.75 & 4.90 & 7.89 & - \\ 
ImageBART~\cite{ImageBART} & 15.09 & 5.51 & 7.32 & 9.57 \\ \hline
StyleGAN2~\cite{StyleGANv2} & 7.25 & 2.35 & 3.86 & 3.8 \\
BigGAN~\cite{BigGAN} & - & - & - & 12.4 \\ \hline
DCT~\cite{DCTransformer} & - & 6.40 & 7.56 & 13.06 \\
VQ-GAN~\cite{VQGAN} & 17.31 & 6.35 & 7.81 & 11.4 \\ \hline
\textbf{RQ-Transformer} & \textcolor{black}{8.64} & \textcolor{black}{3.04} & 7.45 & 10.38 \\ 
\bottomrule
\end{tabular}
\end{table}

\subsection{Unconditional Image Generation} \label{sec:exp_uncond}
The quality of unconditional image generation is evaluated on the LSUN-$\{ \text{cat, bedroom, church} \}$~\cite{yu2015lsun} and FFHQ~\cite{karras2019style} datasets.
The codebook size $K$ is 2048 for FFHQ and 16384 for LSUN.
For the FFHQ dataset, RQ-VAE is trained from scratch for \textcolor{black}{100 epochs}.
We also use early stopping for RQ-Transformer when the validation loss is minimized, since the small size of FFHQ leads to overfitting of AR models.
For the LSUN datasets, we use a pretrained RQ-VAE on ImageNet and finetune the model for one epoch on each dataset.
Considering the dataset size, we use RQ-Transformer of 612M parameters for LSUN-$\{\text{cat, bedroom}\}$ and 370M parameters for LSUN-church and FFHQ.
For the evaluation measure, we use Frechet Inception Distance (FID)~\cite{heusel2017gans} between 50K generated samples and all training samples.
Following the previous studies~\cite{VQGAN,ImageBART}, we also use top-$k$ and top-$p$ sampling to report the best performance.

Table~\ref{tab:result_uncond} shows that our model outperforms the other AR models on unconditional image generation.
For small-scale datasets such as LSUN-church and FFHQ, our model outperforms DCT~\cite{DCTransformer} and VQ-GAN~\cite{VQGAN} with marginal improvements.
However, for a larger scale of datasets such as LSUN-$\{ \text{cat, bedroom} \}$, our model significantly outperforms other AR models and diffusion-based models~\cite{ho2020denoising,ImageBART}.
We conjecture that the performance improvement comes from the shorter sequence length by RQ-VAE, since SQ-Transformer can easily learn the long-range interactions between codes in the short length of the sequence.
In the first two rows of Figure~\ref{fig:Sampling}, we show that RQ-Transformer can unconditionally generate high-quality images.

\begin{table} 
\centering
\small
\caption{Comparison of FIDs and ISs for class-conditioned image generation on ImageNet~\cite{deng2009imagenet} 256$\times$256. $\dagger$ denotes a model without our stochastic sampling and soft labeling. $\ddagger$ denotes the use of \textcolor{black}{rejection sampling or gradient guidance by pretrained classifier}. $\ast$ denotes the use of RQ-VAE trained for 50 epochs.}
\label{tab:result_cIN}
\begin{tabular}{l|c|cc}
\toprule 
  & Params & FID & IS \\ \hline
\rowcolor[gray]{0.95}\multicolumn{4}{l}{\textit{\textbf{without rejection sampling or gradient guidance}}} \\ \hline
\textcolor{black}{ADM}~\cite{ADM} & 554M & 10.94 & 101.0 \\ 
ImageBART~\cite{ImageBART} & 3.5B & 21.19 & 61.6\\ \hline
BigGAN~\cite{BigGAN} & 164M & 7.53 & 168.6 \\
BigGAN-deep~\cite{BigGAN} & 112M & 6.84 & 203.6 \\ \hline
VQ-VAE2~\cite{VQVAE2} & 13.5B & $\sim$31 & $\sim$45 \\
DCT~\cite{DCTransformer} & 738M & 36.5 & n/a \\
VQ-GAN~\cite{VQGAN} & 1.4B & 15.78 & 74.3 \\ \hline
\textbf{RQ-Transformer} & 480M & 15.72 & 86.8\small{$\pm$1.4} \\
\textbf{RQ-Transformer}$^\dagger$ & 821M & 14.06 & 95.8\small{$\pm$2.1}  \\
\textbf{RQ-Transformer}  & 821M & 13.11 & 104.3\small{$\pm$1.5}  \\
\textbf{RQ-Transformer} & 1.4B & 11.56 & 112.4\small{$\pm$1.1} \\
\textbf{RQ-Transformer$^\ast$} & 1.4B & 8.71 & 119.0\small{$\pm$2.5} \\ 
\textbf{RQ-Transformer$^\ast$} & 3.8B & 7.55 & 134.0\small{$\pm$3.0} \\ \hline
\rowcolor[gray]{0.95}\multicolumn{4}{l}{\textit{\textbf{with rejection sampling or gradient guidance}}} \\ \hline
ADM$^\ddagger$~\cite{ADM} & 554M & 4.59 & 186.7 \\ 
ImageBART$^\ddagger$~\cite{ImageBART} & 3.5B & 7.44 & 273.5\small{$\pm$4.1} \\
VQ-VAE2$^\ddagger$~\cite{VQVAE2} & 13.5B & $\sim$10 & $\sim$330 \\
VQ-GAN$^\ddagger$~\cite{VQGAN} & 1.4B & 5.20 & 280.3\small{$\pm$5.5} \\ \hline
\textbf{RQ-Transformer}$^\ddagger$ & 1.4B & 4.45 & 326.0\small{$\pm$3.5} \\
\textbf{RQ-Transformer$^\ast$}$^\ddagger$ & 1.4B & 3.89 & 337.5\small{$\pm$4.6} \\ 
\textbf{RQ-Transformer$^\ast$}$^\ddagger$ & 3.8B & 3.80 & 323.7\small{$\pm$2.8} \\ \hline
Validation Data & - & 1.62 & 234.0 \\ 
\bottomrule
\end{tabular}
\end{table}

\begin{table} 
\centering
\small
\caption{Comparison of FID and CLIP score~\cite{CLIP} on the validation data of CC-3M~\cite{sharma-etal-2018-conceptual} for text-conditioned image generation.}
\label{tab:exp_cc}
\begin{tabular}{l|c|ccc}
\toprule 
     & Params & FID & CLIP-s \\ \hline
VQ-GAN~\cite{VQGAN} & 600M & 28.86 &  0.20 \\ 
ImageBART~\cite{ImageBART} & 2.8B & 22.61 & 0.23 \\ \hline
\textbf{RQ-Transformer} & 654M & 12.33 & 0.26  \\ 
\bottomrule
\end{tabular}
\end{table}

\begin{table} 
\centering
\small
\caption{Comparison of FIDs between ImageNet validation images and their reconstructed images according to codebook size ($K$) and the shape of code map $H\times W \times D$. $\dagger$ denotes the reproduced performance, and \textcolor{black}{$\ast$ denotes 50 epochs of training.}}
\label{tab:ablation}
\begin{tabular}{l|cc|c}
\toprule 
     & $H\times W \times D$ & $K$ & rFID \\ \hline
VQ-GAN~\cite{VQGAN} & 16$\times$16$\times$1 & 16,384 & 4.90 \\ 
VQ-GAN$^\dagger$ & 16$\times$16$\times$1 & 16,384 & 4.32 \\\hline
VQ-GAN & 8$\times$8$\times$1 & 16,384 & 17.95 \\
VQ-GAN & 8$\times$8$\times$1 & 65,536 & 17.66 \\ 
VQ-GAN & 8$\times$8$\times$1 & 131,072 & 17.09 \\ \hline
RQ-VAE & 8$\times$8$\times$2 & 16,384 & 10.77 \\
RQ-VAE & 8$\times$8$\times$4 & 16,384 & 4.73 \\ 
\textcolor{black}{RQ-VAE}$^\ast$ & 8$\times$8$\times$4 & 16,384 & 3.20 \\
\textcolor{black}{RQ-VAE}& 8$\times$8$\times$8 & 16,384 & 2.69 \\ 
\textcolor{black}{RQ-VAE} & 8$\times$8$\times$16 & 16,384 & 1.83 \\ 
\bottomrule
\end{tabular}
\end{table}

\subsection{Conditional Image Generation} \label{sec:exp_cond}
We use ImageNet~\cite{deng2009imagenet} and CC-3M~\cite{sharma-etal-2018-conceptual} for a class- and text-conditioned image generation, respectively.
We train RQ-VAE with $K${$=$}16,384 on ImageNet training data for 10 epochs and reuse the trained RQ-VAE for CC-3M.
\textcolor{black}{
For ImageNet, we also use RQ-VAE trained for 50 epochs to examine the effect of improved reconstruction quality on image generation quality of RQ-Transformer in Table~\ref{tab:result_cIN}.
}
For conditioning, we append the embeddings of class and text conditions to the start of input for spatial transformer.
The texts of CC-3M are represented as a sequence of at most 32 tokens using a byte pair encoding~\cite{sennrich2016neural,huggingface}.

Table~\ref{tab:result_cIN} shows that our model significantly outperforms previous models on ImageNet.
\textcolor{black}{Our RQ-Transformer of 480M parameters is competitive with the previous AR models including VQ-VAE2~\cite{VQVAE2}, DCT~\cite{DCTransformer}, and VQ-GAN~\cite{VQGAN} without rejection sampling, although our model has 3$\times$ less parameters than VQ-GAN.}
In addition, RQ-Transformer of 821M parameters outperforms the previous AR models without rejection sampling.
Our stochastic sampling is also effective for performance improvement, while RQ-Transformer without it still outperforms other AR models.
RQ-Transformer of 1.4B parameters achieves 11.56 of FID score without rejection sampling.
\textcolor{black}{
When we increase the training epoch of RQ-VAE from 10 into 50 and improve the reconstruction quality, RQ-Transformer of 1.4B parameters further improves the performance and achieves 8.71 of FID.
Moreover, when we further increase the number of parameters to 3.8B, RQ-Transformer achieves 7.55 of FID score without rejection sampling and is competitive with BigGAN~\cite{BigGAN}.
}
When ResNet-101~\cite{he2016deep} is used for rejection sampling with 5\% and 12.5\% of acceptance rates for 1.4B and 3.8B parameters, respectively, our model outperforms ADM~\cite{ADM} and achieves the state-of-the-art score of FID. 
Figure~\ref{fig:Sampling} also shows that our model can generate high-quality images.

RQ-Transformer can also generate high-quality images based on various text conditions of CC-3M.
RQ-Transformer shows significantly higher performance than VQ-GAN with a similar number of parameters.
In addition, although RQ-Transformer has 23\% of parameters, our model significantly outperforms ImageBART~\cite{ImageBART} on both FID and CLIP score~\cite{CLIP} (with ViT-B/32~\cite{ViT}).
The results imply that RQ-Transformer can easily learn the relationship between a text and an image when the reduced sequence length is used for the image.
Figure~\ref{fig:Sampling} shows that RQ-Transformer trained on CC-3M can generate high-quality images using various text conditions.
In addition, the text conditions in Figure~\ref{fig:teaser} are novel compositions of visual concepts, which are unseen in training.

\begin{figure}
    \centering
    \includegraphics[height=1.5in]{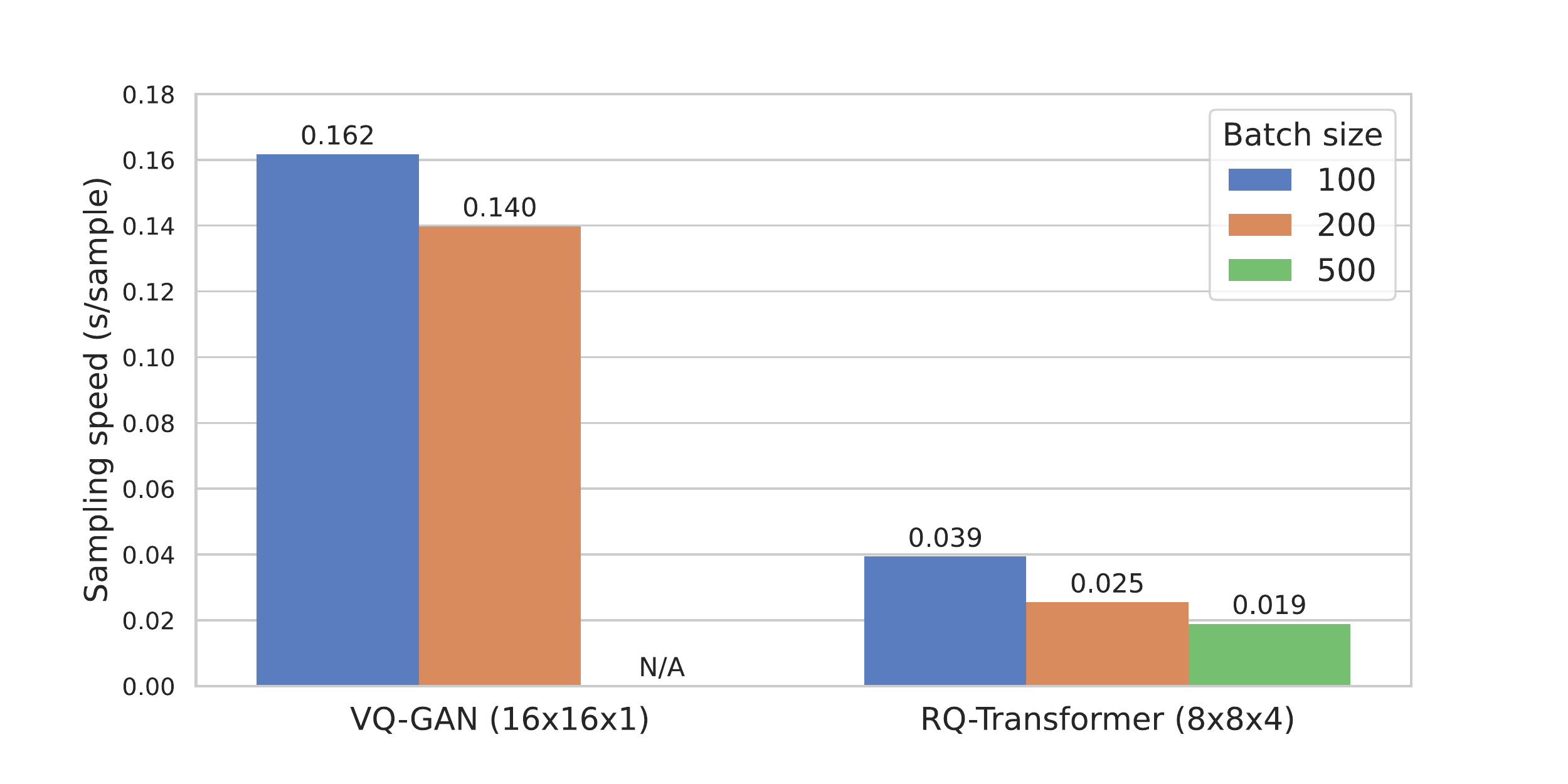}
    \caption{The sampling speed of RQ-Transformer with 1.4B parameters according to batch size and code map shape.}
    \label{fig:efficiency}
\end{figure}

\begin{figure*}
    \centering
    \includegraphics[width=\textwidth]{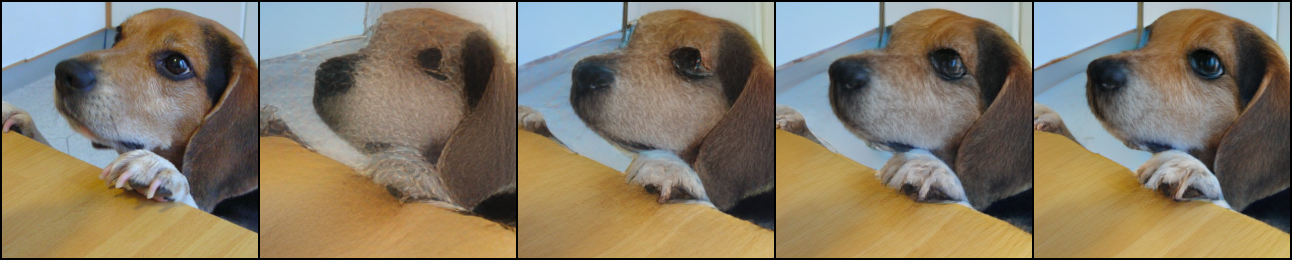}
    \caption{The examples of coarse-to-fine approximation by RQ-VAE. The first example is the original image, and the others are reconstructed from $\hat{\bZ}^{( d)}$. As $d$ increases, the reconstructed images become clear and include fine-grained details of the original image.}
    \label{fig:additive_decoding}
\end{figure*}

\begin{figure}
    \centering
    \includegraphics[height=1.7in]{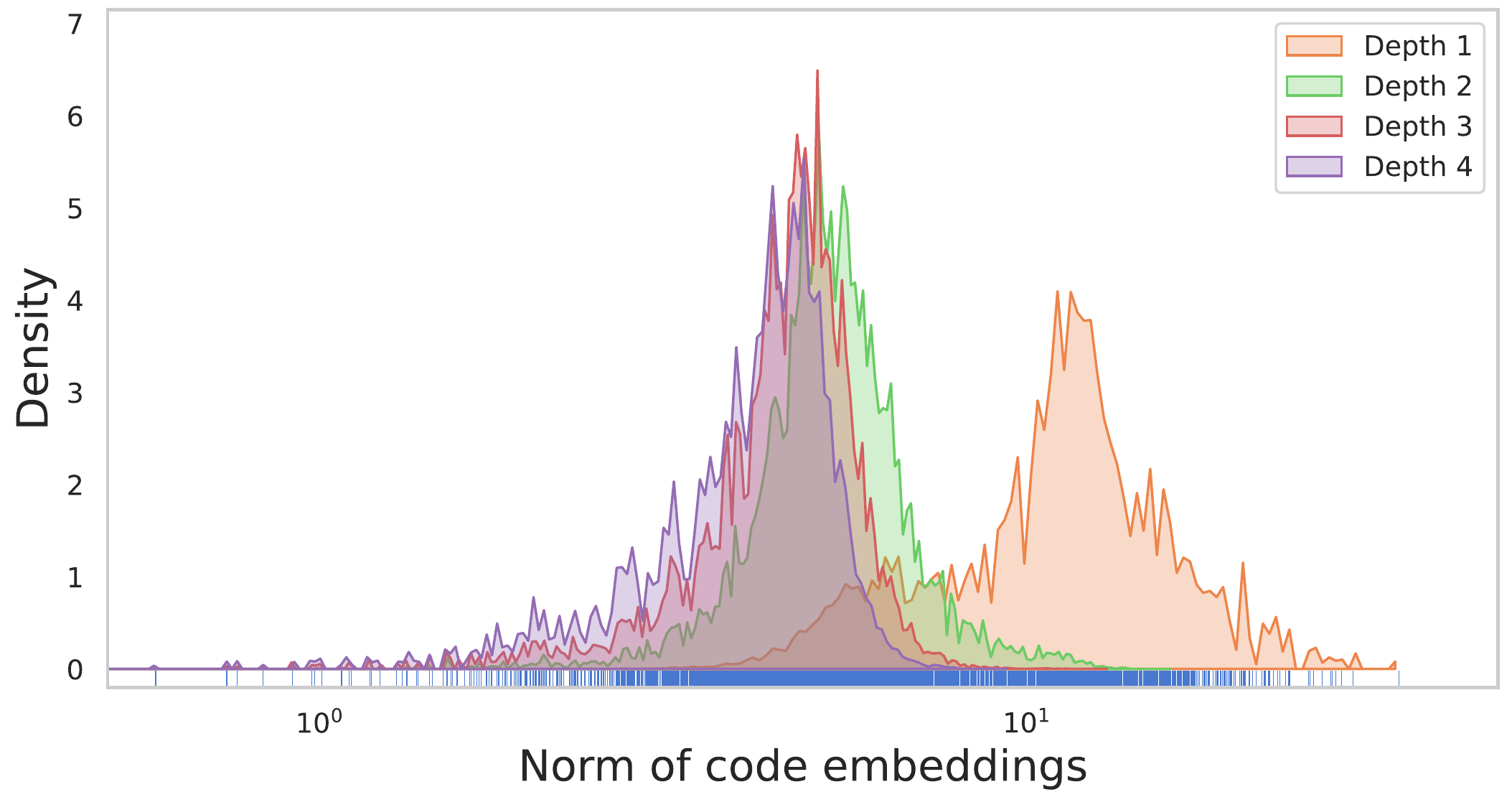}
    \caption{The distribution of used codes at each quantization depth. The blue bar plot represents the code distribution according to the norm of embeddings. ImageNet validation data is used.}
    \label{fig:codebook_usage}
\end{figure}

\subsection{Computational Efficiency of RQ-Transformer} \label{sec:exp_efficiency}

In Figure~\ref{fig:efficiency}, we evaluate the sampling speed of RQ-Transformer and make a comparison with VQ-GAN.
Both the models have 1.4B parameters. The shape of the input code map for VQ-GAN and RQ-Transformer are set to be 16$\times$16$\times$1 and 8$\times$8$\times$4, respectively.
We use a single NVIDIA A100 GPU for each model to generate 5000 samples with 100, 200, and 500 of batch size.
The reported speeds in Figure~\ref{fig:efficiency} do not include the decoding time of the stage 1 model to focus on the effect of RQ-Transformer architecture.
The decoding time of VQ-GAN and RQ-VAE is about 0.008 sec/image.

For the batch size of 100 and 200, RQ-Transformer shows 4.1$\times$ and 5.6$\times$ speed-up compared with VQ-GAN.
Moreover, thanks to the memory saving from the short sequence length of RQ-VAE, RQ-Transformer can increase the batch size up to 500, which is not allowed for VQ-GAN.
Thus, RQ-Transformer can further accelerate the sampling speed, which is $0.02$ seconds per image, and be 7.3$\times$ faster than VQ-GAN with batch size 200.
Thus, RQ-Transformer is more computationally efficient than previous AR models, while achieving state-of-the-art results on high-resolution image generation benchmarks.


\subsection{Ablation Study on RQ-VAE} \label{sec:exp_abl}
We conduct an ablation study to understand the effect of RQ with respect to the codebook size ($K$) and the shape of the code map ($H \times W \times D$).
We measure the rFID, which is FID between original images and reconstructed images, on ImageNet validation data.
Table~\ref{tab:ablation} shows that increasing the quantization depth $D$ is more effective to improve the reconstruction quality than increasing the codebook size $K$.
Here, we remark that RQ-VAE with $D\!=\!1$ is equivalent to VQ-GAN.
For a fixed codebook size $K${$=$}16,384, the rFID significantly deteriorates as the spatial resolution $H \times W$ is reduced from 16$\times$16 to 8$\times$8.
Even when the codebook size is increased to $K${$=$}131,072, the rFID cannot recover the rFID with 16$\times$16 feature maps, since the restoration of rFID requires the codebook of size $K${$=$}16,384$^4$ in terms of the rate-distortion trade-off.
Contrastively, note that the rFIDs are significantly improved when we increase the quantization depth $D$ with a codebook of fixed size $K${$=$}16,384.
\textcolor{black}{
Thus, our RQ-VAE can further reduce the spatial resolution than VQ-GAN, while conserving the reconstruction quality.
Although RQ-VAE with $D>4$ can further improve the reconstruction quality, we use RQ-VAE with 8$\times$8$\times$4 code map for AR modeling of images, considering the computational costs of RQ-Transformer.
In addition, the longer training of RQ-VAE can further improve the reconstruction quality, but we train RQ-VAE for 10 epochs as the default due to its increased training time.
}


Figure~\ref{fig:additive_decoding} and~\ref{fig:codebook_usage} substantiate our claim that RQ-VAE conducts the coarse-to-fine estimation of feature maps.
For example, Figure~\ref{fig:additive_decoding} shows the reconstructed images $G(\hat{\bZ}^{( d)})$ of a quantized feature map at depth $d$ in Eq.~\ref{eq:SQ_details}.
When we only use the codes at $d=1$, the reconstructed image is blurry and only contains coarse information of the original image.
However, as $d$ increases and the information of remaining codes is sequentially added, the reconstructed image includes more clear and fine-grained details of the image.
We visualize the distribution of the code usage at each depth $d$ over the norm of code embeddings in Figure~\ref{fig:codebook_usage}.
Since RQ conducts the coarse-to-fine approximation of a feature map, a smaller norm of code embeddings are used as $d$ increases.
Moreover, the overlaps between the code usage distributions show that many codes are shared in different levels of depth $d$.
Thus, the shared codebook of RQ-VAE can maximize the utility of its codes.

\section{Conclusion}
Discrete representation of visual images is important for an AR model to generate high-resolution images.
In this work, we have proposed RQ-VAE and RQ-Transformer for high-quality image generation.
Under a fixed codebook size, RQ-VAE can precisely approximate a feature map of an image to represent the image as a short sequence of codes.
Thus, RQ-Transformer effectively learns to predict the codes to generate high-quality images with low computational costs.
Consequently, our approach outperforms the previous AR models on various image generation benchmarks such as LSUNs, FFHQ, ImageNet, and CC-3M.

Our study has \textcolor{black}{three} main limitations.
First, our model does not outperform StyleGAN2~\cite{StyleGANv2} on unconditional image generation, especially with a small-scale dataset such as FFHQ, due to overfitting of AR models.
Thus, regularizing AR models is worth exploration for high-resolution image generation on a small dataset.
Second, our study does not enlarge the model and training data for text-to-image generation.
As a previous study~\cite{DALL-E,henighan2020scaling} shows that a huge transformer can effectively learn the zero-shot text-to-image generation, increasing the number of parameters is an interesting future work.
\textcolor{black}{Third, AR models can only capture unidirectional contexts to generate images compared to other generative models. Thus, modeling of bidirectional contexts can further improve the quality of image generation and enable AR models to be used for image manipulation such as image inpainting and outpainting~\cite{ImageBART}.}

\textcolor{black}{
Although our study significantly reduces the computational costs for AR modeling of images, training of large-scale AR models is still expensive, consumes high amounts of electrical energy, and can leave a huge carbon footprint, as the scale of model and training dataset becomes large.
Thus, efficient training of large-scale AR models is still worth exploration to avoid environmental pollution.
}

\section{Acknowledgements}
This work was supported by Institute of Information \& communications Technology Planning \& Evaluation(IITP) grant funded by the Korea government(MSIT) (No.2018-0-01398: Development of a Conversational, Self-tuning DBMS; No.2021-0-00537: Visual Common Sense).

{\small
\bibliographystyle{ieee_fullname}
\bibliography{egbib}

\begin{thebibliography}{10}\itemsep=-1pt

\bibitem{AQ}
Artem Babenko and Victor Lempitsky.
\newblock Additive quantization for extreme vector compression.
\newblock In {\em Proceedings of the IEEE Conference on Computer Vision and
  Pattern Recognition}, pages 931--938, 2014.

\bibitem{bengio2015scheduled}
Samy Bengio, Oriol Vinyals, Navdeep Jaitly, and Noam Shazeer.
\newblock Scheduled sampling for sequence prediction with recurrent neural
  networks.
\newblock {\em arXiv preprint arXiv:1506.03099}, 2015.

\bibitem{BigGAN}
Andrew Brock, Jeff Donahue, and Karen Simonyan.
\newblock Large scale {GAN} training for high fidelity natural image synthesis.
\newblock In {\em International Conference on Learning Representations}, 2019.

\bibitem{GPT3}
Tom Brown, Benjamin Mann, Nick Ryder, Melanie Subbiah, Jared~D Kaplan, Prafulla
  Dhariwal, Arvind Neelakantan, Pranav Shyam, Girish Sastry, Amanda Askell,
  Sandhini Agarwal, Ariel Herbert-Voss, Gretchen Krueger, Tom Henighan, Rewon
  Child, Aditya Ramesh, Daniel Ziegler, Jeffrey Wu, Clemens Winter, Chris
  Hesse, Mark Chen, Eric Sigler, Mateusz Litwin, Scott Gray, Benjamin Chess,
  Jack Clark, Christopher Berner, Sam McCandlish, Alec Radford, Ilya Sutskever,
  and Dario Amodei.
\newblock Language models are few-shot learners.
\newblock In H. Larochelle, M. Ranzato, R. Hadsell, M.~F. Balcan, and H. Lin,
  editors, {\em Advances in Neural Information Processing Systems}, volume~33,
  pages 1877--1901. Curran Associates, Inc., 2020.

\bibitem{sklearn_api}
Lars Buitinck, Gilles Louppe, Mathieu Blondel, Fabian Pedregosa, Andreas
  Mueller, Olivier Grisel, Vlad Niculae, Peter Prettenhofer, Alexandre
  Gramfort, Jaques Grobler, Robert Layton, Jake VanderPlas, Arnaud Joly, Brian
  Holt, and Ga{\"{e}}l Varoquaux.
\newblock {API} design for machine learning software: experiences from the
  scikit-learn project.
\newblock In {\em ECML PKDD Workshop: Languages for Data Mining and Machine
  Learning}, pages 108--122, 2013.

\bibitem{iGPT}
Mark Chen, Alec Radford, Rewon Child, Jeffrey Wu, Heewoo Jun, David Luan, and
  Ilya Sutskever.
\newblock Generative pretraining from pixels.
\newblock In {\em International Conference on Machine Learning}, pages
  1691--1703. PMLR, 2020.

\bibitem{VDVAE}
Rewon Child.
\newblock Very deep vaes generalize autoregressive models and can outperform
  them on images.
\newblock In {\em International Conference on Learning Representations}, 2020.

\bibitem{cooper1990computational}
Gregory~F Cooper.
\newblock The computational complexity of probabilistic inference using
  bayesian belief networks.
\newblock {\em Artificial intelligence}, 42(2-3):393--405, 1990.

\bibitem{deng2009imagenet}
Jia Deng, Wei Dong, Richard Socher, Li-Jia Li, Kai Li, and Li Fei-Fei.
\newblock Imagenet: A large-scale hierarchical image database.
\newblock In {\em 2009 IEEE conference on computer vision and pattern
  recognition}, pages 248--255. Ieee, 2009.

\bibitem{Jukebox}
Prafulla Dhariwal, Heewoo Jun, Christine Payne, Jong~Wook Kim, Alec Radford,
  and Ilya Sutskever.
\newblock Jukebox: A generative model for music.
\newblock {\em arXiv preprint arXiv:2005.00341}, 2020.

\bibitem{ADM}
Prafulla Dhariwal and Alex Nichol.
\newblock Diffusion models beat gans on image synthesis.
\newblock {\em arXiv preprint arXiv:2105.05233}, 2021.

\bibitem{ViT}
Alexey Dosovitskiy, Lucas Beyer, Alexander Kolesnikov, Dirk Weissenborn,
  Xiaohua Zhai, Thomas Unterthiner, Mostafa Dehghani, Matthias Minderer, Georg
  Heigold, Sylvain Gelly, et~al.
\newblock An image is worth 16x16 words: Transformers for image recognition at
  scale.
\newblock {\em arXiv preprint arXiv:2010.11929}, 2020.

\bibitem{ImageBART}
Patrick Esser, Robin Rombach, Andreas Blattmann, and Björn Ommer.
\newblock Imagebart: Bidirectional context with multinomial diffusion for
  autoregressive image synthesis, 2021.

\bibitem{VQGAN}
Patrick Esser, Robin Rombach, and Bjorn Ommer.
\newblock Taming transformers for high-resolution image synthesis.
\newblock In {\em Proceedings of the IEEE/CVF Conference on Computer Vision and
  Pattern Recognition}, pages 12873--12883, 2021.

\bibitem{ferdowsi2017regularized}
Sohrab Ferdowsi, Slava Voloshynovskiy, and Dimche Kostadinov.
\newblock Regularized residual quantization: a multi-layer sparse dictionary
  learning approach.
\newblock {\em arXiv preprint arXiv:1705.00522}, 2017.

\bibitem{GAN}
Ian Goodfellow, Jean Pouget-Abadie, Mehdi Mirza, Bing Xu, David Warde-Farley,
  Sherjil Ozair, Aaron Courville, and Yoshua Bengio.
\newblock Generative adversarial nets.
\newblock {\em Advances in neural information processing systems}, 27, 2014.

\bibitem{PQ}
R. Gray.
\newblock Vector quantization.
\newblock {\em IEEE ASSP Magazine}, 1(2):4--29, 1984.

\bibitem{he2016deep}
Kaiming He, Xiangyu Zhang, Shaoqing Ren, and Jian Sun.
\newblock Deep residual learning for image recognition.
\newblock In {\em Proceedings of the IEEE conference on computer vision and
  pattern recognition}, pages 770--778, 2016.

\bibitem{henighan2020scaling}
Tom Henighan, Jared Kaplan, Mor Katz, Mark Chen, Christopher Hesse, Jacob
  Jackson, Heewoo Jun, Tom~B Brown, Prafulla Dhariwal, Scott Gray, et~al.
\newblock Scaling laws for autoregressive generative modeling.
\newblock {\em arXiv preprint arXiv:2010.14701}, 2020.

\bibitem{heusel2017gans}
Martin Heusel, Hubert Ramsauer, Thomas Unterthiner, Bernhard Nessler, and Sepp
  Hochreiter.
\newblock Gans trained by a two time-scale update rule converge to a local nash
  equilibrium.
\newblock {\em Advances in neural information processing systems}, 30, 2017.

\bibitem{ho2020denoising}
Jonathan Ho, Ajay Jain, and Pieter Abbeel.
\newblock Denoising diffusion probabilistic models.
\newblock {\em arXiv preprint arXiv:2006.11239}, 2020.

\bibitem{isola2017image}
Phillip Isola, Jun-Yan Zhu, Tinghui Zhou, and Alexei~A Efros.
\newblock Image-to-image translation with conditional adversarial networks.
\newblock In {\em Proceedings of the IEEE conference on computer vision and
  pattern recognition}, pages 1125--1134, 2017.

\bibitem{johnson2016perceptual}
Justin Johnson, Alexandre Alahi, and Li Fei-Fei.
\newblock Perceptual losses for real-time style transfer and super-resolution.
\newblock In {\em European conference on computer vision}, pages 694--711.
  Springer, 2016.

\bibitem{juang1982multiple}
Biing-Hwang Juang and A Gray.
\newblock Multiple stage vector quantization for speech coding.
\newblock In {\em ICASSP'82. IEEE International Conference on Acoustics,
  Speech, and Signal Processing}, volume~7, pages 597--600. IEEE, 1982.

\bibitem{karras2019style}
Tero Karras, Samuli Laine, and Timo Aila.
\newblock A style-based generator architecture for generative adversarial
  networks.
\newblock In {\em Proceedings of the IEEE/CVF Conference on Computer Vision and
  Pattern Recognition}, pages 4401--4410, 2019.

\bibitem{StyleGANv2}
Tero Karras, Samuli Laine, Miika Aittala, Janne Hellsten, Jaakko Lehtinen, and
  Timo Aila.
\newblock Analyzing and improving the image quality of stylegan.
\newblock In {\em Proceedings of the IEEE/CVF Conference on Computer Vision and
  Pattern Recognition}, pages 8110--8119, 2020.

\bibitem{Adam}
Diederik~P. Kingma and Jimmy Ba.
\newblock Adam: {A} method for stochastic optimization.
\newblock In Yoshua Bengio and Yann LeCun, editors, {\em 3rd International
  Conference on Learning Representations, {ICLR} 2015, San Diego, CA, USA, May
  7-9, 2015, Conference Track Proceedings}, 2015.

\bibitem{LedigC2017vggloss}
Christian Ledig, Lucas Theis, Ferenc Huszar, Jose Caballero, Andrew Cunningham,
  Alejandro Acosta, Andrew Aitken, Alykhan Tejani, Johannes Totz, Zehan Wang,
  and Wenzhe Shi.
\newblock Photo-realistic single image super-resolution using a generative
  adversarial network.
\newblock In {\em CVPR}, 2017.

\bibitem{li2021trq}
Yue Li, Wenrui Ding, Chunlei Liu, Baochang Zhang, and Guodong Guo.
\newblock Trq: Ternary neural networks with residual quantization.
\newblock In {\em Proceedings of the AAAI Conference on Artificial
  Intelligence}, volume~35, pages 8538--8546, 2021.

\bibitem{li2017performance}
Zefan Li, Bingbing Ni, Wenjun Zhang, Xiaokang Yang, and Wen Gao.
\newblock Performance guaranteed network acceleration via high-order residual
  quantization.
\newblock In {\em Proceedings of the IEEE international conference on computer
  vision}, pages 2584--2592, 2017.

\bibitem{AdamW}
Ilya Loshchilov and Frank Hutter.
\newblock Decoupled weight decay regularization.
\newblock In {\em International Conference on Learning Representations}, 2019.

\bibitem{SQ}
Julieta Martinez, Holger~H Hoos, and James~J Little.
\newblock Stacked quantizers for compositional vector compression.
\newblock {\em arXiv preprint arXiv:1411.2173}, 2014.

\bibitem{DCTransformer}
Charlie Nash, Jacob Menick, Sander Dieleman, and Peter Battaglia.
\newblock Generating images with sparse representations.
\newblock In Marina Meila and Tong Zhang, editors, {\em Proceedings of the 38th
  International Conference on Machine Learning}, volume 139 of {\em Proceedings
  of Machine Learning Research}, pages 7958--7968. PMLR, 18--24 Jul 2021.

\bibitem{PixelRNN}
Aaron~Van Oord, Nal Kalchbrenner, and Koray Kavukcuoglu.
\newblock Pixel recurrent neural networks.
\newblock In {\em Proceedings of The 33rd International Conference on Machine
  Learning}, 2016.

\bibitem{PixelCNN}
A{\"a}ron van~den Oord, Nal Kalchbrenner, Oriol Vinyals, Lasse Espeholt, Alex
  Graves, and Koray Kavukcuoglu.
\newblock Conditional image generation with pixelcnn decoders.
\newblock In {\em Proceedings of the 30th International Conference on Neural
  Information Processing Systems}, pages 4797--4805, 2016.

\bibitem{CLIP}
Alec Radford, Jong~Wook Kim, Chris Hallacy, Aditya Ramesh, Gabriel Goh,
  Sandhini Agarwal, Girish Sastry, Amanda Askell, Pamela Mishkin, Jack Clark,
  Gretchen Krueger, and Ilya Sutskever.
\newblock Learning transferable visual models from natural language
  supervision.
\newblock In Marina Meila and Tong Zhang, editors, {\em Proceedings of the 38th
  International Conference on Machine Learning}, volume 139 of {\em Proceedings
  of Machine Learning Research}, pages 8748--8763. PMLR, 18--24 Jul 2021.

\bibitem{DALL-E}
Aditya Ramesh, Mikhail Pavlov, Gabriel Goh, Scott Gray, Chelsea Voss, Alec
  Radford, Mark Chen, and Ilya Sutskever.
\newblock Zero-shot text-to-image generation.
\newblock In Marina Meila and Tong Zhang, editors, {\em Proceedings of the 38th
  International Conference on Machine Learning}, volume 139 of {\em Proceedings
  of Machine Learning Research}, pages 8821--8831. PMLR, 18--24 Jul 2021.

\bibitem{ExposureBias}
Marc'Aurelio Ranzato, Sumit Chopra, Michael Auli, and Wojciech Zaremba.
\newblock Sequence level training with recurrent neural networks.
\newblock In Yoshua Bengio and Yann LeCun, editors, {\em 4th International
  Conference on Learning Representations, {ICLR} 2016, San Juan, Puerto Rico,
  May 2-4, 2016, Conference Track Proceedings}, 2016.

\bibitem{VQVAE2}
Ali Razavi, Aaron van~den Oord, and Oriol Vinyals.
\newblock Generating diverse high-fidelity images with vq-vae-2.
\newblock In {\em Advances in neural information processing systems}, pages
  14866--14876, 2019.

\bibitem{PixelCNN++}
Tim Salimans, Andrej Karpathy, Xi Chen, and Diederik~P. Kingma.
\newblock Pixelcnn++: A pixelcnn implementation with discretized logistic
  mixture likelihood and other modifications.
\newblock In {\em ICLR}, 2017.

\bibitem{sennrich2016neural}
Rico Sennrich, Barry Haddow, and Alexandra Birch.
\newblock Neural machine translation of rare words with subword units.
\newblock In {\em Proceedings of the 54th Annual Meeting of the Association for
  Computational Linguistics (Volume 1: Long Papers)}, pages 1715--1725, 2016.

\bibitem{shannon1959coding}
Claude~E Shannon et~al.
\newblock Coding theorems for a discrete source with a fidelity criterion.
\newblock {\em IRE Nat. Conv. Rec}, 4(142-163):1, 1959.

\bibitem{sharma-etal-2018-conceptual}
Piyush Sharma, Nan Ding, Sebastian Goodman, and Radu Soricut.
\newblock Conceptual captions: A cleaned, hypernymed, image alt-text dataset
  for automatic image captioning.
\newblock In {\em Proceedings of the 56th Annual Meeting of the Association for
  Computational Linguistics (Volume 1: Long Papers)}, pages 2556--2565,
  Melbourne, Australia, July 2018. Association for Computational Linguistics.

\bibitem{VGG}
Karen Simonyan and Andrew Zisserman.
\newblock Very deep convolutional networks for large-scale image recognition.
\newblock In Yoshua Bengio and Yann LeCun, editors, {\em 3rd International
  Conference on Learning Representations, {ICLR} 2015, San Diego, CA, USA, May
  7-9, 2015, Conference Track Proceedings}, 2015.

\bibitem{VQVAE}
Aaron van~den Oord, Oriol Vinyals, and koray kavukcuoglu.
\newblock Neural discrete representation learning.
\newblock In I. Guyon, U.~V. Luxburg, S. Bengio, H. Wallach, R. Fergus, S.
  Vishwanathan, and R. Garnett, editors, {\em Advances in Neural Information
  Processing Systems}, volume~30. Curran Associates, Inc., 2017.

\bibitem{transformer}
Ashish Vaswani, Noam Shazeer, Niki Parmar, Jakob Uszkoreit, Llion Jones,
  Aidan~N Gomez, {\L}ukasz Kaiser, and Illia Polosukhin.
\newblock Attention is all you need.
\newblock In {\em Advances in neural information processing systems}, pages
  5998--6008, 2017.

\bibitem{huggingface}
Thomas Wolf, Lysandre Debut, Victor Sanh, Julien Chaumond, Clement Delangue,
  Anthony Moi, Pierric Cistac, Tim Rault, Rémi Louf, Morgan Funtowicz, Joe
  Davison, Sam Shleifer, Patrick von Platen, Clara Ma, Yacine Jernite, Julien
  Plu, Canwen Xu, Teven~Le Scao, Sylvain Gugger, Mariama Drame, Quentin Lhoest,
  and Alexander~M. Rush.
\newblock Transformers: State-of-the-art natural language processing.
\newblock In {\em Proceedings of the 2020 Conference on Empirical Methods in
  Natural Language Processing: System Demonstrations}, pages 38--45, Online,
  Oct. 2020. Association for Computational Linguistics.

\bibitem{yu2015lsun}
Fisher Yu, Ari Seff, Yinda Zhang, Shuran Song, Thomas Funkhouser, and Jianxiong
  Xiao.
\newblock Lsun: Construction of a large-scale image dataset using deep learning
  with humans in the loop.
\newblock {\em arXiv preprint arXiv:1506.03365}, 2015.

\end{thebibliography}
}

\clearpage
\appendix
\onecolumn

\begin{table} 
\centering
\small
\caption{The hyperparameters for implementing RQ-Transformer. We follow the same notation in the main paper. $n_e$ represents the dimensionality of features in RQ-Transformer, and \# heads represents the number of heads in self-attentions of RQ-Transformer.}
\label{tab:appendix_rq_transformer}
\begin{tabular}{l|cccccccccc}
\toprule 
Dataset & $N_{\text{spatial}}$ & $N_{\text{depth}} $ & \# params & $K$ & $T$ & $D$ &  $n_z$ & $n_e$ & \# heads & $\tau$ \\ \hline 
LSUN-cat~\cite{yu2015lsun} & 26 & 4 & 612M & 16384 & 64 & 4 & 256 & 1280 & 20 & 0.5 \\
LSUN-bedroom~\cite{yu2015lsun} & 26 & 4 & 612M & 16384 & 64 & 4 & 256 & 1280 & 20 & 0.5 \\
LSUN-church~\cite{yu2015lsun} & 24 & 4 & 370M & 16384 & 64 & 4 & 256 & 1024 & 16 & 0.5 \\
FFHQ~\cite{karras2019style} & 24 & 4 & 370M & 2048 & 64 & 4 & 256 & 1024 & 16 & 1.0 \\
ImageNet~\cite{deng2009imagenet} & 12 & 4 & 480M & 16384 & 12 & 4 & 256 & 1536 & 24 & 0.5\\
ImageNet~\cite{deng2009imagenet} & 24 & 4 & 821M & 16384 & 64 & 4 & 256 & 1536 & 24 & 0.5\\
ImageNet~\cite{deng2009imagenet} & 42 & 6 & 1388M & 16384 & 64 & 4 & 256 & 1536 & 24 & 0.5 \\
ImageNet~\cite{deng2009imagenet} & 42 & 6 & 3822M & 16384 & 64 & 4 & 256 & 2560 & 40 & 0.5 \\
CC-3M~\cite{sharma-etal-2018-conceptual} & 24 & 4 & 654M & 16384 & \textcolor{black}{95} & 4 & 256 & 1280 & 20 & 0.5 \\
\bottomrule
\end{tabular}
\end{table}

\section{Implementation Details}
\subsection{Architecture of RQ-VAE}
For the architecture of RQ-VAE, we follow the architecture of VQ-GAN~\cite{VQGAN} for a fair comparison.
However, we add two residual blocks with 512 channels each followed by a down-/up-sampling block to extract feature maps of resolution 8$\times$8.

\subsection{Architecture of RQ-Transformer}
The RQ-Transformer, which consists of the spatial transformer and the depth transformer, adopts a stack of self-attention blocks~\cite{transformer} for each compartment.
In Table~\ref{tab:appendix_rq_transformer}, we include the detailed information of hyperparameters to implement our RQ-Transformers.
\textcolor{black}{All RQ-Transformers in Table~\ref{tab:appendix_rq_transformer} uses RQ-VAE with 8$\times$8$\times$4 shape of codes.}
\textcolor{black}{For CC-3M, the length of text conditions is 32, and the last token in text conditions predicts the code at the first position of images. Thus, the total sequence length ($T$) of RQ-Transformer is 95.}

\subsection{Training Details}
For ImageNet, RQ-VAE is trained for 10 epochs with batch size 128.
We use the Adam optimizer~\cite{Adam} with $\beta_1=0.5$ and $\beta_2=0.9$, and learning rate is set 0.00004.
The learning rate is linearly warmed up during the first 0.5 epoch.
We do not use learning rate decay, weight decaying, nor dropout.
For the adversarial and perceptual loss, we follow the experimental setting of VQ-GAN~\cite{VQGAN}. In particular, the weight for the adversarial loss is set 0.75 and the weight for the perceptual loss is set 1.0.
To increase the codebook usage of RQ-VAE, we use random restart of unused codes proposed in JukeBox~\cite{Jukebox}.
For LSUN-$\{\text{cat, bedroom, church}\}$, we use the pretrained RQ-VAE on ImageNet and finetune it for one epoch with 0.000004 of learning rate.
For FFHQ, we train RQ-VAE for 150 epochs of training data with 0.00004 of learning rate and five epochs of warm-up.
For CC-3M, we use the pretrained RQ-VAE on ImageNet without finetuning.

All RQ-Transformers are trained using the AdamW optimizer~\cite{AdamW} with $\beta_1=0.9$ and $\beta_2=0.95$.
We use the cosine learning rate schedule with 0.0005 of the initial learning rate. 
The RQ-Transformer is trained for 90, 200, 300 epochs for LSUN-bedroom, -cat, and -church respectively.
The weight decay is set 0.0001, and the batch size is 16 for FFHQ and 2048 for other datasets.
In all experiments, the dropout rate of each self-attention block is set 0.1 \textcolor{black}{except 0.3 for 3.8B parameters of RQ-Transformer}.
We use eight NVIDIA A100 GPUs to train RQ-Transformer of 1.4B parameters, and four GPUs to train RQ-Transformers of other sizes.
The training time is $<$9 days for LSUN-cat, LSUN-bedroom, $<$4.5 days for ImageNet, and CC-3M, and $<$1 day for LSUN-church and FFHQ.
We use the early stopping at 39 epoch for the FFHQ dataset, considering the overfitting of the RQ-Transformer due to the small scale of the dataset.

\section{Additional Results of Generated Images by RQ-Transformer}
\subsection{Additional Examples of Unconditional Image Generation for LSUNs and FFHQ}
\begin{figure}
    \centering
    \includegraphics[width=\textwidth]{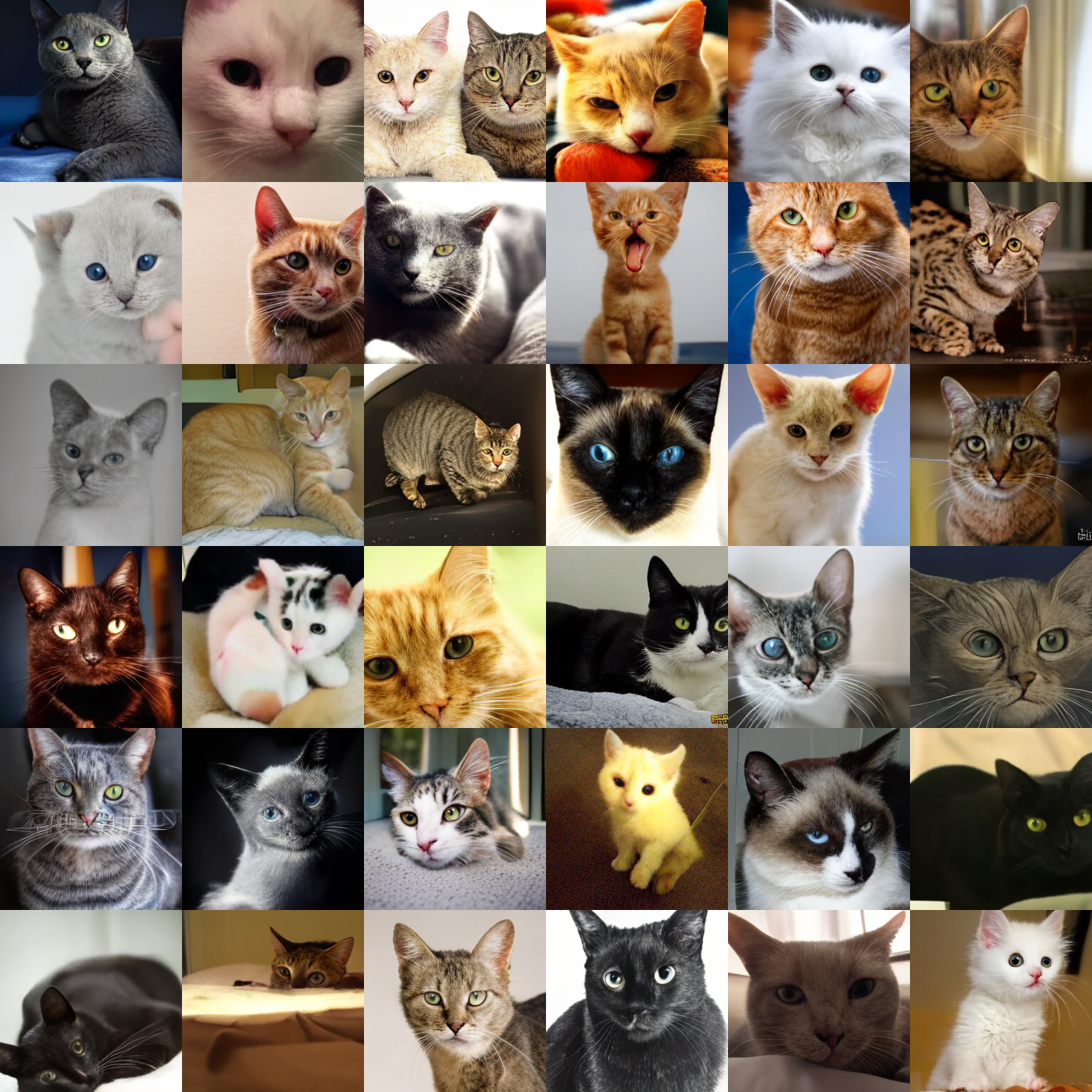}
    \caption{Additional examples of unconditional image generation by our model trained on LSUN-cat.}
    \label{fig:appendix_cat}
\end{figure}

\begin{figure}
    \centering
    \includegraphics[width=\textwidth]{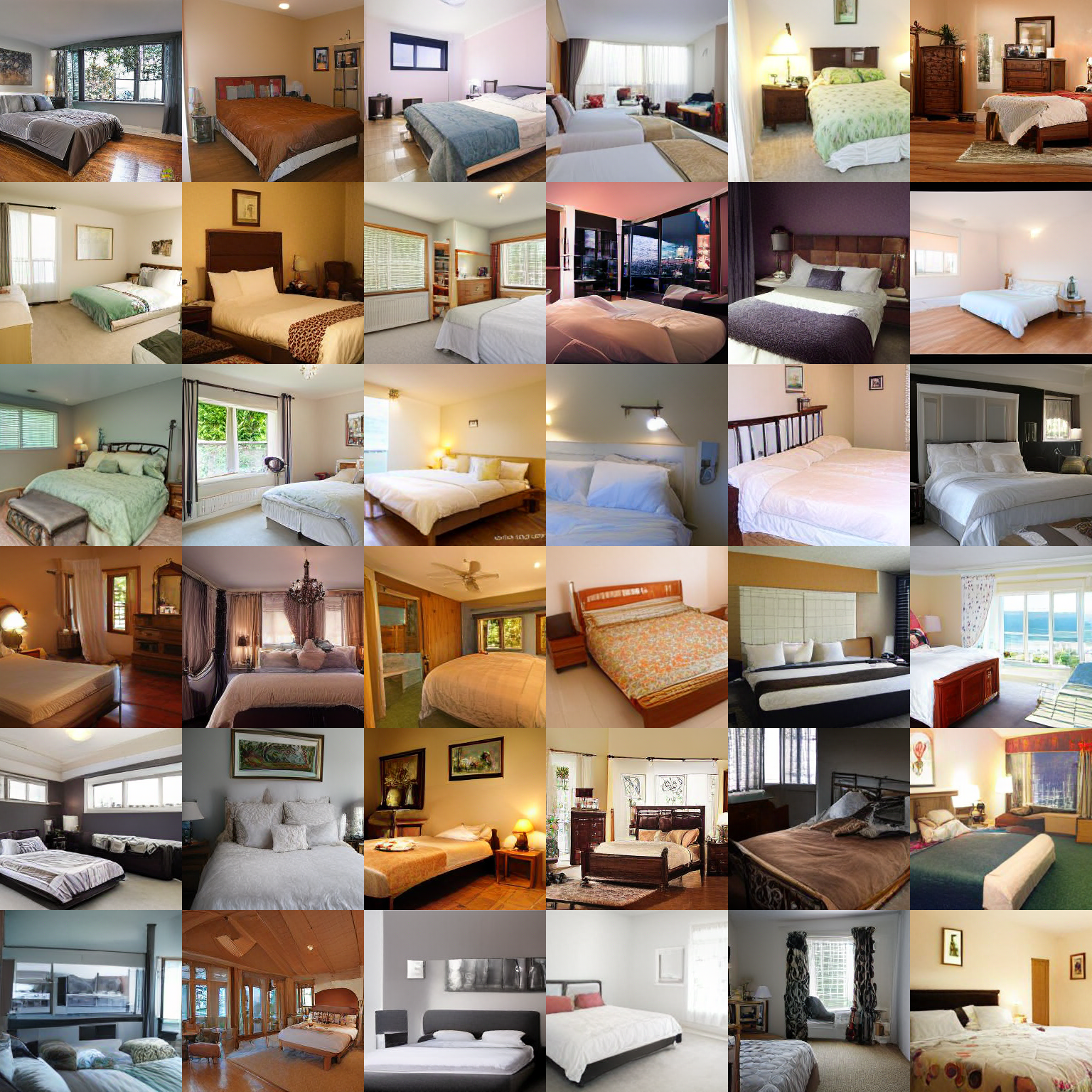}
    \caption{Additional examples of unconditional image generation by our model trained on LSUN-bedroom.}
    \label{fig:appendix_bedroom}
\end{figure}

\begin{figure}
    \centering
    \includegraphics[width=\textwidth]{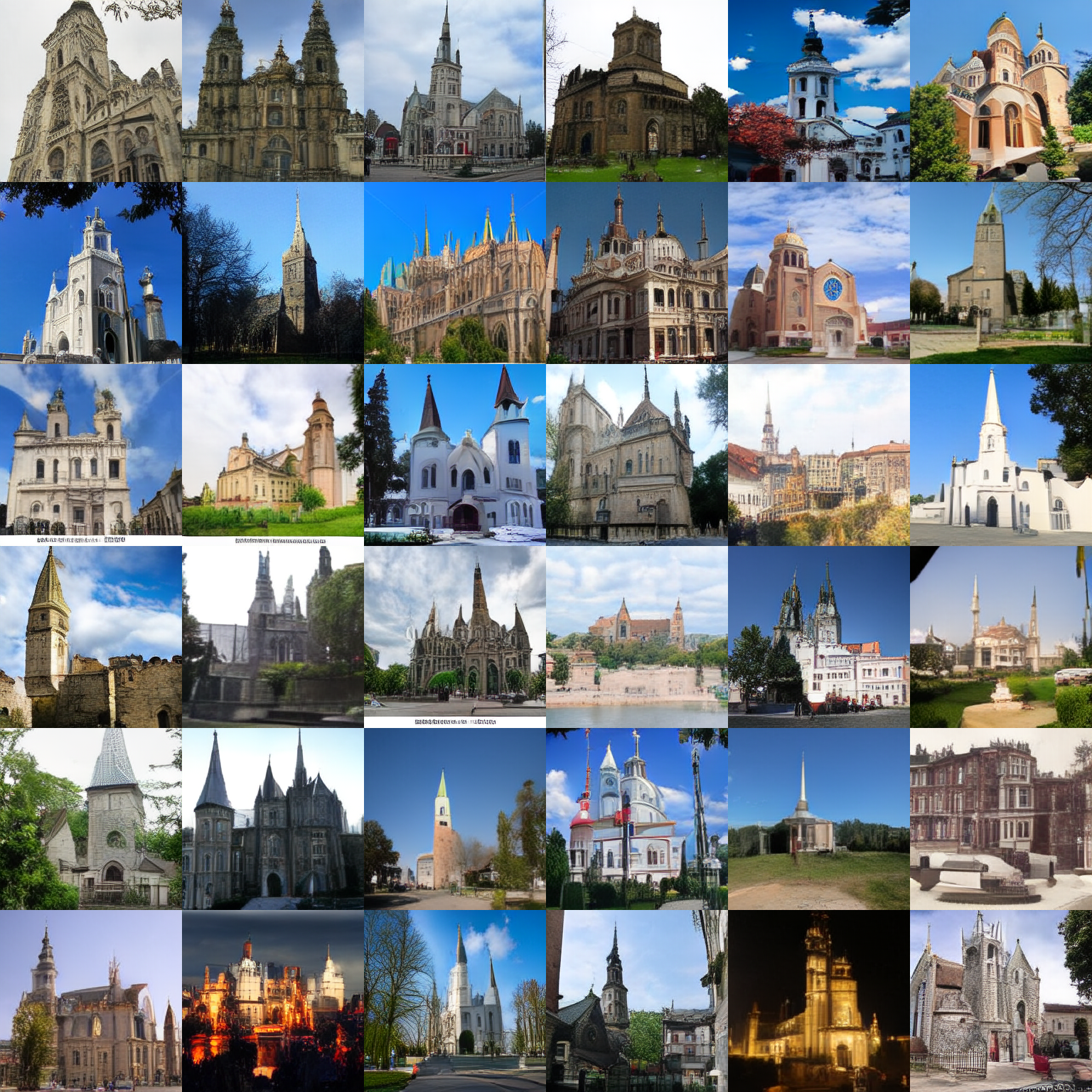}
    \caption{Additional examples of unconditional image generation by our model trained on LSUN-church.}
    \label{fig:appendix_church}
\end{figure}

\begin{figure}
    \centering
    \includegraphics[width=\textwidth]{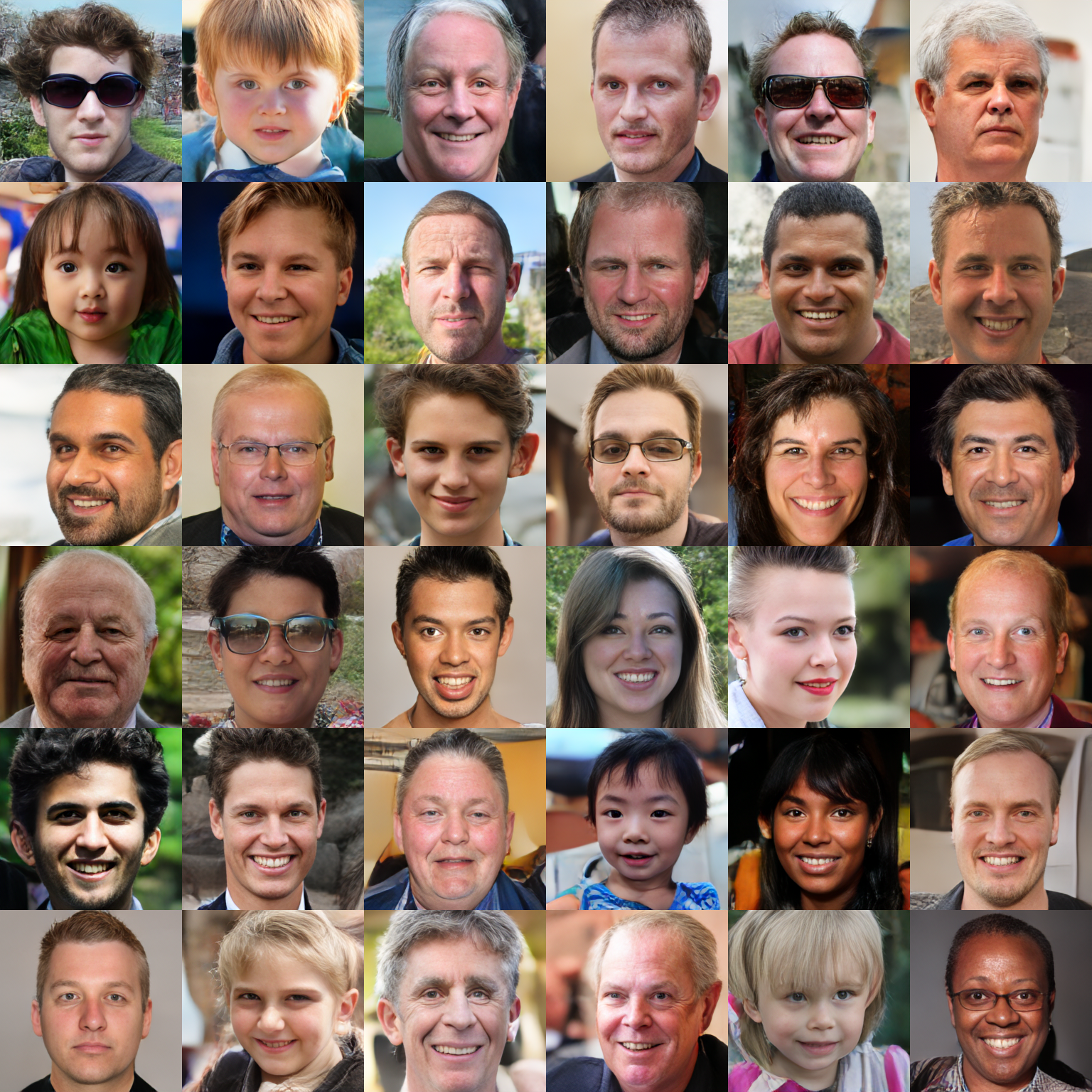}
    \caption{Additional examples of unconditional image generation by our model trained on FFHQ.}
    \label{fig:appendix_ffhq}
\end{figure}

We show the additional examples of unconditional image generation by RQ-VAEs trained on LSUN-$\{\text{cat, bedroom, church}\}$ and FFHQ.
Figure~\ref{fig:appendix_cat},~\ref{fig:appendix_bedroom},~\ref{fig:appendix_church}, and~\ref{fig:appendix_ffhq} show the results of LSUN-cat, LSUN-bedroom LSUN-church, and FFHQ, respectively.
For the top-$k$ (top-$p$) sampling, 512 (0.9), 8192 (0.85), 1400 (1.0), and 2048 (0.95) are used respectively.

\subsection{Nearest Neighbor Search of Generated Images for FFHQ}
\begin{figure}
    \centering
    \includegraphics[width=\textwidth]{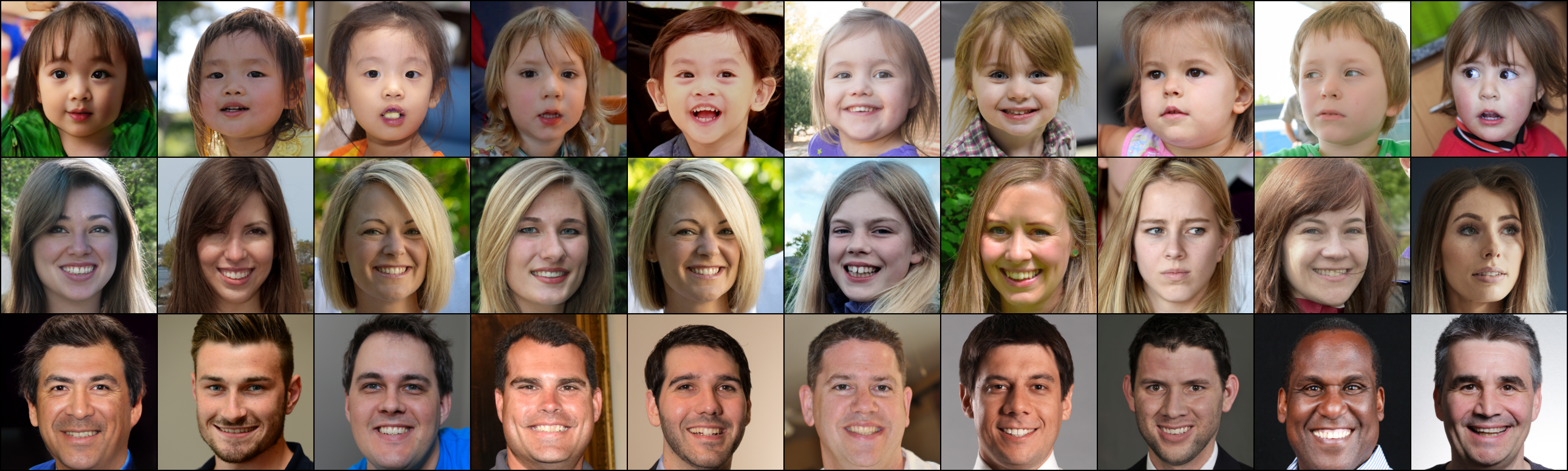}
    \caption{Visualization of nearest neighbors in the FFHQ training samples to our generated samples. In each row, the first image is our generation. The nearest neighbors to the first image are visualized according to the similarity of VGG-16 features in descending order.}
    \label{fig:memorization_test}
\end{figure}

For the training of FFHQ, we use early stopping for RQ-Transformer when the validation loss is minimized, since RQ-Transformer can memorize all training samples due to the small scale of FFHQ.
Despite the use of early stopping, we further examine whether our model memorizes the training samples or generates new images.
To visualize the nearest neighbors in the training images of FFHQ to generated images, we use a KD-tree~\cite{sklearn_api}, which is constructed by the VGG-16 features~\cite{VGG} of training images.
Figure~\ref{fig:memorization_test} shows that our model does not memorize the training data, but generates new face images for unconditional sample generation of FFHQ.

\subsection{Ablation Study on Soft Labeling and Stochastic Sampling}
\textcolor{black}{
For 821M parameters of RQ-Transformer trained on ImageNet, RQ-Transformer achieves 14.06 of FID score when neither stochastic sampling nor soft labeling is used. When stochastic sampling is applied to the training of RQ-Transformer, 13.24 of FID score is achieved. When only soft labeling is used without stochastic sampling, RQ-Transformer achieves 14.87 of FID score, and the performance worsens. However, when both stochastic sampling and soft labeling are used together, RQ-Transformer achieves 13.11 of FID score, which is improved performance than baseline.}

\subsection{Additional Examples of Class-Conditioned Image Generation for ImageNet}

\begin{figure}
    \centering
    \includegraphics[width=0.8\textwidth]{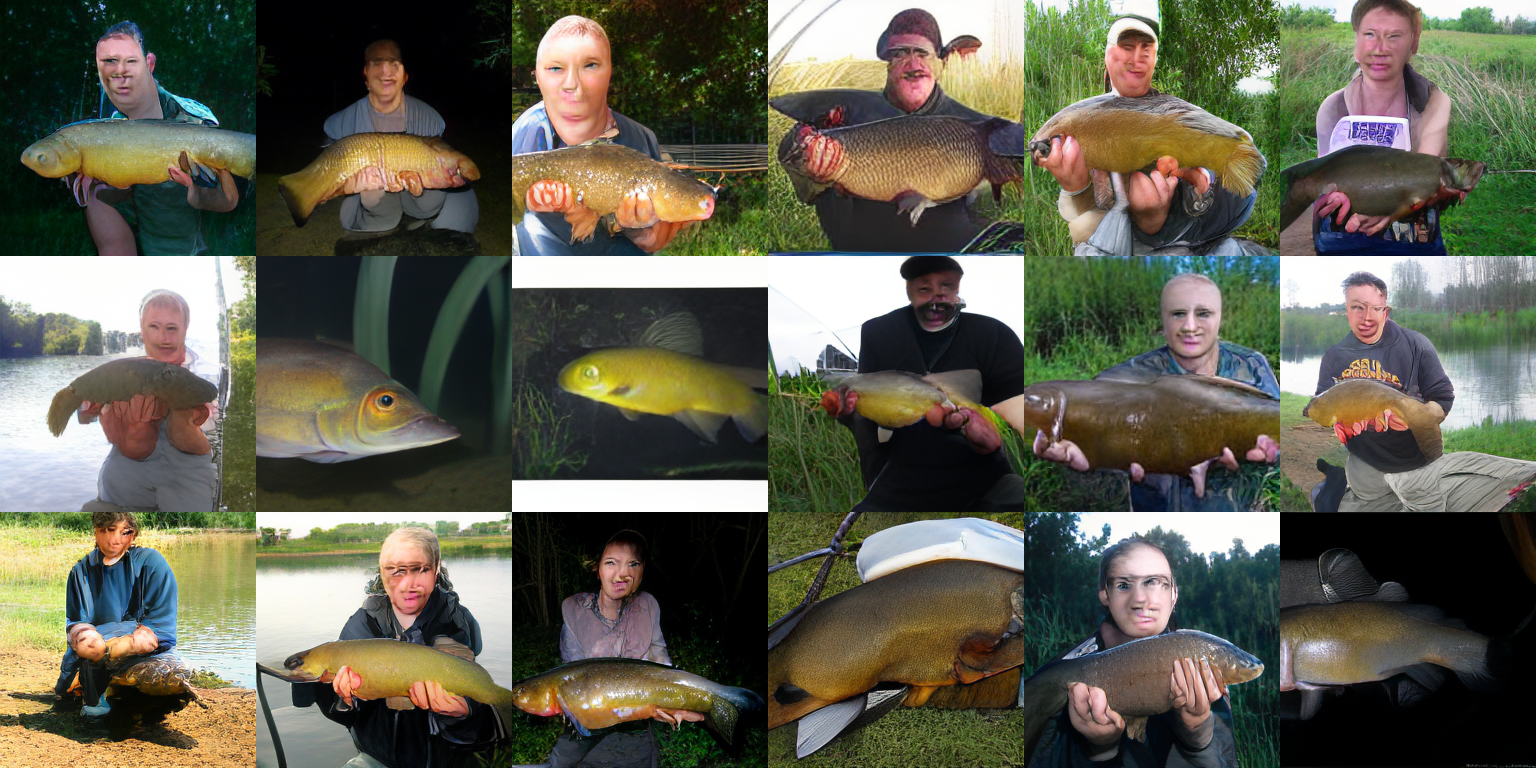}
    \includegraphics[width=0.8\textwidth]{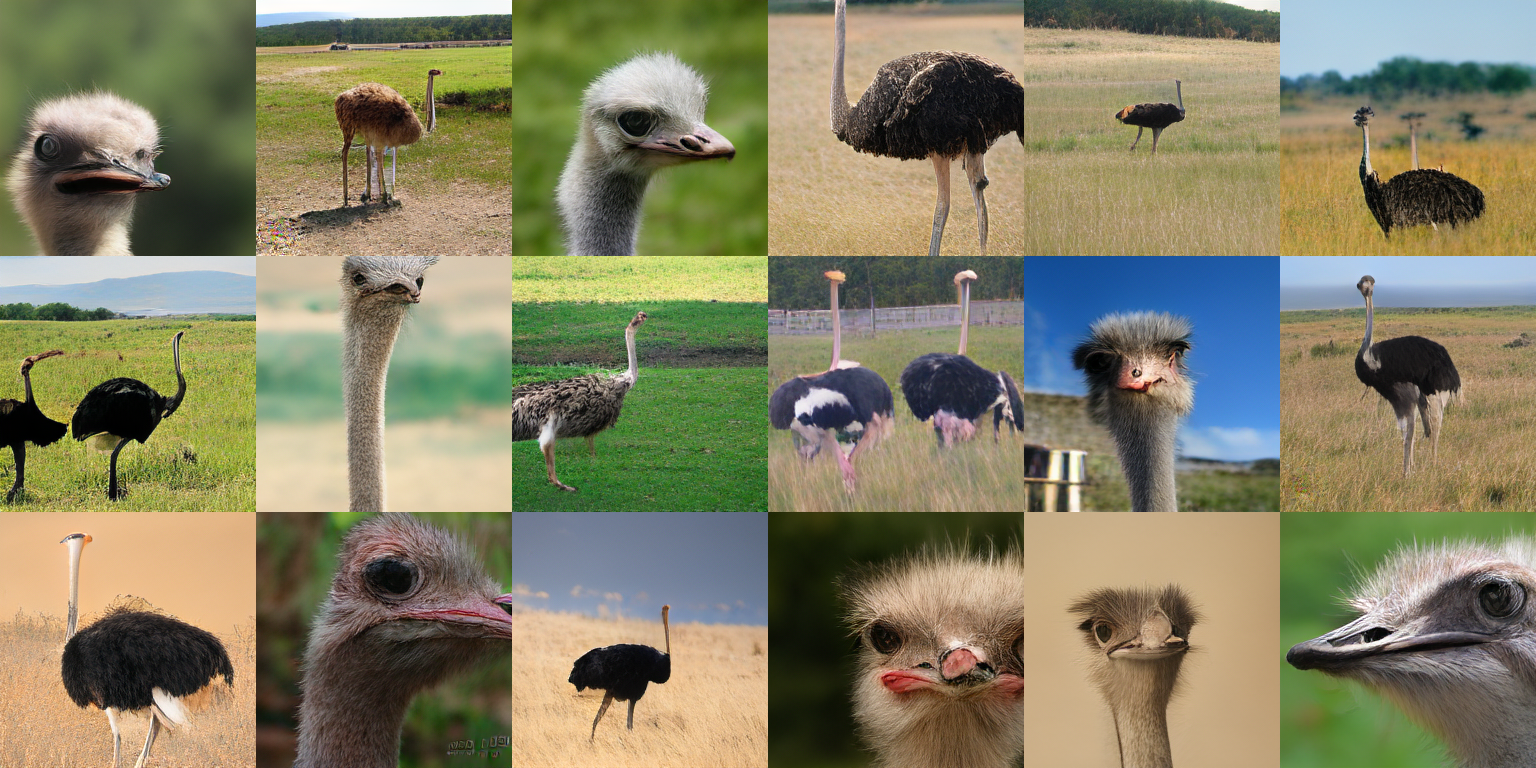}
    \includegraphics[width=0.8\textwidth]{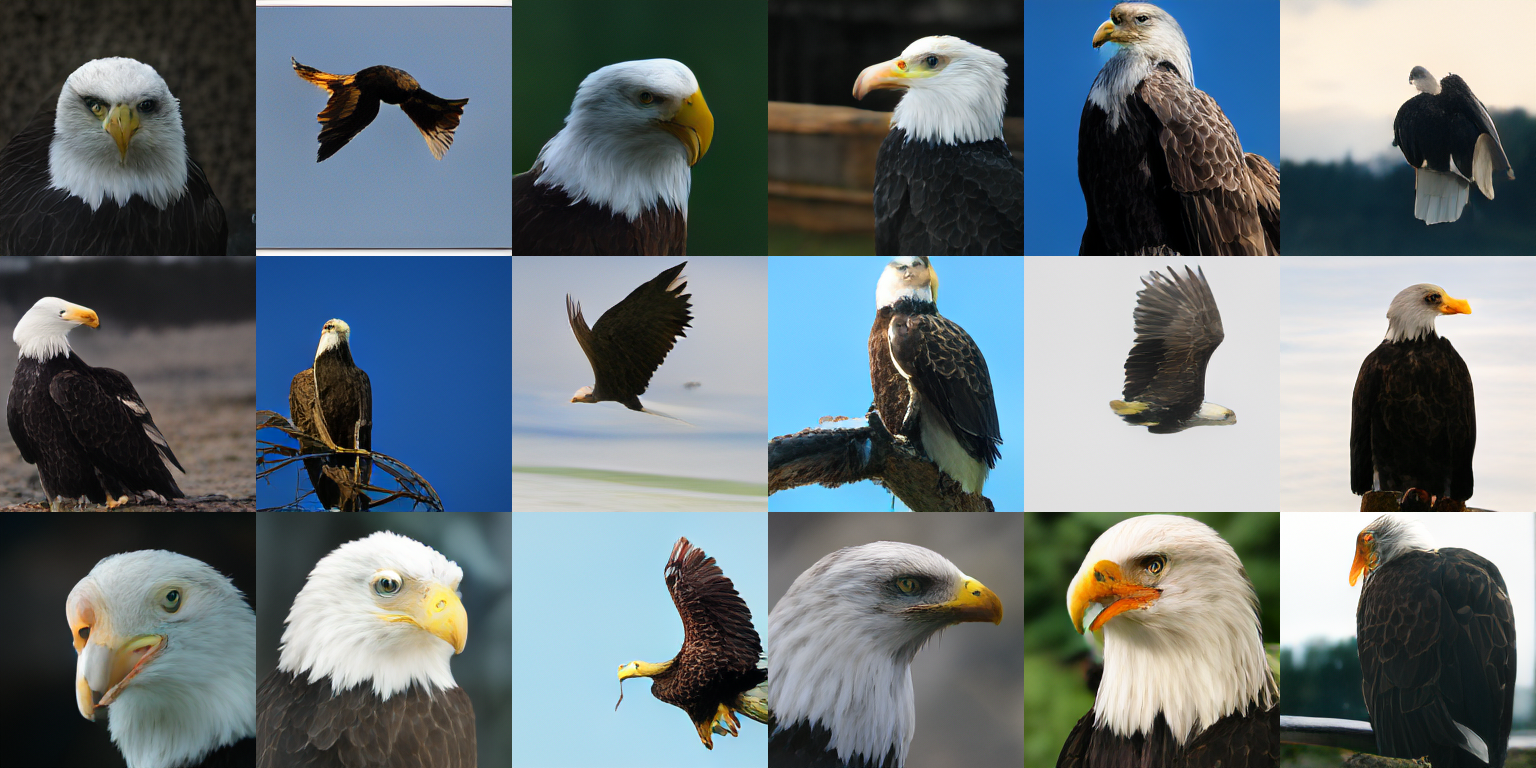}
    \caption{Additional examples of conditional image generation by 1.4B parameters of RQ-Transformer trained on ImageNet. Top: Tench (0). Middle: Ostrich (9). Bottom: Bald eagle (22).}
    \label{fig:appendix_cIN_0}
\end{figure}

\begin{figure}
    \centering
    \includegraphics[width=0.8\textwidth]{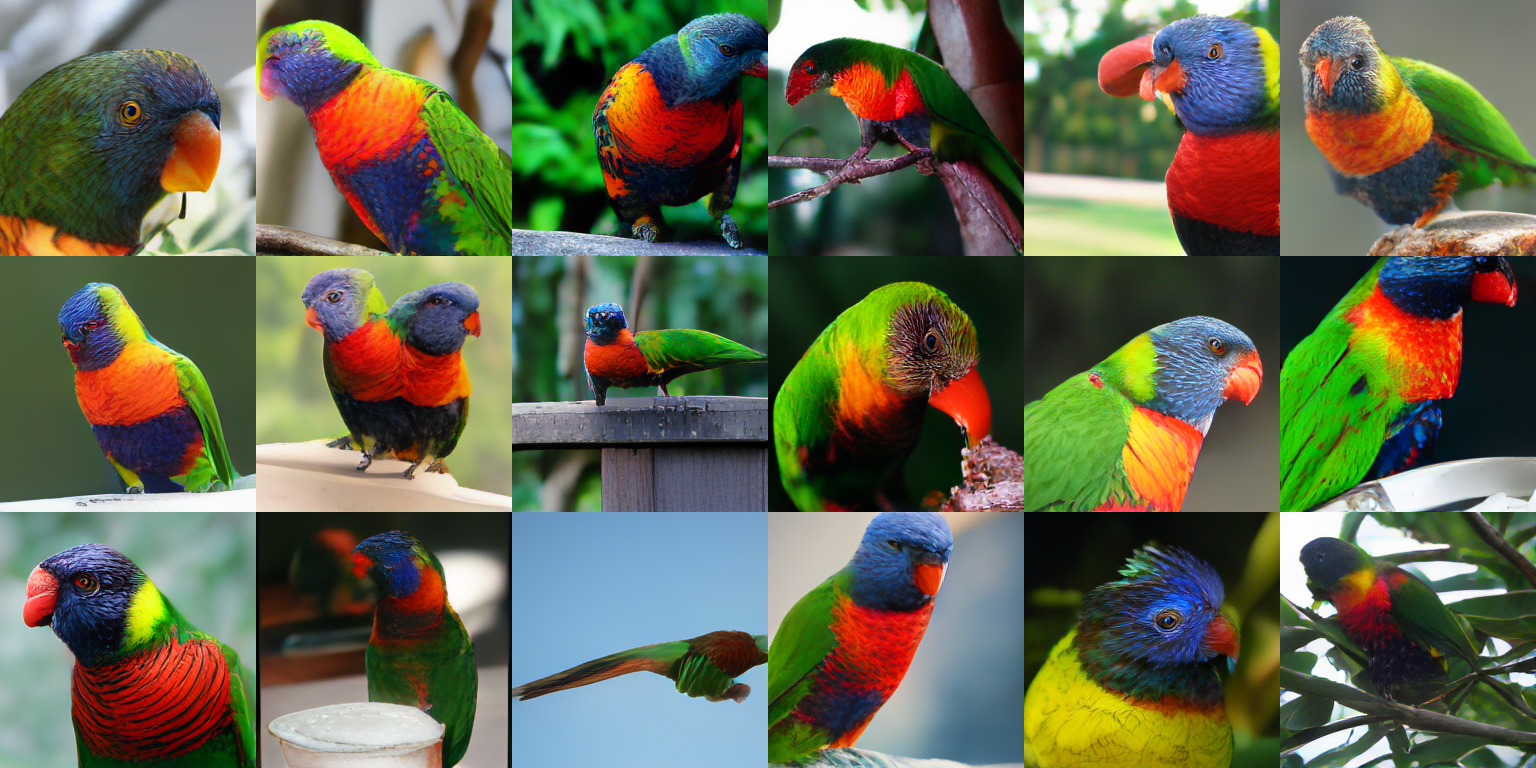}
    \includegraphics[width=0.8\textwidth]{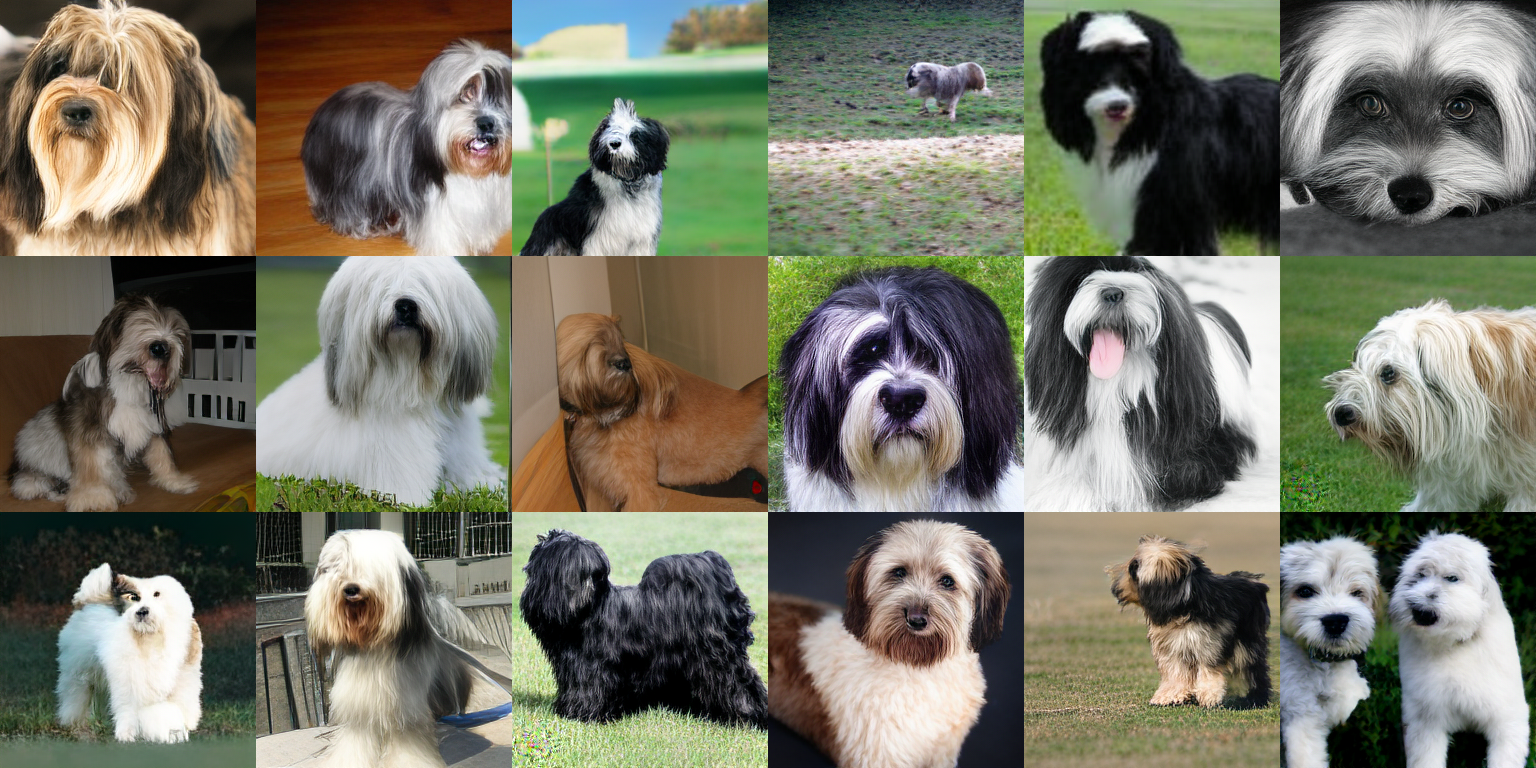}
    \includegraphics[width=0.8\textwidth]{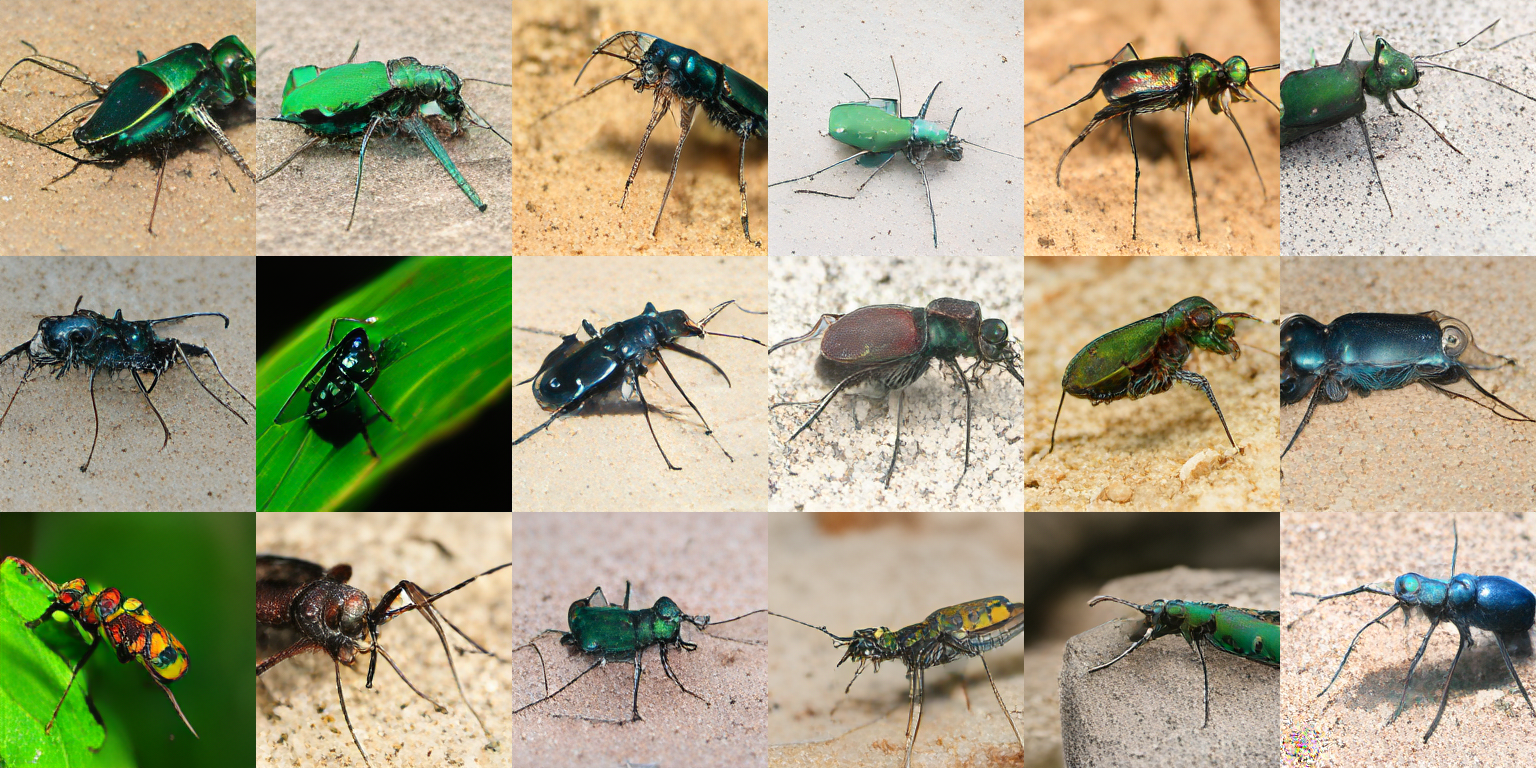}
    \caption{Additional examples of conditional image generation by 1.4B parameters of RQ-Transformer trained on ImageNet. Top: Lorikeet (90). Middle: Tibetan terrier (200). Bottom: Tiger beetle (300). }
    \label{fig:appendix_cIN_1}
\end{figure}

\begin{figure}
    \centering
    \includegraphics[width=0.8\textwidth]{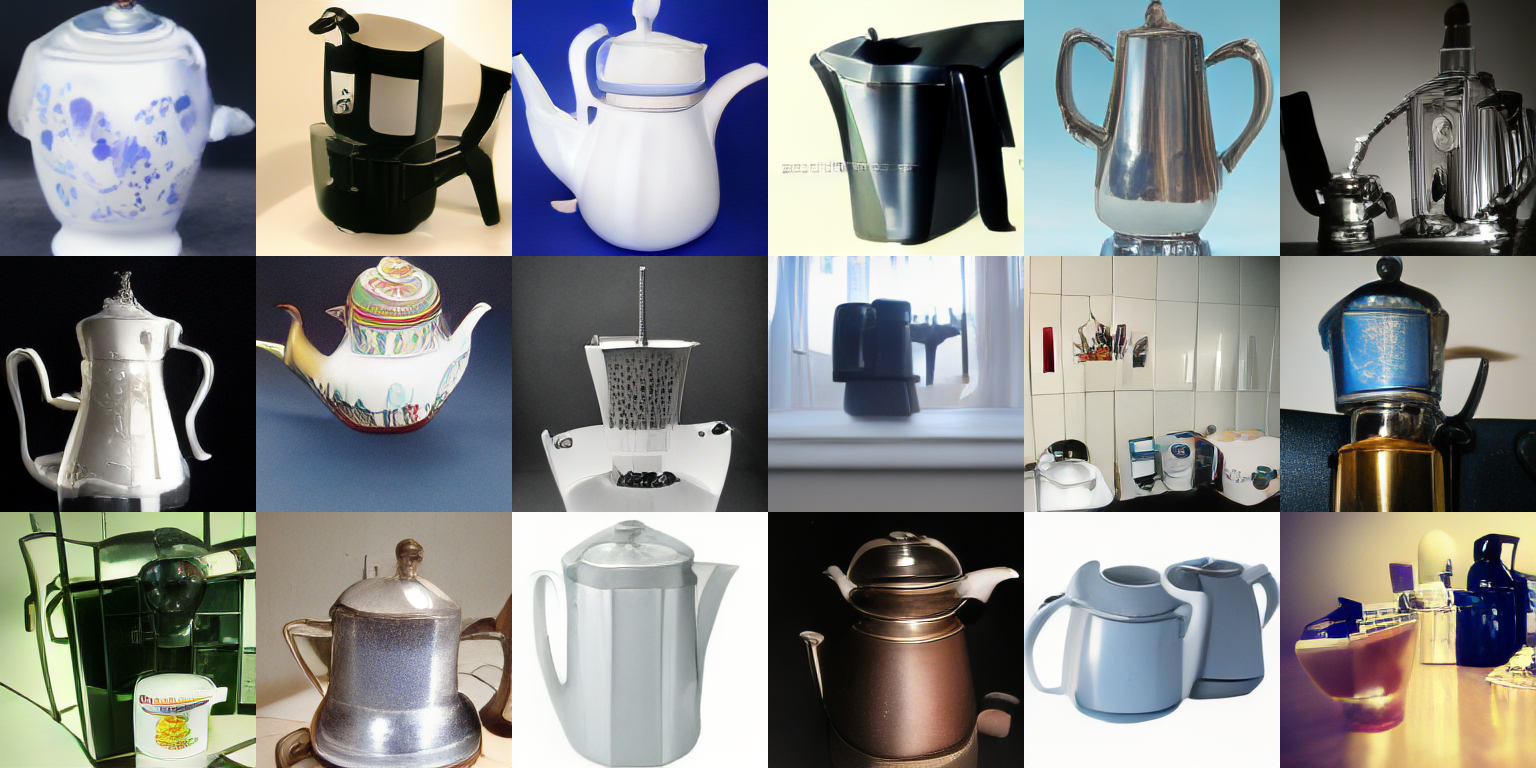}
    \includegraphics[width=0.8\textwidth]{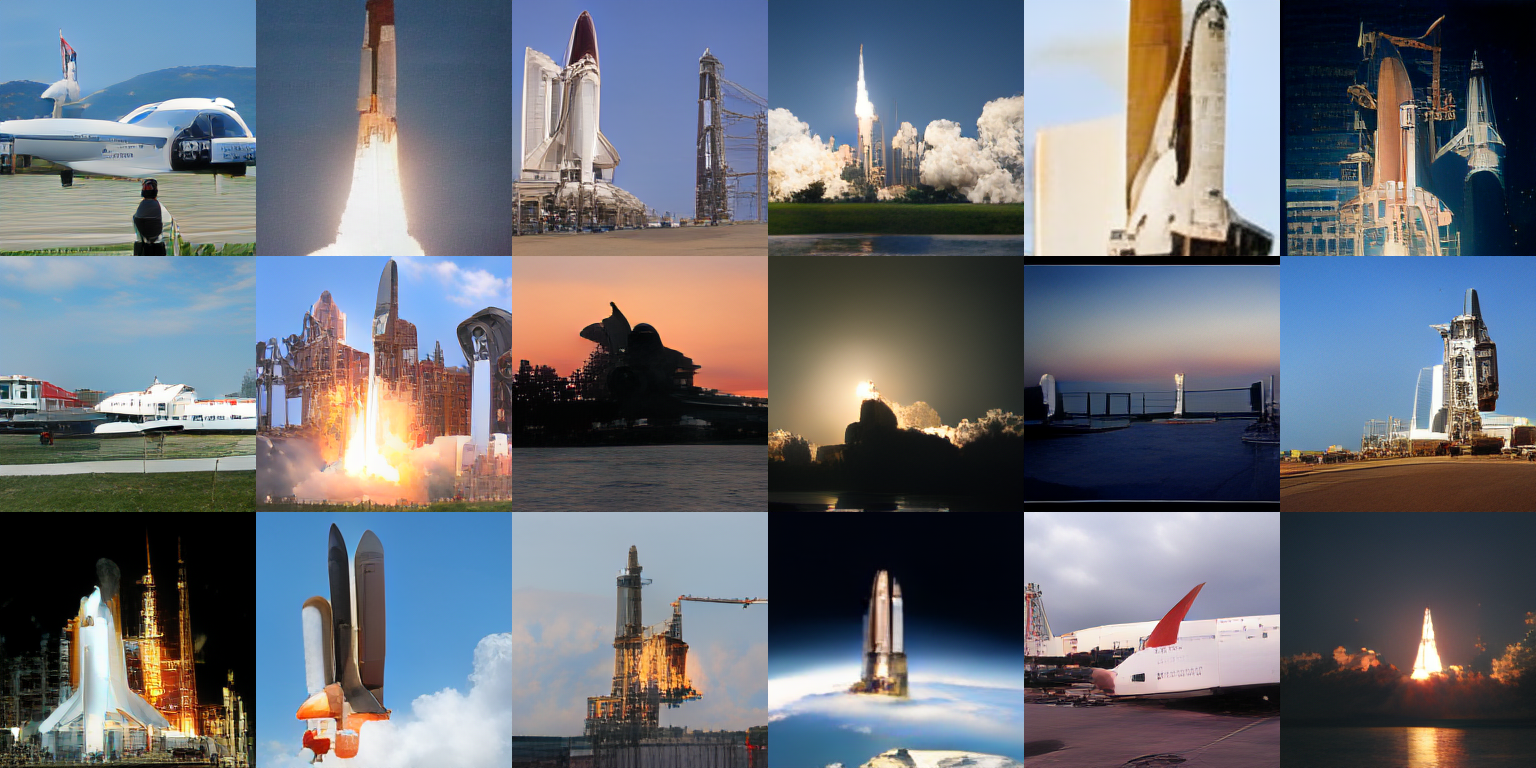}
    \includegraphics[width=0.8\textwidth]{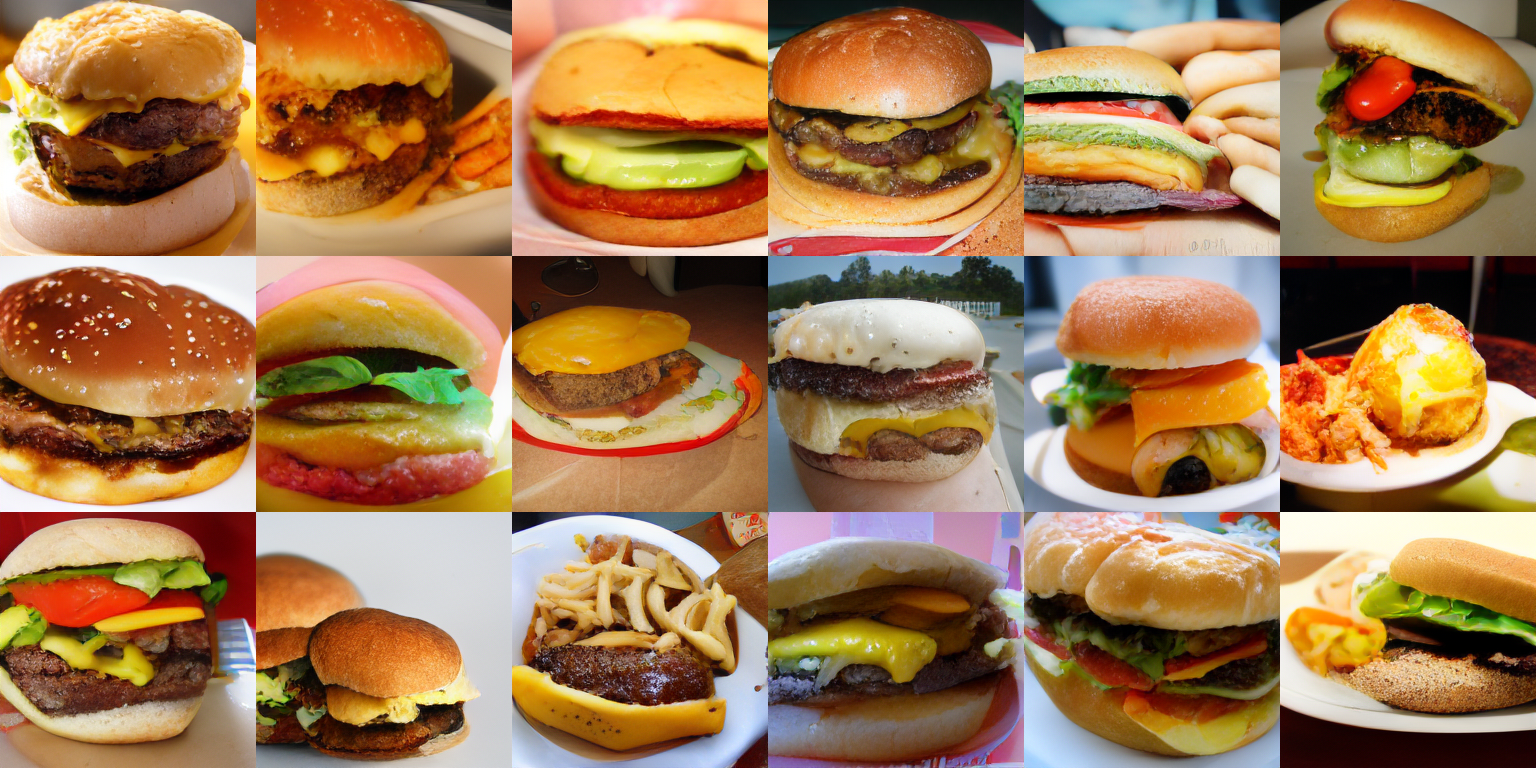}
    \caption{Additional examples of conditional image generation by 1.4B parameters of RQ-Transformer trained on ImageNet. Top: Coffee pot (505). Middle: Space shuttle (812). Bottom: Cheeseburger (933).}
    \label{fig:appendix_cIN_2}
\end{figure}

\begin{figure}
    \centering
    \includegraphics[width=0.8\textwidth]{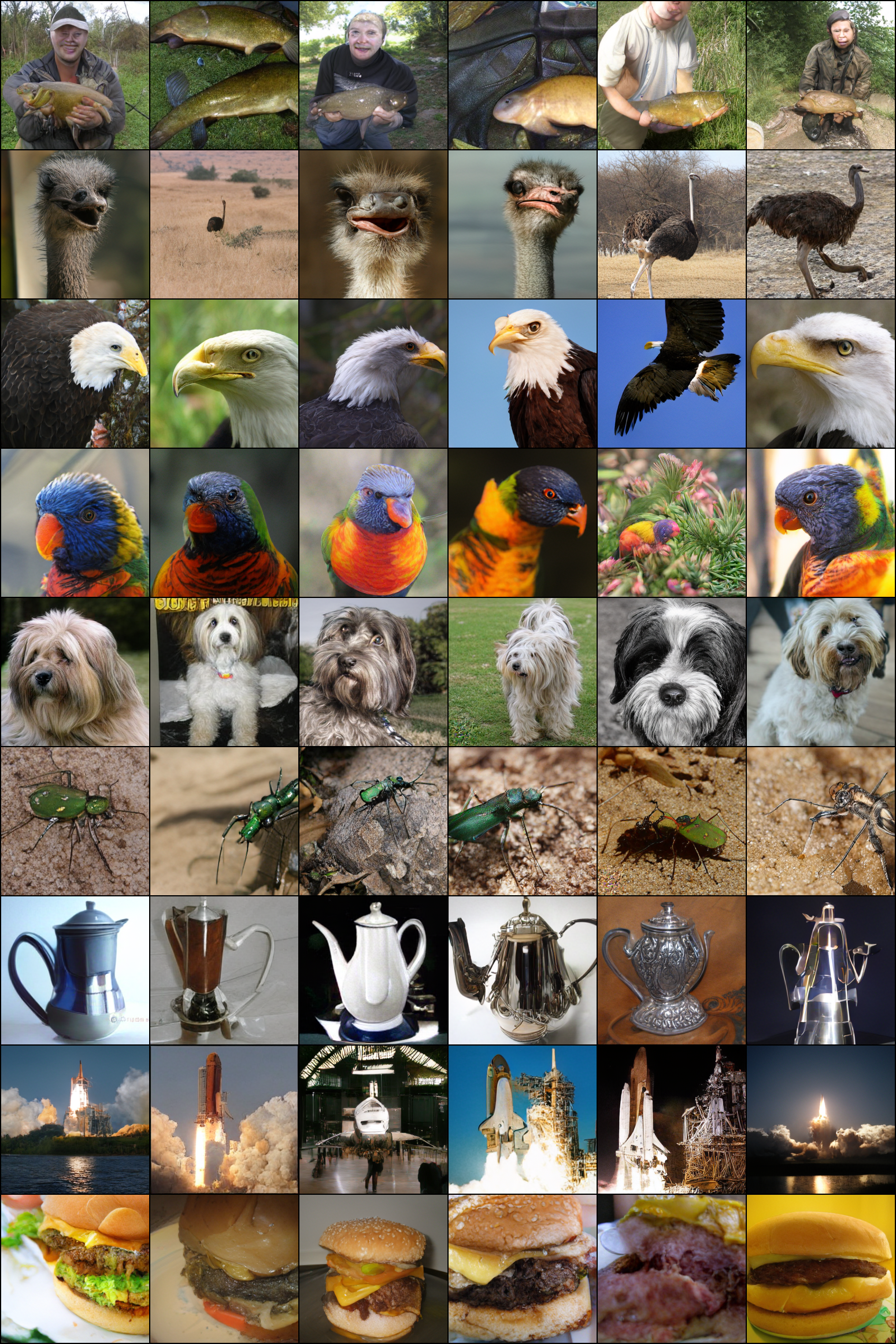}
    \caption{Additional examples of conditional image generation by 3.8B parameters of RQ-Transformer trained on ImageNet. The classes of images in each line are tench (0), ostrich (9), bald eagle (22), lorikeet (90), tibetan terrier (200),  tiger beetle (300), coffee pot (505), space shuttle (812), and cheeseburger (933), respectively.}
    \label{fig:appendix_cIN_3}
\end{figure}

\begin{figure}
    \centering
    \includegraphics[width=\textwidth]{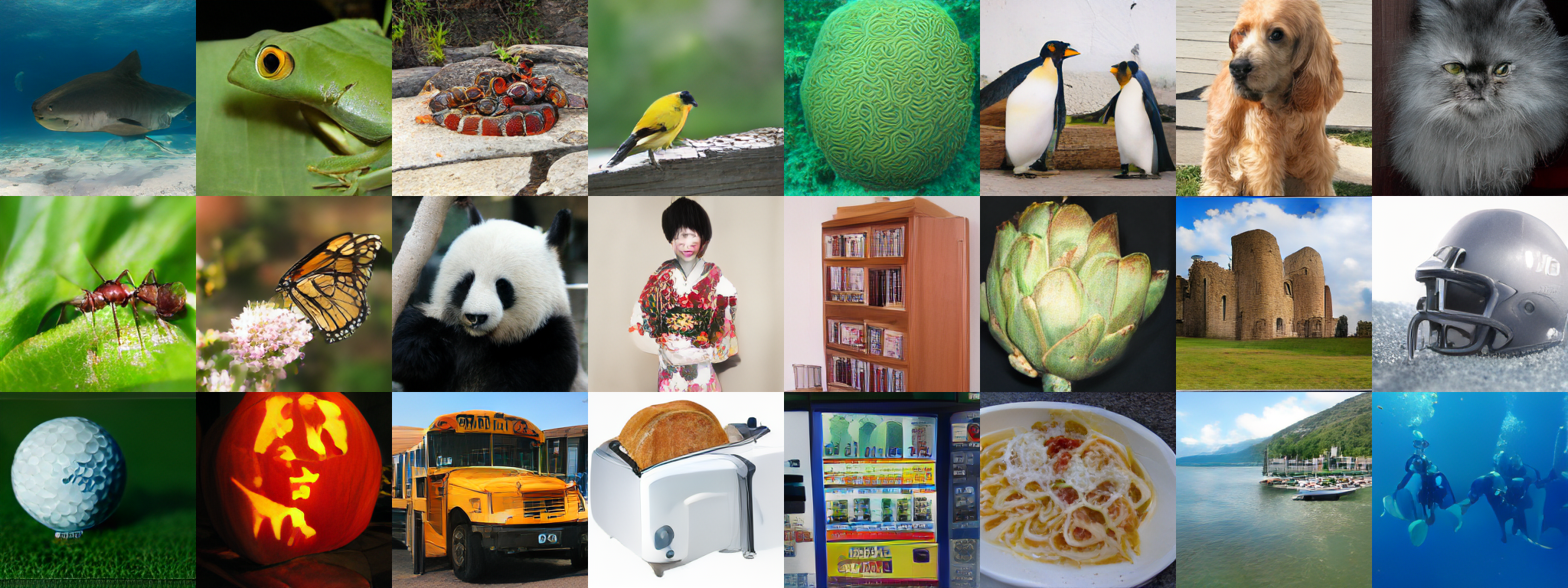}
    \includegraphics[width=\textwidth]{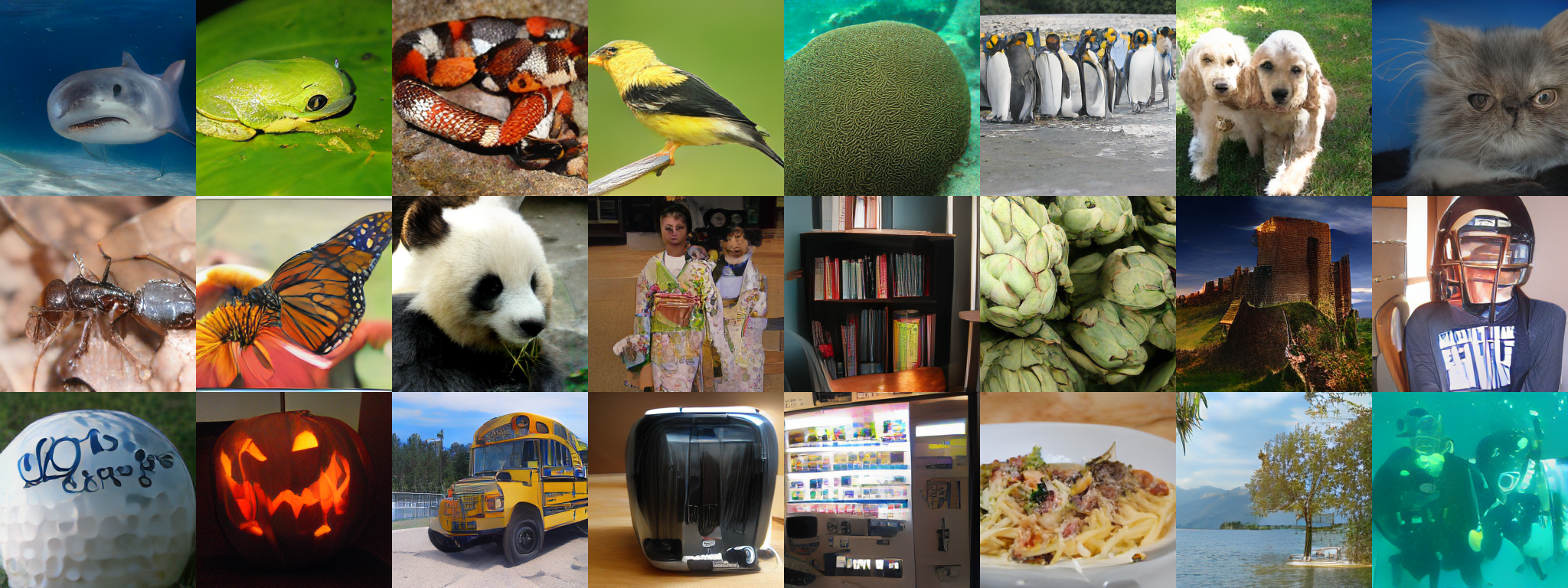}
    \includegraphics[width=\textwidth]{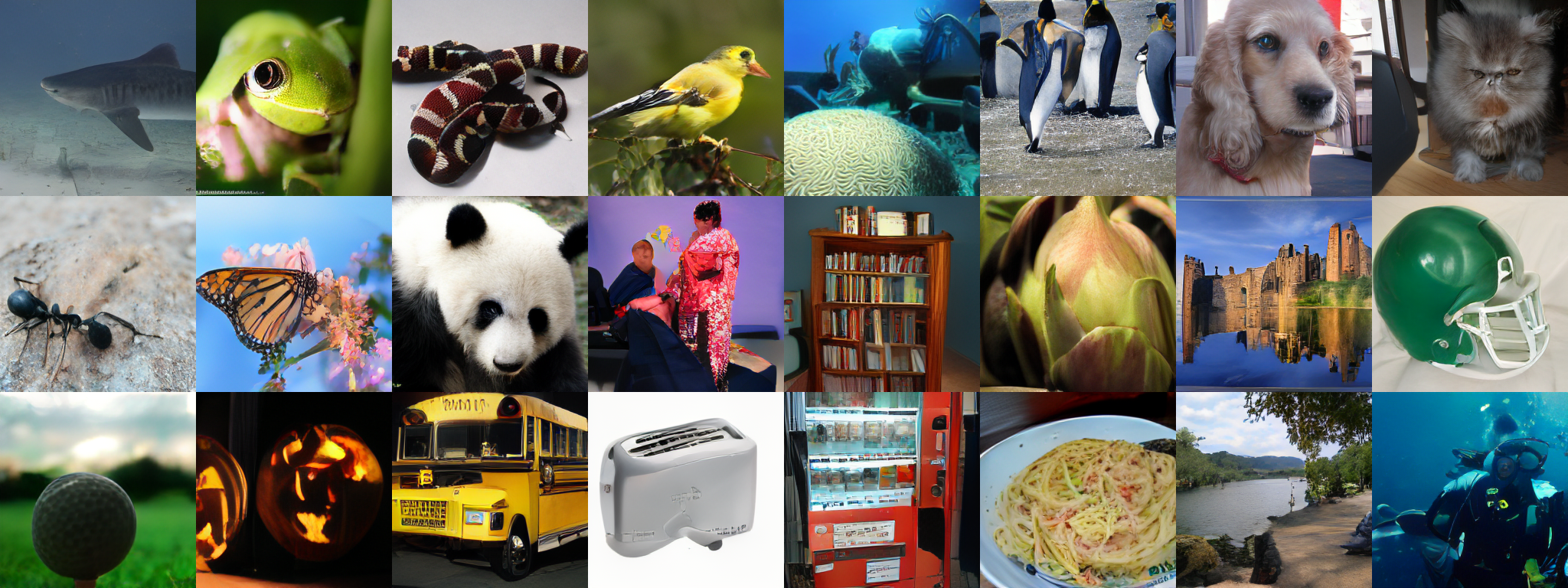}
    \caption{Additional examples of conditional image generation by 1.4B parameters of RQ-Transformer trained on ImageNet with rejection sampling. All images are sampled with top-$p$=1.0 with different acceptance rates of the rejection sampling and top-$k$ values. (Top) Acceptance rate=0.5 and top-$k$=512, Middle: Acceptance rate=0.25 and top-$k$=1024. Bottom: Acceptance rate=0.05 and top-$k$=2048.}
    \label{fig:appendix_cIN_rej}
\end{figure}

We visualize the additional examples of class-conditional image generation by RQ-Transformer trained on ImageNet. Figure \ref{fig:appendix_cIN_0}, \ref{fig:appendix_cIN_1}, and \ref{fig:appendix_cIN_2} show the generated samples by RQ-Transformer with 1.4B parameters conditioned on a few selected classes. Those images are sampled with top-$k$ 512 and top-$p$ 0.95.
In addition, Figure~\ref{fig:appendix_cIN_rej} shows the generated samples using the rejection sampling with ResNet-101 is applied with various acceptance rates. 
The images in Figure~\ref{fig:appendix_cIN_rej} are sampled with fixed $p=1.0$ for top-$p$ sampling and different acceptance rates of the rejection sampling and top-$k$ values.
The (top-$k$, acceptance rate)s are (512, 0,5), (1024, 0.25), and (2048, 0.05), and their corresponding FID scores are 7.08, 5.62, and 4.45. 
Figure~\ref{fig:appendix_cIN_3} shows the generated samples of RQ-Transformer with 3.8B parameters using rejection sampling with (4098, 0.125).

\subsection{Additional Examples of Text-Conditioned Image Generation for CC-3M}

\begin{figure}
    \centering
    \begin{tabular}{m{0.18\textwidth}m{0.78\textwidth}}
         \textit{A photograph of crowd of people enjoying night market.} & \includegraphics[width=0.78\textwidth]{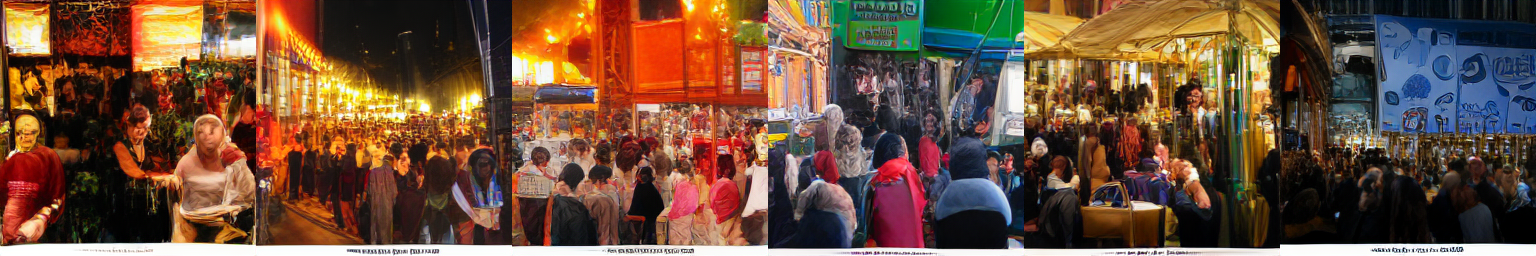} \\
         \textit{A photograph of crowd of people under cherry blossom trees.} & \includegraphics[width=0.78\textwidth]{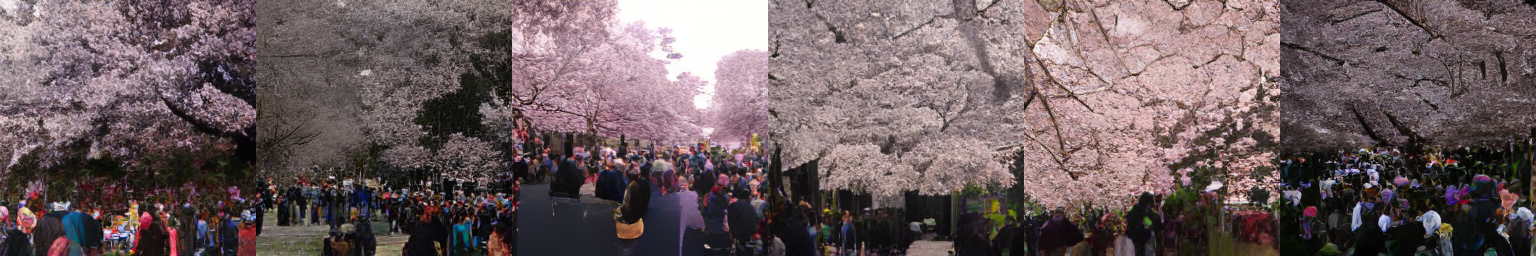} \\ \hline \\ [-0.85em]
         \textit{A small house in the wilderness.} & \includegraphics[width=0.78\textwidth]{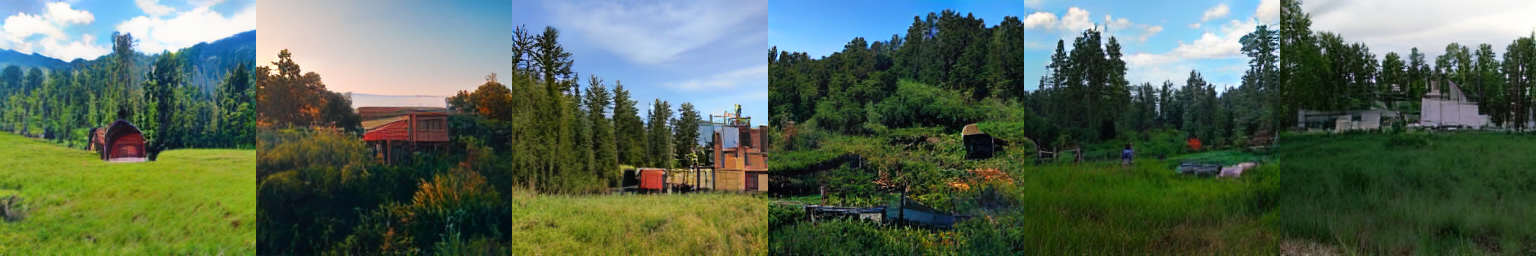} \\
         \textit{A small house on the shore.} & \includegraphics[width=0.78\textwidth]{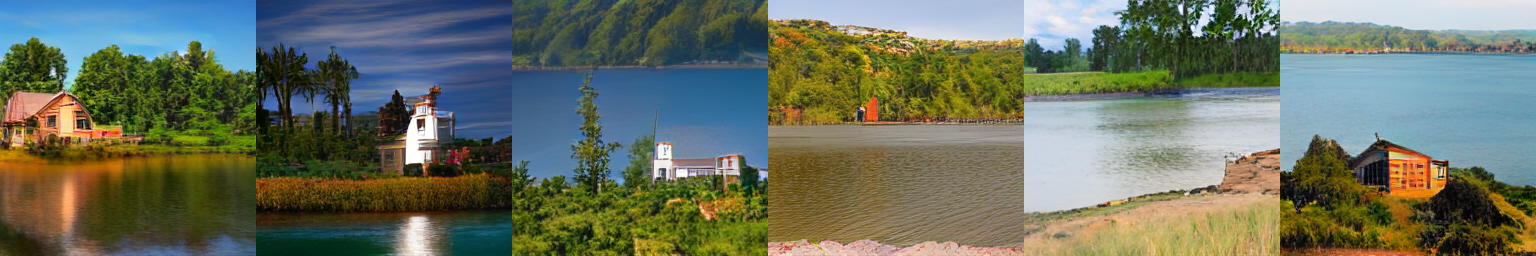} \\ \hline \\ [-0.85em]
         \textit{Sunset over the skyline of a city.} & \includegraphics[width=0.78\textwidth]{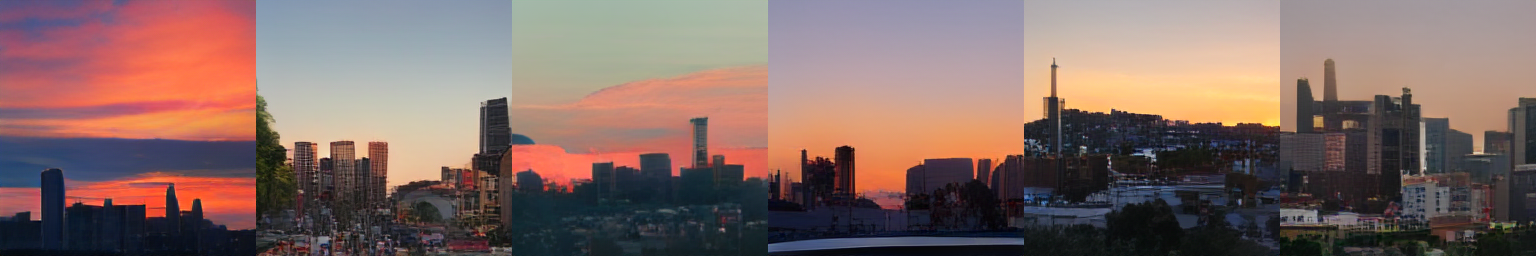} \\
         \textit{Night landscape of the skyline of a city.} & \includegraphics[width=0.78\textwidth]{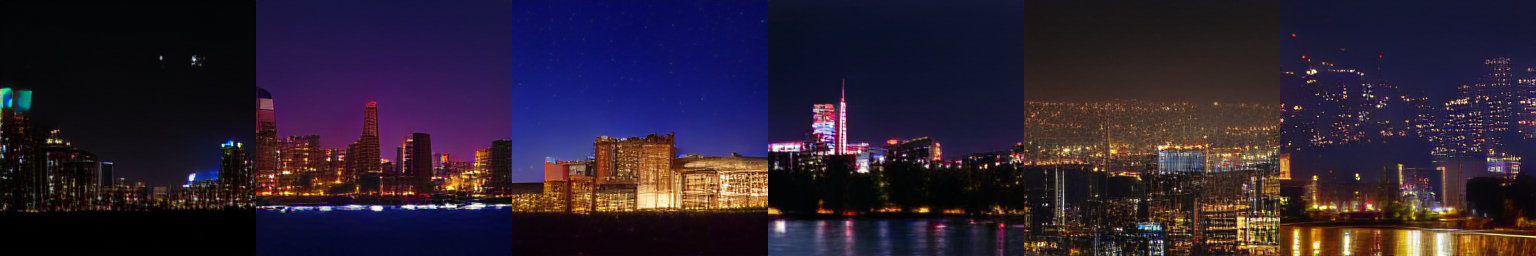} \\ \hline \\ [-0.85em]
         \textit{An illustration of a cathedral.} & \includegraphics[width=0.78\textwidth]{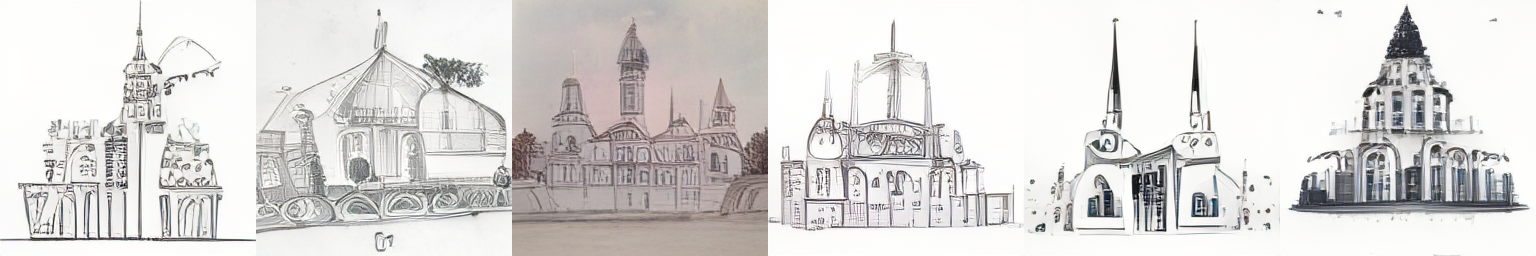} \\
         \textit{A painting of a cathedral.} & \includegraphics[width=0.78\textwidth]{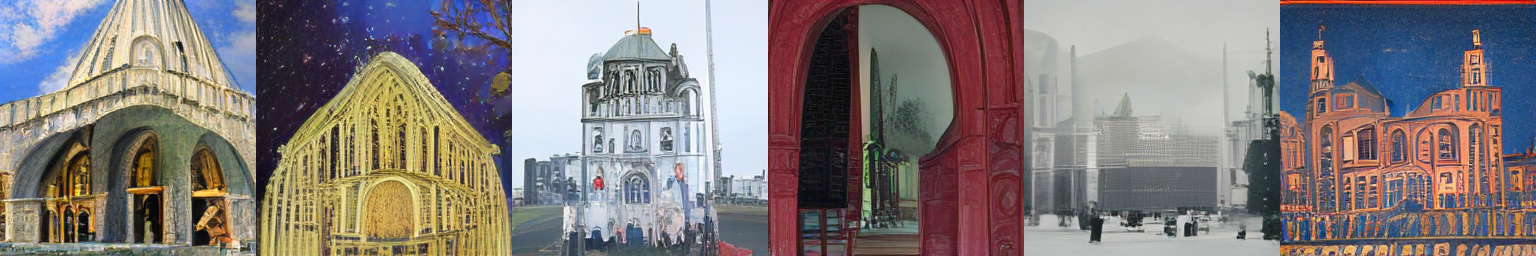} 
    \end{tabular}
    \caption{Additional examples of text-conditional image generation by our model trained on CC-3M. The text conditions are customized prompts, which are unseen during the training of RQ-Transformer. All images are sampled with top-$k$ 1024 and top-$p$ 0.9.}
    \label{fig:appendix_cc3m_control}
\end{figure}

\begin{figure}
    \centering
    \includegraphics[width=\textwidth]{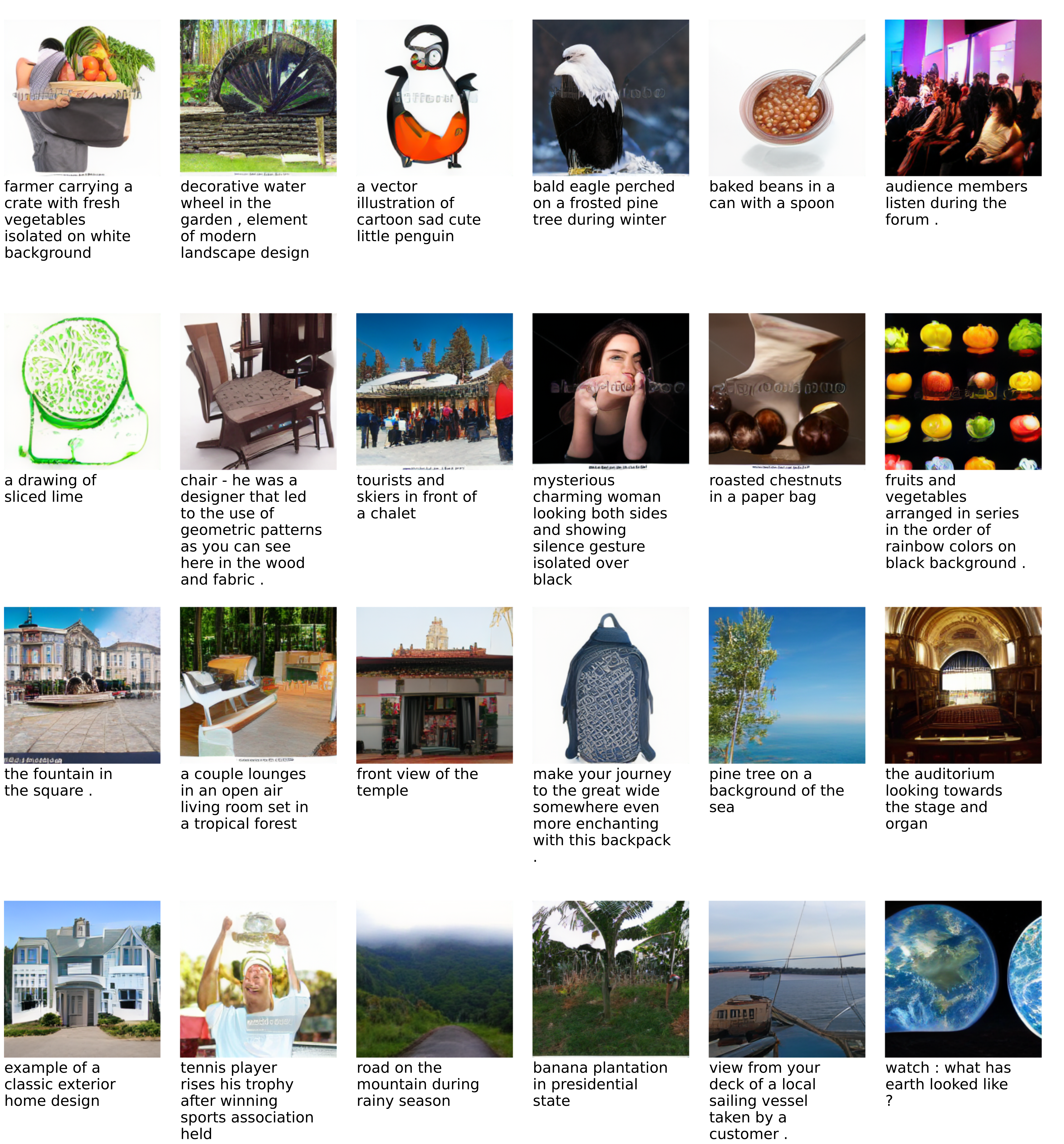}
    \caption{Additional examples of text-conditioned image generation by our model trained on CC-3M. Text prompts are randomly chosen from the validation data. Best 1 of 16 with re-ranking as in \cite{DALL-E, ImageBART}.}
    \label{fig:appendix_cc3m_val}
\end{figure}

We visualize the additional examples of text-conditioned image generation by RQ-Transformer trained on CC-3M. Figure \ref{fig:appendix_cc3m_control} shows the generated samples conditioned by various texts, which are unseen during training. Specifically, we manually choose four pairs of sentences, which share visual content with different contexts and styles, to validate the compositional generalization of our model. All images are sampled with top-$k$ 1024 and top-$p$ 0.9. Additionally, Figure \ref{fig:appendix_cc3m_val} shows the samples conditioned on randomly chosen texts from the validation set of CC-3M. For the images in Figure~\ref{fig:appendix_cc3m_val}, we use the re-ranking with CLIP similarity score \cite{CLIP} as in \cite{DALL-E} and \cite{ImageBART} and select the image with the highest CLIP score among 16 generated images.

\subsection{The Effects of Top-$k$ \& Top-$p$ Sampling on FID Scores}
\begin{figure}
    \centering
    \includegraphics[width=0.48\textwidth]{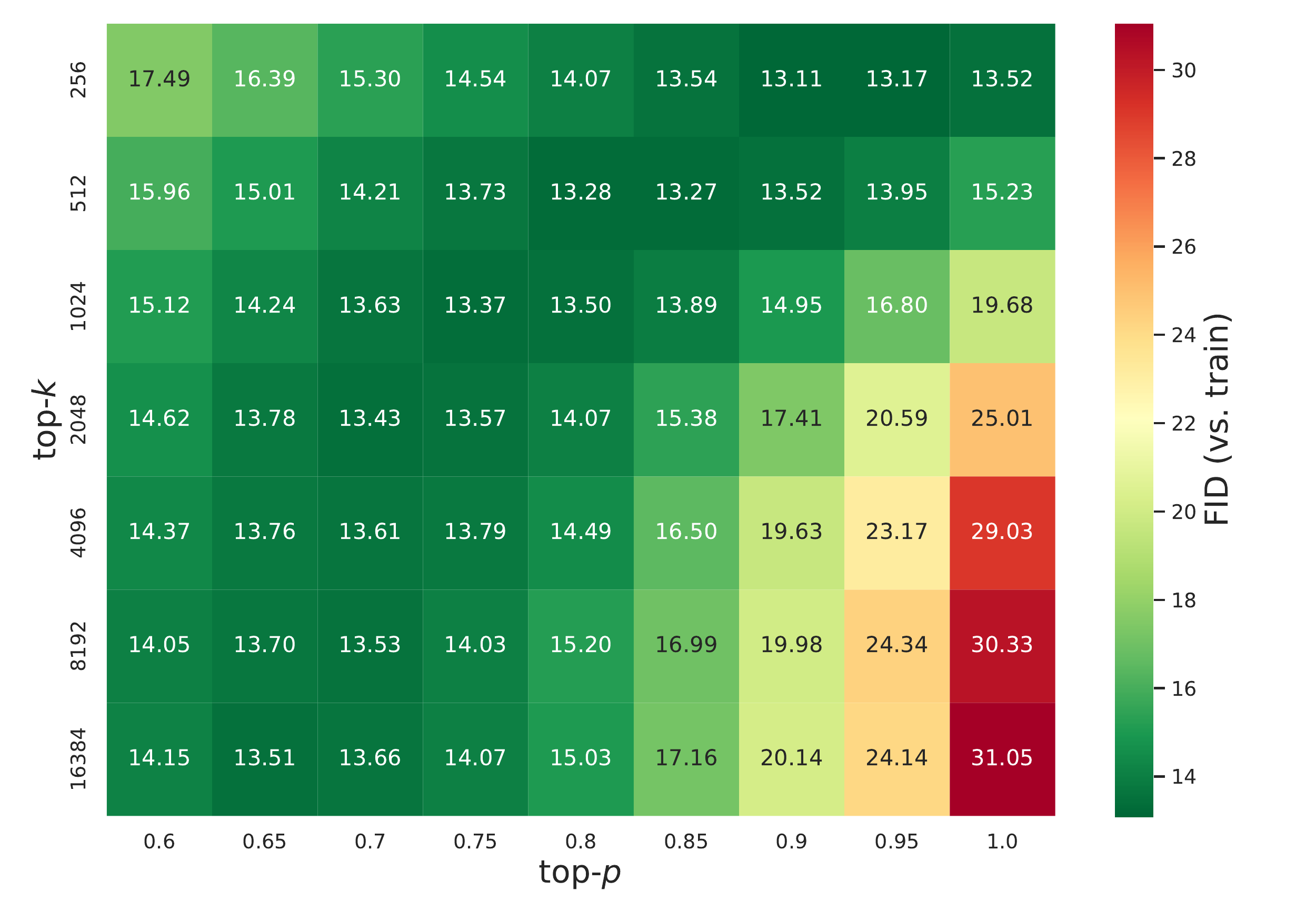}
    \includegraphics[width=0.48\textwidth]{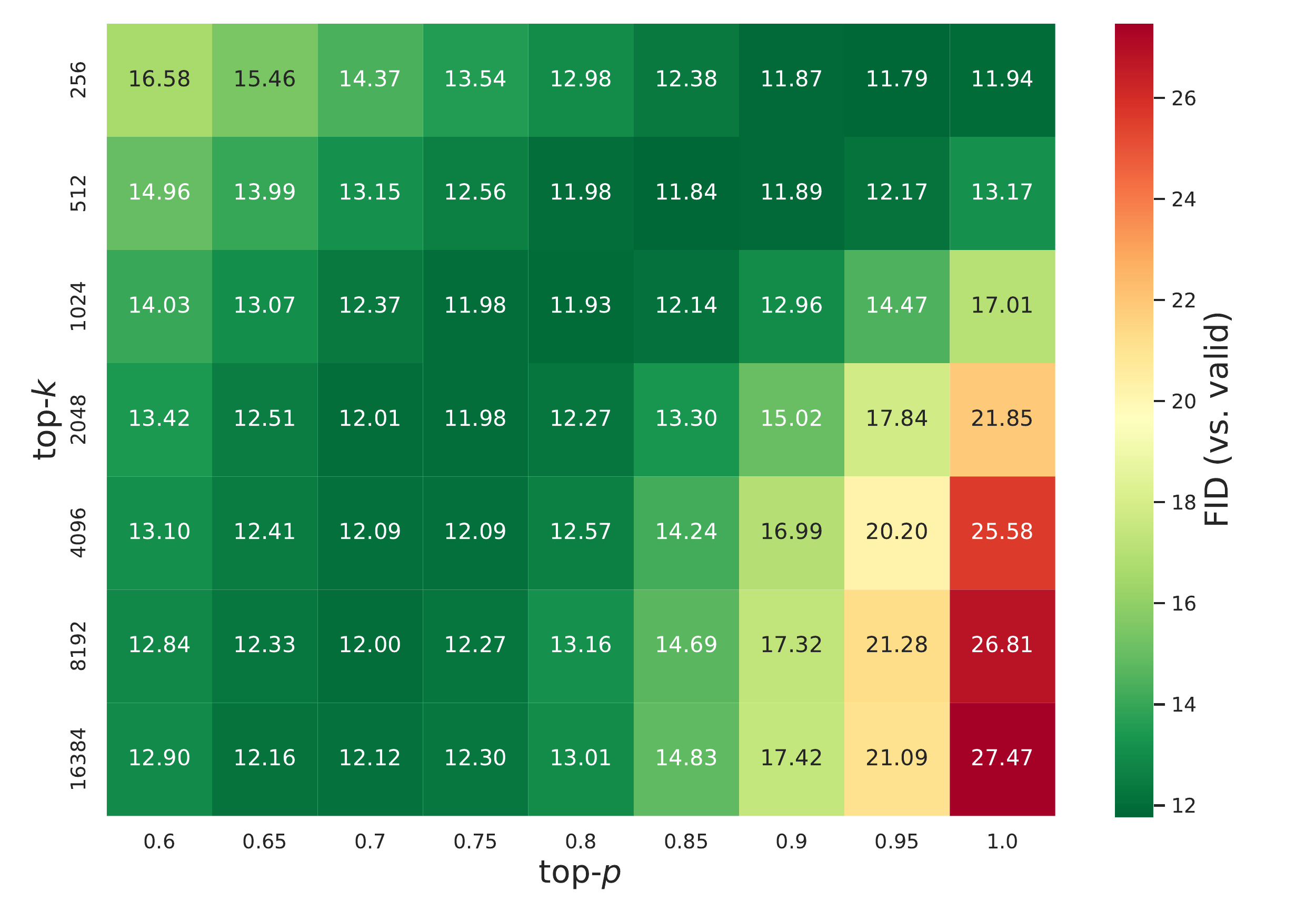}
    \caption{FID of 50K generated samples of RQ-Transformer (821M) against the training and the validation split of ImageNet.}
    \label{fig:appendix_fid_821m}
\end{figure}

\begin{figure}
    \centering
    \includegraphics[width=0.48\textwidth]{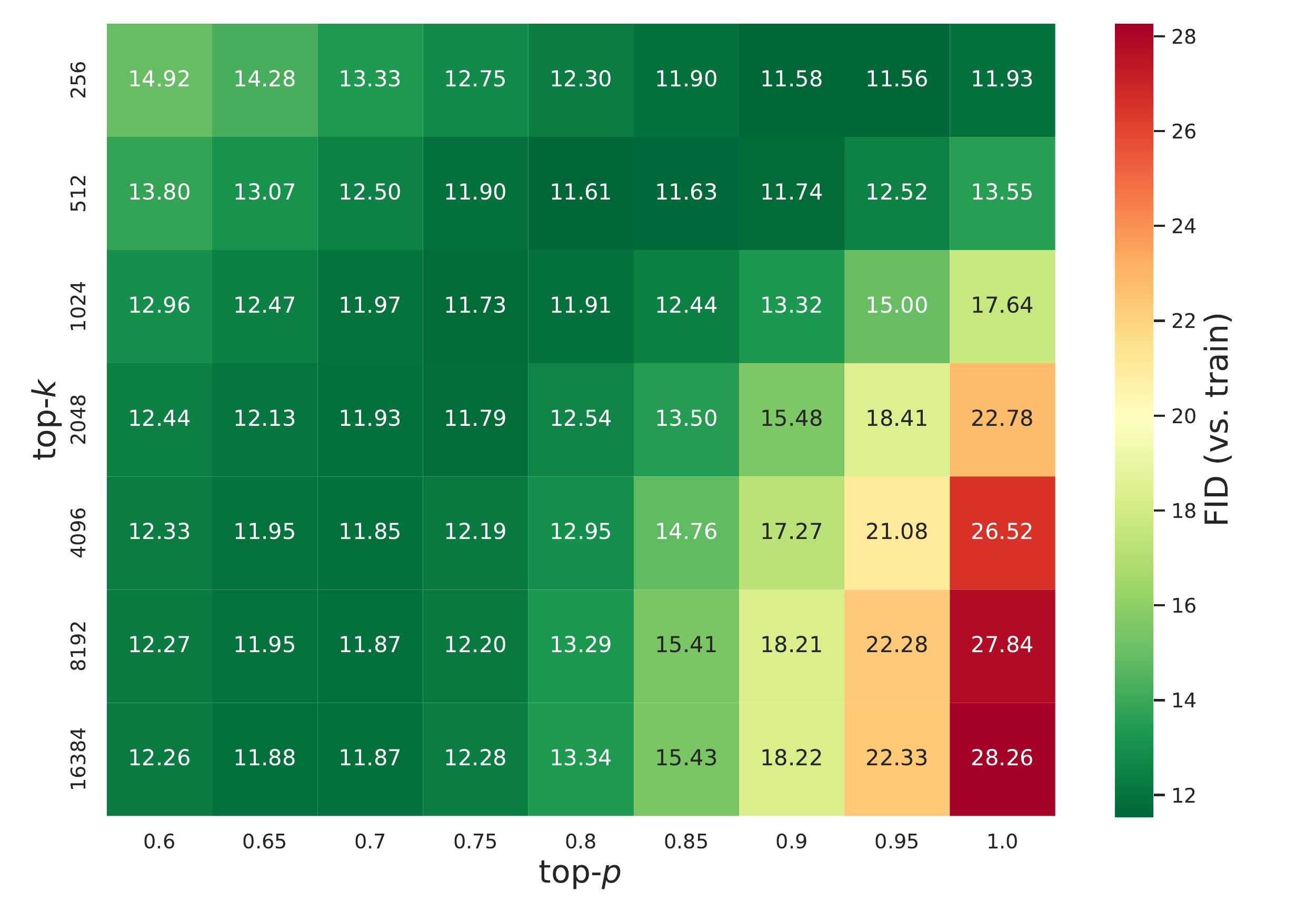}
    \includegraphics[width=0.48\textwidth]{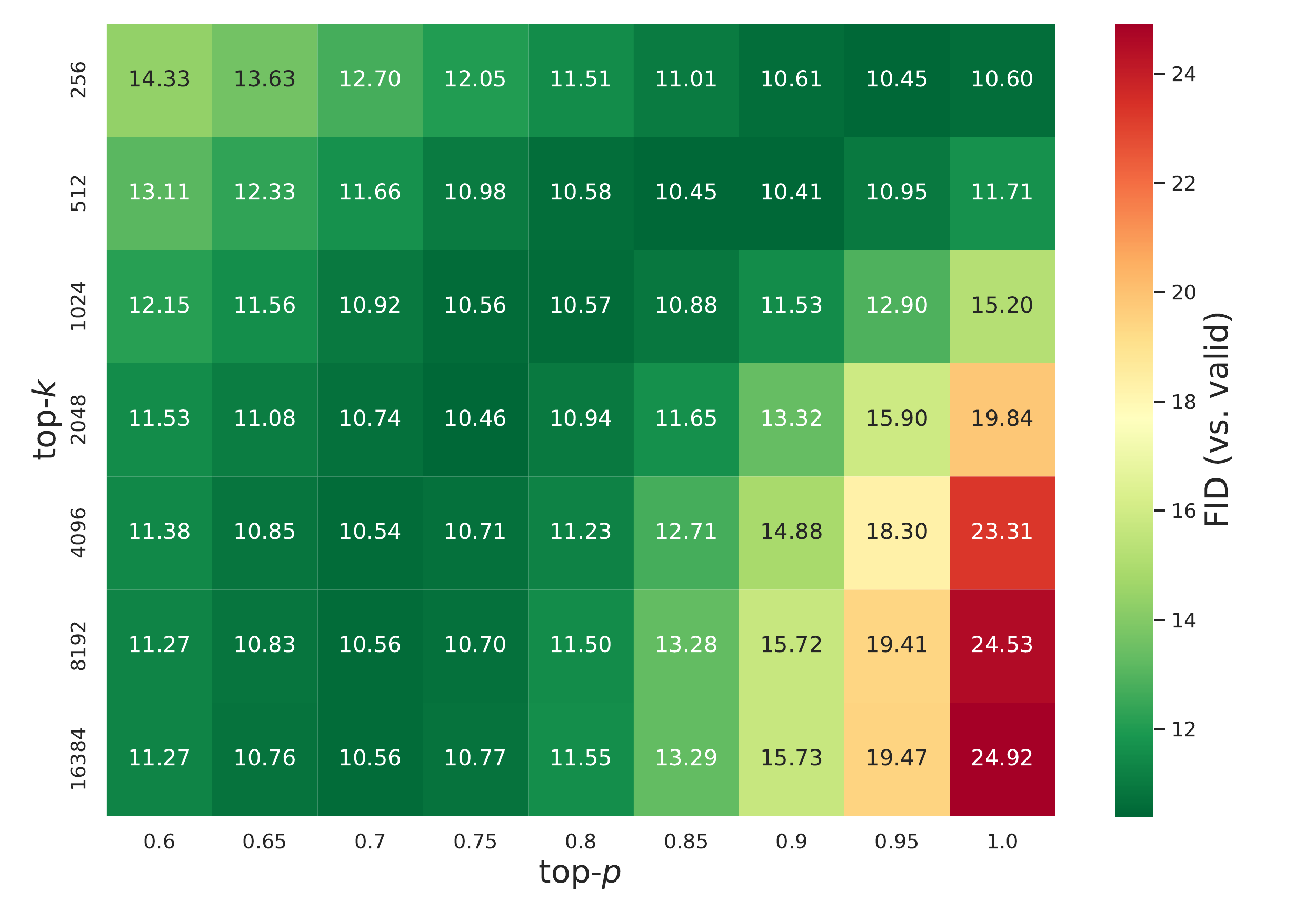}
    \caption{FID of 50K generated samples of RQ-Transformer (1.4B) against the training and the validation split of ImageNet.}
    \label{fig:appendix_fid_1400m}
\end{figure}

\begin{figure}
    \centering
    \includegraphics[width=0.48\textwidth]{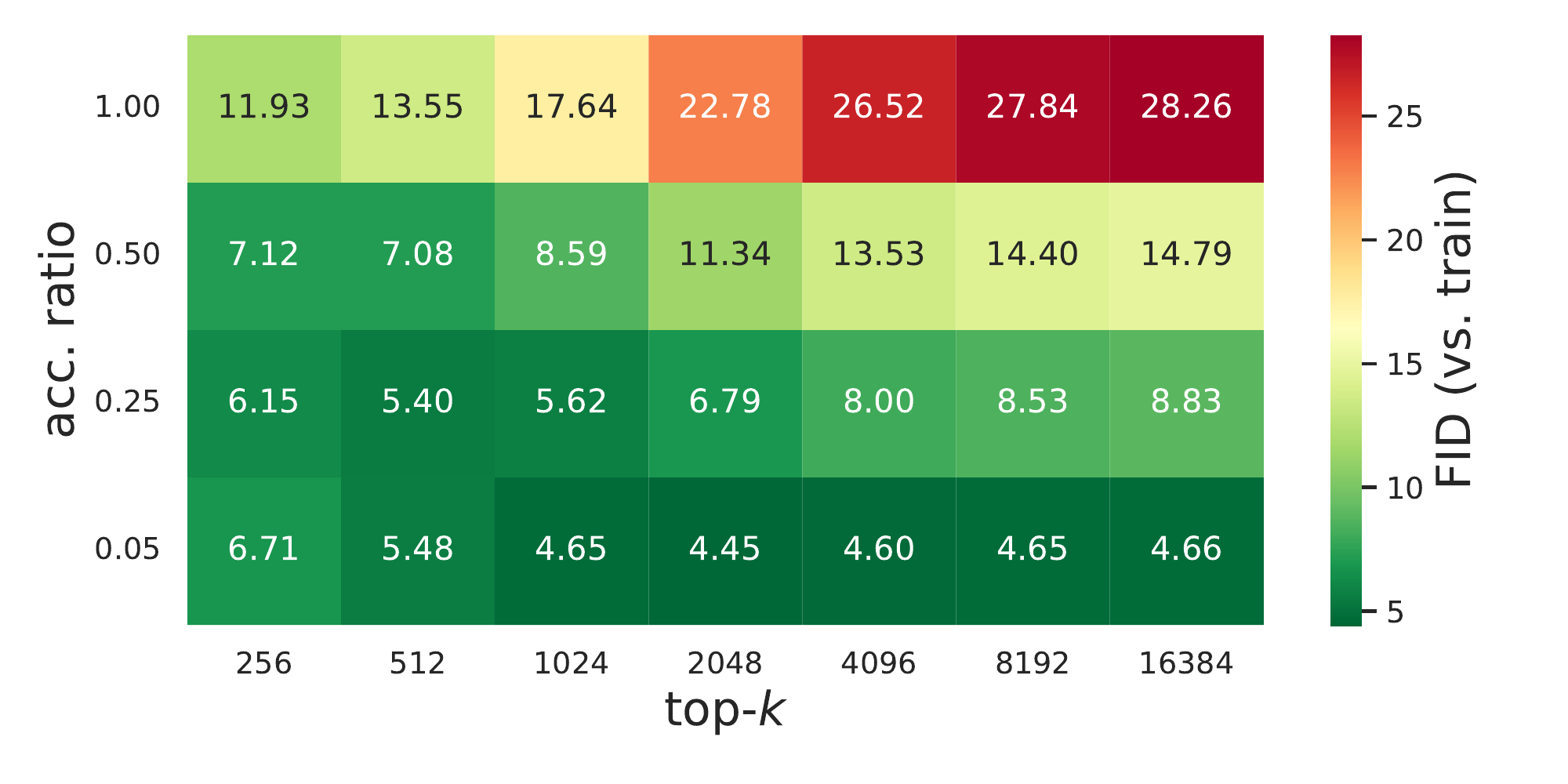}
    \includegraphics[width=0.48\textwidth]{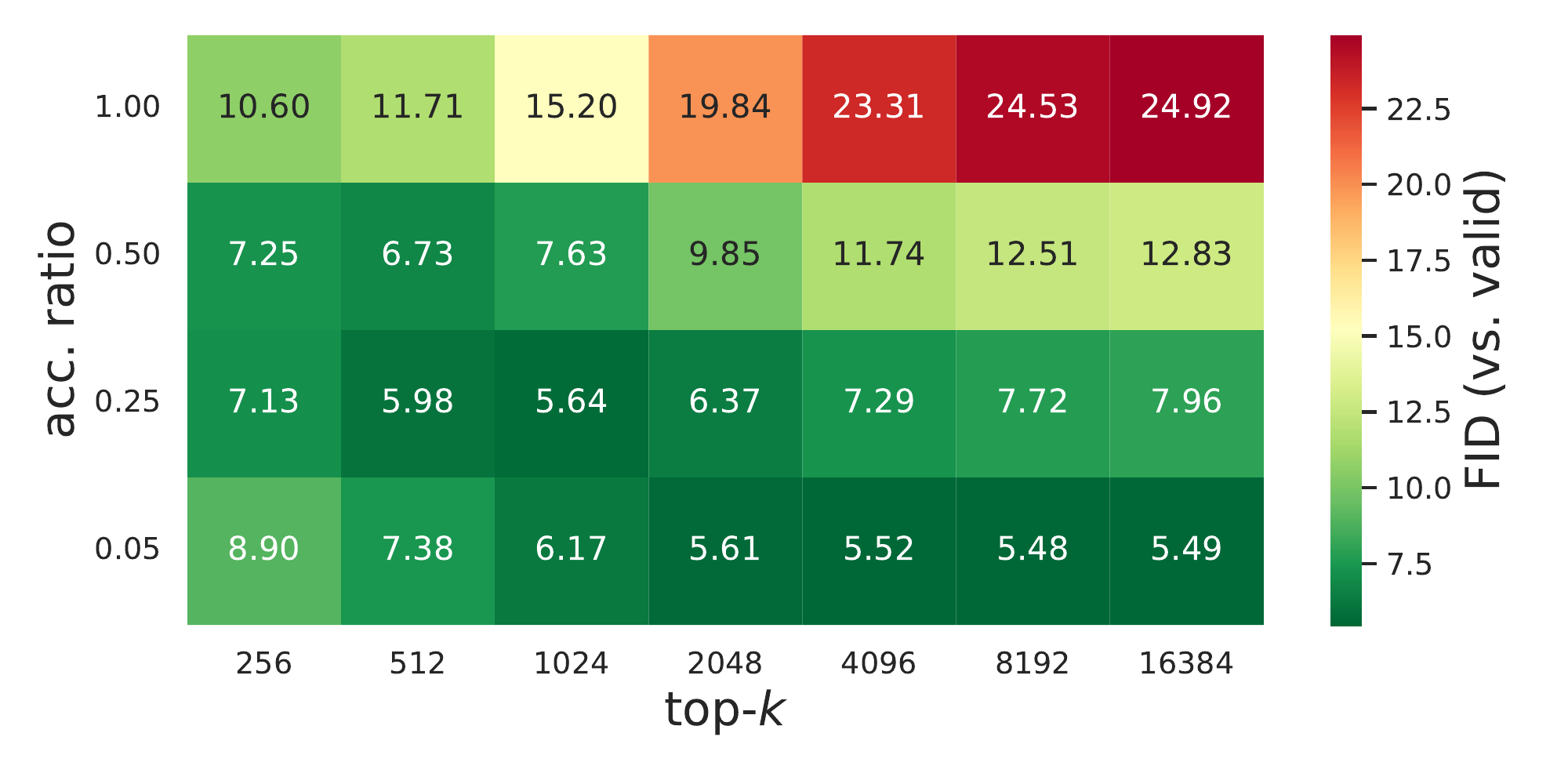}
    \caption{FID of rejection-sampled 50K samples of RQ-Transformer (1.4B) against the training and the validation split of ImageNet.}
    \label{fig:appendix_fid_1400m_rej}
\end{figure}

\begin{figure}
    \centering
    \includegraphics[width=0.48\textwidth]{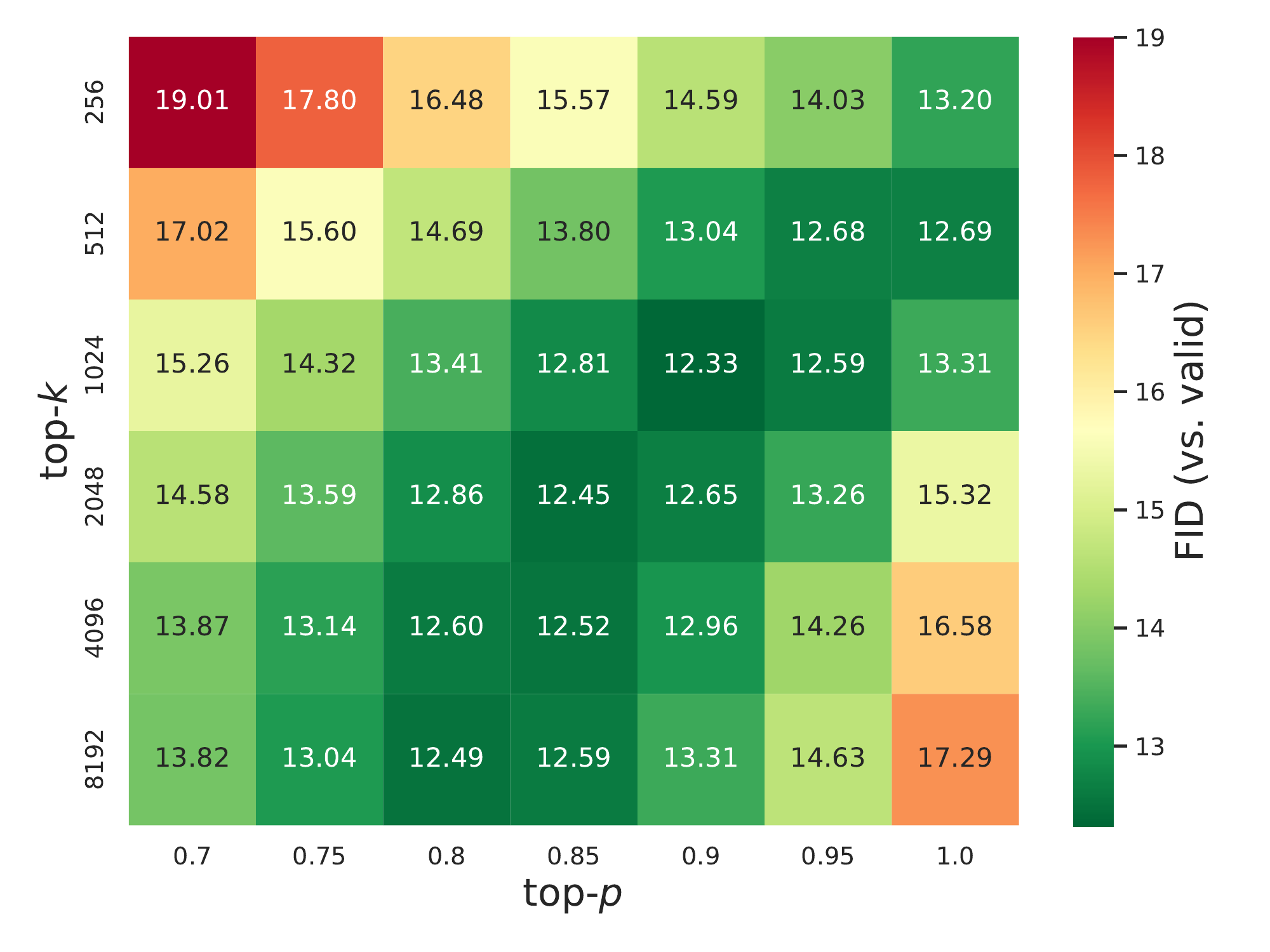}
    \includegraphics[width=0.48\textwidth]{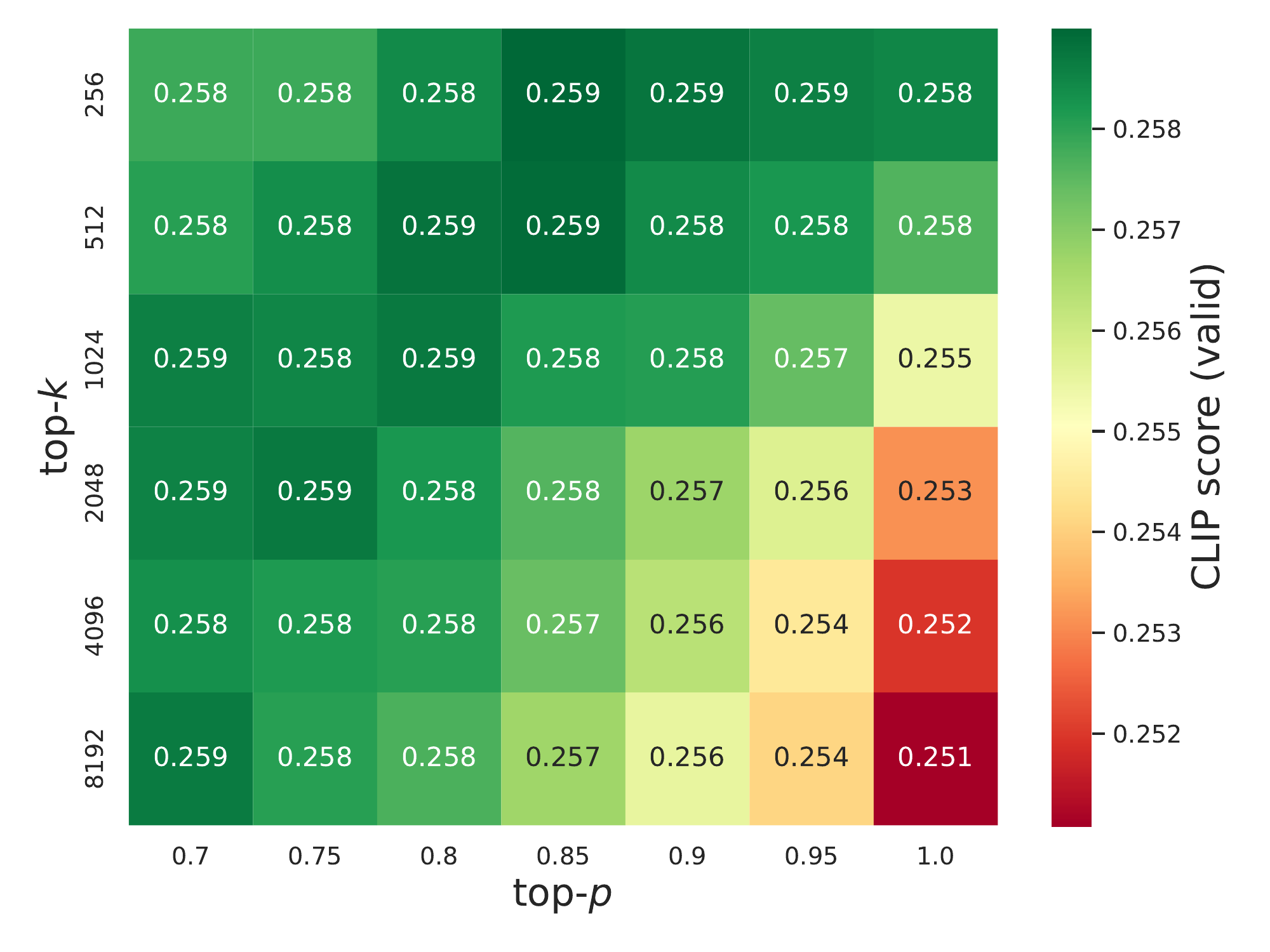}
    \caption{FID and CLIP score of RQ-Transformer (654M) on CC-3M, evaluated against the validation set. Images are generated conditioned on each sentence in the validation set.}
    \label{fig:appendix_fid_cc3m}
\end{figure}

In this section, we show the FID scores of the RQ-Transformer trained on ImageNet according to the choice of $k$ and $p$ for top-$k$ and top-$p$ sampling, respectively.
Figure~\ref{fig:appendix_fid_821m} and~\ref{fig:appendix_fid_1400m} shows the FID scores of 821M and 1400M parameters of RQ-Transformer according to different $k$s and $p$s.
Although we report the global minimum FID score, the minimum FID score at each $k$ is not significantly deviating from the global minimum.
For instance, the minimum FID attained by RQ-Transformer with 1.4B parameters is 11.58 while the minimum for each $k$ is at most 11.87.
When the rejection sampling of generated images is used to select high-quality images, Figure~\ref{fig:appendix_fid_1400m_rej} shows that higher top-$k$ values are effective as the acceptance rate decreases, since various and high-quality samples can be generated with higher top-$k$ values.
Finally, for the CC-3M dataset, Figure~\ref{fig:appendix_fid_cc3m} shows the FID scores and CLIP similarity scores according to different top-$k$ and top-$p$ values.


\section{Additional Results of Reconstruction Images by RQ-VAE}

\begin{table} 
\centering
\small
\caption{Results of coarse-to-fine approximation by the RQ-VAE with 8$\times$8$\times$4 shape of $\bM$. Reconstruction loss $\Lrec$, commitment loss $\Lcommit$, perceptual loss, and reconstruction FID (rFID) are measured on ImageNet validation data.}
\label{tab:appendix_coarse_to_fine}
\begin{tabular}{c|cccc}
\toprule 
$\hat{\bX}$ & $\Lrec$ & $\Lcommit$ & Perceptual loss & rFID \\ \hline
$G(\hat{\bZ}^{(1)})$ & 0.018 & 0.12 & 0.12 & 100.86 \\
$G(\hat{\bZ}^{(2)})$ & 0.014 & 0.10 & 0.090 & 22.74 \\
$G(\hat{\bZ}^{(3)})$ & 0.012 & 0.091 & 0.075 & 7.66 \\
$G(\hat{\bZ}^{(4)})$ & 0.010 & 0.082 & 0.068 & 4.73 \\
\bottomrule
\end{tabular}
\end{table}

\subsection{Coarse-to-Fine Approximation of Feature Maps by RQ-VAE}
\begin{figure}
    \centering
    \includegraphics[width=\textwidth]{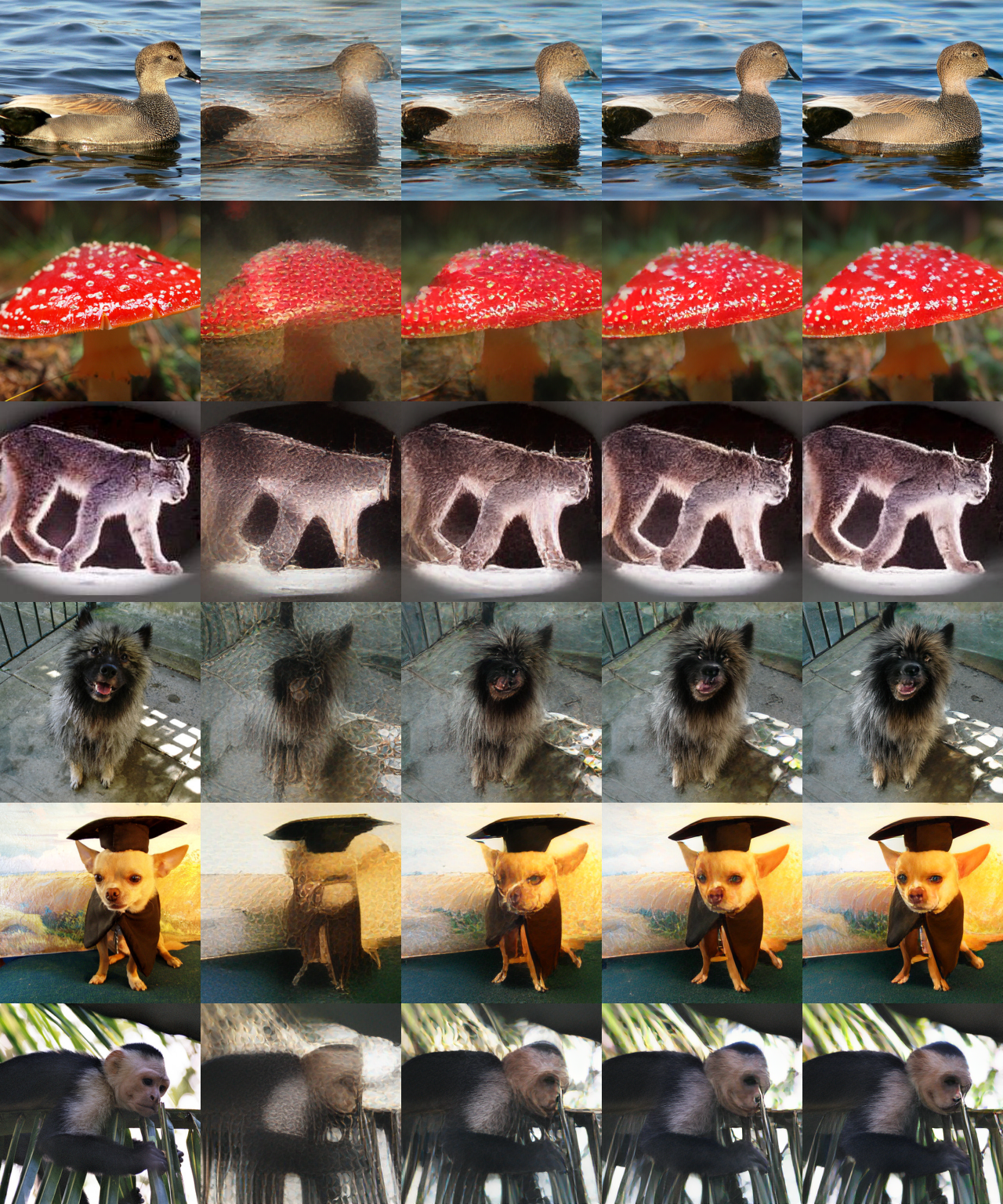}
    \caption{Additional examples of coarse-to-fine approximation by RQ-VAE with the 8$\times$8$\times$4 code map. The first example in each row is the original image, and the others are constructed from $\hat{\bZ}^{(d)}$ as $d$ increases.}
    \label{fig:appendix_additive_decoding}
\end{figure}

In this section, we further explain that RQ-VAE with depth $D$ conducts the coarse-to-fine approximation of a feature map.
Table~\ref{tab:appendix_coarse_to_fine} shows the reconstruction error $\Lrec$, the commitment loss $\Lcommit$, and the perceptual loss~\cite{johnson2016perceptual}, when RQ-VAE uses the partial sum $\hat{\bZ}^{(d)}$ of up to $d$ code embeddings for the quantized feature map of an image.
All three losses, which are the reconstruction and perceptual loss of a reconstructed image, and the commitment loss $\Lcommit$ of the feature map, monotonically decrease as $d$ increases.
The results imply that RQ-VAE can precisely approximate the feature map of an image, when RQ-VAE iteratively quantizes the feature map and its residuals.
Figure~\ref{fig:appendix_additive_decoding} also shows that the reconstructed images contain more fine-grained information of the original images as $d$ increases.
Thus, the experimental results validate that our RQ-VAE conducts the coarse-to-fine approximation, and RQ-Transformer can learn to generate the feature vector at the next position in a coarse-to-fine manner.

\subsection{The Effects of Adversarial and Perceptual Losses on Training of RQ-VAE}
\begin{figure}
    \centering
    \includegraphics[width=\textwidth]{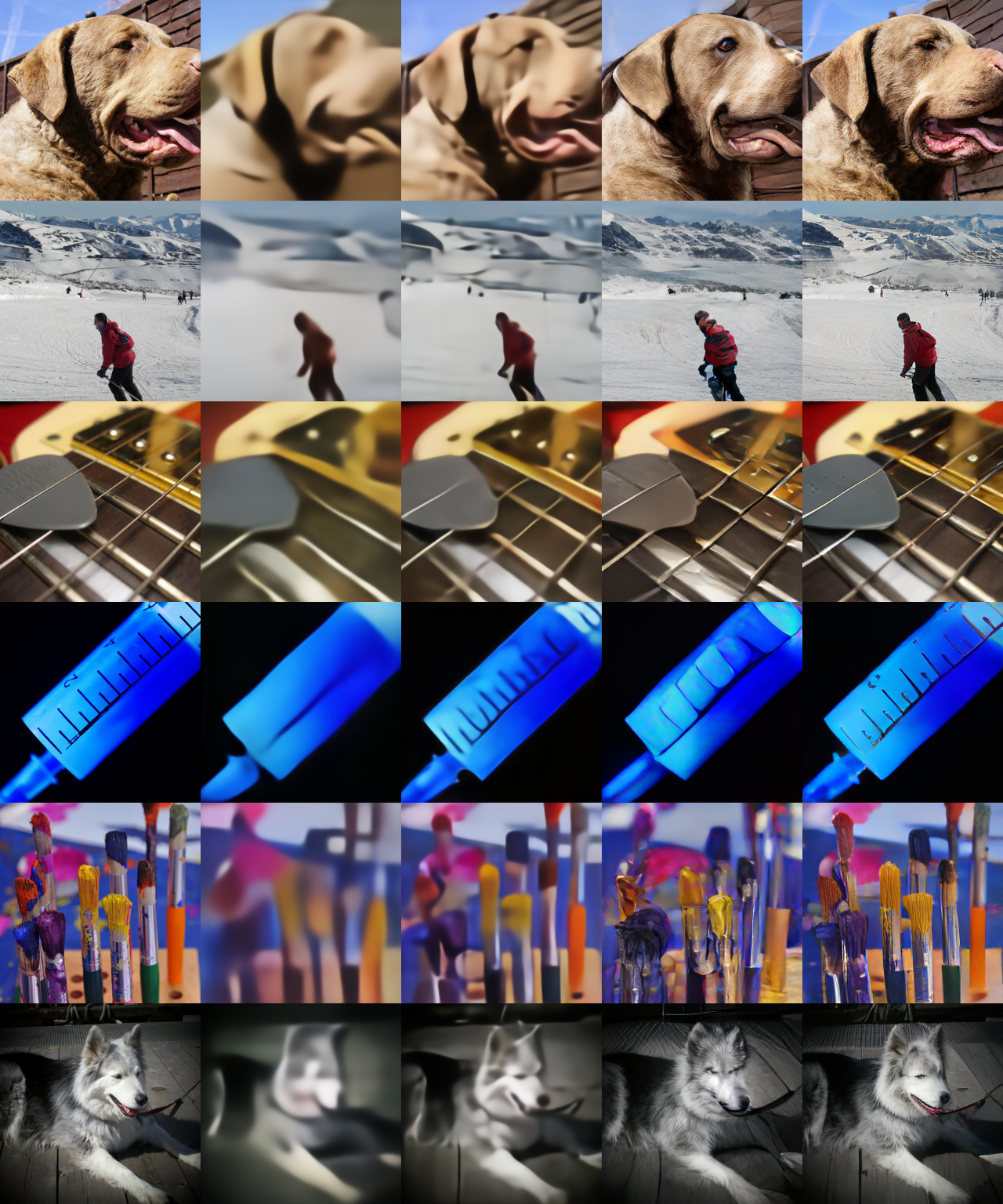}
    \caption{Reconstruction images by RQ-VAE with and without adversarial training. The first image in each row is the original image. The second and third images are reconstructed images by RQ-VAE without adversarial training. The second image is reconstructed by RQ-VAE using 8$\times$8$\times$1 code map, and the third image is reconstructed by RQ-VAE using 8$\times$8$\times$4 code map. The fourth and fifth images are reconstructed images by RQ-VAE with adversarial training. The fourth images are reconstructed by 8$\times$8$\times$1 code map, and the fifth images are reconstructed by 8$\times$8$\times$4 code map,}
    \label{fig:appendix_wo_gan}
\end{figure}

In Figure~\ref{fig:appendix_wo_gan}, we visualize the reconstructed images by RQ-VAEs, which are trained without and with adversarial and perceptual losses.
When the adversarial and perceptual losses are not used (the second and third columns), the reconstructed images are blurry, since the codebook is insufficient to include all information of local details in the original images.
However, despite the blurriness, note that RQ-VAE with $D=4$ (the third column) much improves the quality of reconstructed images than VQ-VAE (or RQ-VAE with $D=1$, the second column).

Although the adversarial and perceptual losses are used to improve the quality of image reconstruction, RQ is still important to generate high-quality reconstructed images with low distortion.
When the adversarial and perceptual losses are used in the training of RQ-VAEs (the fourth and fifth columns), the reconstructed images are much clear and include fine-grained details of the original images.
However, the reconstructed images by VQ-VAE (or RQ-VAE with $D=1$, the fourth column) include the unrealistic artifacts and the high distortion of the original images. 
Contrastively, when RQ with $D=4$ is used to encode the information of the original images, the reconstructed images by RQ-VAE (the fifth column) are significantly realistic and do not distort the visual information in the original images.

\subsection{Using $D$ Non-Shared Codebooks of Size $D/K$ for RQ-VAE}
\textcolor{black}{
As mentioned in Section~\ref{section:RQ}, a single codebook $\mathcal{C}$ of size $K$ is shared for every quantization depth $D$ instead of $D$ non-shared codebooks of size $D/K$. 
When we replace the shared codebook of size 16,384 with four non-shared codebooks of size 4,096, rFID of RQ-VAE increases from 4.73 to 5.73, since the non-shared codebooks can approximate at most $(K/D)^D$ clusters only. 
In fact, a shared codebook with $K$=4,096 has 5.94 of rFID, which is similar to 5.73 above.
Thus, the shared codebook is more effective to increase the quality of image reconstruction with limited codebook size than non-shared codebooks.
}

\end{document}